\RequirePackage{amsthm}
\documentclass[sn-mathphys,Numbered]{sn-jnl}

\usepackage{graphicx}%
\usepackage{multirow}%
\usepackage{amsmath,amssymb,amsfonts}%
\usepackage{amsthm}%
\usepackage{mathrsfs}%
\usepackage[title]{appendix}%
\usepackage{xcolor}%
\usepackage{textcomp}%
\usepackage{manyfoot}%
\usepackage{booktabs}%
\usepackage{bookmark}
\usepackage{algorithm}%
\usepackage{algorithmicx}%
\usepackage{algpseudocode}%
\usepackage{listings}%
\usepackage{caption}%

\usepackage{commath}
\usepackage{mathtools}
\usepackage{epigraph}
\usepackage{latexsym}
\usepackage[T1]{fontenc}

\theoremstyle{thmstyleone}%
\theoremstyle{thmstyletwo}%

\theoremstyle{thmstylethree}%

\begin{document}

\title[Utility Inspired Generalizations of TOPSIS]{Utility Inspired Generalizations of TOPSIS}

\author*[1]{\fnm{Izabela} \sur{Szcz\k{e}ch}}\email{iszczech@cs.put.poznan.pl}
\author[1]{\fnm{Robert} \sur{Susmaga}}\email{rsusmaga@cs.put.poznan.pl}

\affil[1]{\orgdiv{Institute of Computing Science}, \orgname{Poznan University of Technology}, \orgaddress{\street{Piotrowo 2}, \city{Poznan}, \postcode{60-965}, \country{Poland}}}

\abstract{TOPSIS, a popular method for ranking alternatives is based on aggregated distances to ideal and anti-ideal points. As such, it was considered to be essentially different from widely popular and acknowledged  `utility-based methods', which build rankings from weight-averaged utility values. 
Nonetheless, TOPSIS has recently been shown to be a natural generalization of these `utility-based methods' on the grounds that the distances it uses can be decomposed into so called weight-scaled means (WM) and weight-scaled standard deviations (WSD) of utilities. However, the influence that these two components exert on the final ranking cannot be in any way influenced in the standard TOPSIS.
This is why, building on our previous results, in this paper we put forward modifications that make TOPSIS aggregations responsive to WM and WSD, achieving some amount of well interpretable control over how the rankings are influenced by WM and WSD. The modifications constitute a natural generalization of the standard TOPSIS method because, thanks to them, the generalized TOPSIS may turn into the original TOPSIS or, otherwise, following the decision maker's preferences, may trade off WM for WSD or WSD for WM. In the latter case, TOPSIS gradually reduces to a regular `utility-based method'.
All in all, we believe that the proposed generalizations constitute
an interesting practical tool for influencing the ranking by controlled application of a new form of decision maker's preferences.}



\maketitle

\section{Introduction}
Multi-Criteria Decision Support Systems (MCDSS) assist decision makers in solving problems that analyze and process real-world objects (alternatives) evaluated on multiple conflicting attributes (criteria). What is often referred to as MCDA (Multi-Criteria Decision Aid) is a subfield of MCDSS concerned with, specifically, selecting the most preferred objects, assigning them to preference classes (called sorting), or ranking them; for an extended overview of MCDA techniques, models, and frameworks, see e.g. \cite{BS_02,Bisdorff15,Cinelli22,GEF_16,Ishizaka13}.

Among methods concerned with the task of ranking alternatives from the most preferred to the least preferred, a commonly chosen one is TOPSIS (Technique for Order Preference by Similarity to Ideal Solution)~\cite{HWAYOO81}.
This popular approach operates on the principle of distances to ideal and anti-ideal alternatives. The calculated distances are later processed using an aggregation function, referred to as the `relative closeness', which naturally renders a final ranking of alternatives. In this respect, TOPSIS from its early beginning diverged from methods based on the additive utility principle (shortly: `utility-based methods'), i.e. methods that in practice build their rankings using the weight-averaged values of utility, e.g. the SAW method \cite{HWAYOO81} or the UTA family of methods \cite{JLS-1978,LAGSIS82}. This seemingly irreconcilable difference had been retained till papers \cite{Susmaga_2023MSD,susmaga2023WMSD} showed that the distances to the ideal and anti-ideal points may be decomposed into what is referred to as the \textit{weight-scaled mean} (WM) and \textit{weight-scaled standard deviation} (WSD) of utilities. Under the assumption of linearity of utility functions used in `utility-based methods', the introduced relation establishes a fairly clear reconciliation of these methods and TOPSIS: by considering WSD \emph{in addition to} WM, TOPSIS naturally generalizes the `utility-based methods' (which thus use only WM). Notice however, that the relative effect of WM and WSD on the final ranking generated by the classic TOPSIS cannot be in any way influenced. This is unfortunate, as controlling this effect might be deeply useful in producing new, non-standard aggregations that will be either more or less dependent on WSD than the standard ones. 

To address that issue, this paper attempts to generalize TOPSIS so that the effect exerted by WM and WSD on the final ranking is explicitely controlled by the decision maker. This generalization will thus provide a tool for shifting TOPSIS towards or away from the `utility-based methods'.
Additionally, further modifications of TOPSIS are put forward to exploit other forms of interaction between WM and WSD.

The rest of the paper is organized as follows. In Section~\ref{sec:Related-works}, we elaborate on the selected methodological adaptations of TOPSIS that can be found in the literature to point out how our generalizations differ from the existing approaches. 
Next, in Section~\ref{sec:Formalization_TOPSIS_WMSD}, we recall our recently proposed formalization of the internal logic of TOPSIS (including the decomposition of distances) and a new, so-called WMSD-space, which allows to visually explain the most important aspects of the method. 
On these grounds, in Sections~\ref{sec:Elliptic-generalization} and ~\ref{sec:Lexicographic-generalization}, we put forward the generalizations of TOPSIS based on new aggregation functions, some of them redesigned to admit external parameters. 
These sections show how the parametrised aggregations make TOPSIS a much more versatile method, in particular by rendering its ranking mechanism arbitrarily close to `utility-based methods'. They also enumerate all potential limitations of the introduced generalizations.
All the introduced notions, including the new, parametrised aggregations are exemplified and illustrated with much detail in an extensive case study included in Section~\ref{sec:Case-Study}.
The paper ends with conclusions and lines of potentially interesting future investigations. 

\section{Related Works}
\label{sec:Related-works}
There are very many different aspects of TOPSIS that have so far been addressed and described. Among them, there are assorted adaptations of the method (\cite{ABOHADJAM19,KUO2017152}), issues with the method, e.g. the so-called rank-reversal problem (\cite{AIRES201984,MARROY06}), and numerous TOPSIS applications (\cite{BEHZADIAN201213051,Zavadskas16,Zyoud17}). 
Therefore the following brief list of TOPSIS-related papers will be confined 
to those papers that describe selected methodological adaptations of TOPSIS, 
in particular its extensions and generalizations.

A very popular type of generalizations concerns the form of the input data to the method. 
The most prominent here are the interval and fuzzy extensions, the latter being numerous enough
to merit their own surveys, e.g. \cite{NADABAN2016823}, which reviews the development 
of the fuzzy paradigm in TOPSIS, explores the method's different variants within this paradigm and presents multiple real-life applications. 

A similar kind of generalizations concerns the form of the preferential information
that is taken into account by the method to control its behaviour. Even though TOPSIS does not originally admit any
parameters to be controlled by explicit preferential information, this is exactly
what has been implemented in \cite{Zielniewicz13}, where preference-ordered pairs of alternatives are passed to TOPSIS. In this adaptation of the method the Euclidean distance measure is replaced with a parametrised distance measure, and explicit preferential information is used to construct such a version of the measure that will generate rankings compatible with the provided information. 
A reference to the distance measure constitutes a natural segue to other instances in which the  distance measure itself was generalized. First of all, the Minkowski distance measure is a natural generalization of the Euclidean distance measure, which in \cite{Wachowicz13} is assumed to be standard within TOPSIS, although several other distance measures are also suggested. Two other papers with alternative distance measures are: \cite{Vega14}, where the Euclidean measure was replaced with the Mahalanobis measure (allowing for correlated criteria), and \cite{Dmytrow18}, where the Euclidean measure was replaced with what is referred to as the GDM measure (allowing for mixed-domain criteria).

Finally, some developments of TOPSIS were aimed towards adapting the method
to solving different MCDA problems, in particular sorting. A good example of this
is described in \cite{Lima_Silva_20,Lima_Silva_23}, where TOPSIS-based methodology was applied to assigning alternatives to pre-defined quality classes. 

Although numerous modifications and adaptations of TOPSIS have been proposed, none of them fully explains the differences between this method and the `utility-based methods', in particular the differences concerning their ranking producing mechanisms. 
This paper fills this gap by exploiting the WMSD-space and putting forward adequate TOPSIS generalizations. 
A preliminary version of this paper was presented at MODeM'23 workshop~\cite{SS2023}.

\section{Formalization of the Inner Workings of TOPSIS}
\label{sec:Formalization_TOPSIS_WMSD}

Data processed by TOPSIS are provided in the form of a discrete set of $m$ alternatives from set $\mathbb{A}$ described in terms of $n$ criteria from set $\mathbb{C}$ (the $m \times n$ table with descriptions of alternatives described by criteria is usually referred to as the decision matrix). The descriptions of the alternatives are then weighted by non-uniform criterion weights $[w_1, w_2, ..., w_n]$ to incorporate criterion-specific preference information from a decision maker (if no weighting is required, then the weights are uniform, i.e. $w_j$ is set to $1$ for every $j \in \{1, 2, ..., n\}$). In the following, the vector of weights will be denoted as $\mathbf{w}$ (i.e. $\mathbf{w} = [w_1, w_2, ..., w_n]$) and all weights assumed to be positive. The individual elements of $\mathbf{w}$ are in practice used to multiply the alternative criterion descriptions. Additionally, the mean of those elements, denoted as $mean(\mathbf{w}) = \frac{\sum^{n}_{j=1} w_j}{n}$ will be used below for some scaling-related purposes. It is also consistently assumed that the counterdomains (ranges) of all criteria are real-valued intervals.

The processing itself assumes establishing two different points (`artificial alternatives'), further referred to as the ideal ($I$) and the anti-ideal ($A$). Next, the Euclidean distances between each alternative $a_i$ and these two points: $d_I = \delta(a_i,I)$ and $d_A = \delta(a_i,A)$, are calculated.
Finally, the distances are aggregated using a coefficient referred to as `relative closeness': 
\begin{equation}
  \mathsf{R}(d_I,d_A) = \frac{d_A}{d_A+d_I} = \frac{\delta(a_i,A)}{\delta(a_i,A)+\delta(a_i,I)}.  
\end{equation}
Notice that because $\mathsf{R}(d_I,d_A)$ expresses a share of $d_A$ in the sum of $d_A$ and $d_I$, its counterdomain is $[0, 1]$ (independently of the counterdomains of $d_A$ and $d_I$).
The resulting coefficients give rise to a total pre-order, which constitutes the final ranking of the alternatives generated by TOPSIS.

\begin{figure*}[!htb]
\centering
\includegraphics[width=0.95\textwidth]{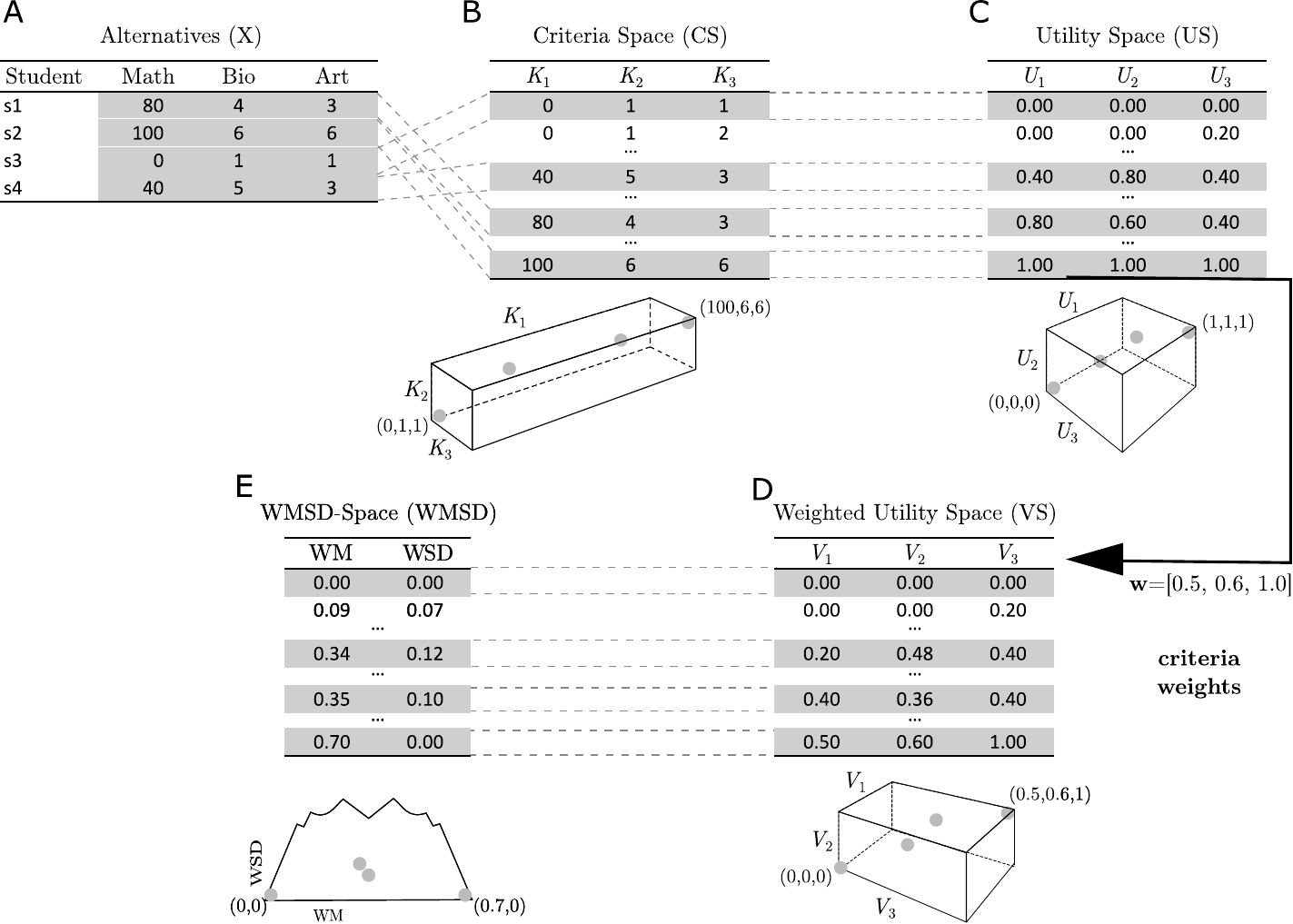}
\caption{A schema explaining different representations of data analyzed in this paper. (A) The original dataset (decision matrix) describing $m = 4$ students (alternatives) using final grades from $n = 3$ subjects (criteria). The same dataset is then depicted as a subset of (B) the criterion space, the set of all possible alternatives described by the three criteria describing students; (C) the utility space, the re-scaled equivalent of criterion space; (D) the weighted utility space, the preference-modified version of the utility space, with weights $\mathbf{w} = [0.5, 0.6, 1.0]$; (E) the WMSD-space, the space defined by the values of WM and WSD, with its shape determined by $\mathbf{w}$.}
\label{fig:all-spaces}
\end{figure*}

To provide the most general (and thus: data set independent) analyses and conclusions, in the following we consider all possible value combinations for a given set of criteria, i.e. the $n$-dimensional space of criteria ($CS$) as in Fig.~\ref{fig:all-spaces} (A~and~B). The approach is most general because every imaginable data set is in this way `covered' by $CS$ (formally: all images of every such set are included in $CS$).

Since working with criteria that vary in their ranges and types (`gain', `cost') may be troublesome, all the criteria are subjected to linear \emph{min-max transformations} \cite{Susmaga_2023MSD}. As a result, the criterion space is transformed to what will be referred to as an $n$-dimensional utility space $US = [0, 1] \times [0, 1] \times ... \times [0, 1]$, in which vector $\mathbf{0}$ is the image of $A$, while vector $\mathbf{1}$ is the image of $I$ (Fig.~\ref{fig:all-spaces}C). 
To incorporate the process of criteria weighting, modifications are applied to the components of $US$. In practice the $[0, 1]$ ranges are re-scaled by criterion weights from $\mathbf{w} = [w_1, w_2, ..., w_n]$ to form $[0, w_j]$ ranges. In result, $US$ is transformed to what will be referred to as $n$-dimensional weight-scaled utility space $VS = [0, w_1] \times [0, w_2] \times ... \times [0, w_n]$, in which vector $\mathbf{0}$ is the image of $A$, while vector $\mathbf{w}$ is the image of $I$ (Fig.~\ref{fig:all-spaces}D). 
Consequently, the `relative closeness' coefficient will now be established  on the basis of distances calculated in $VS$. 
Notice that the transformations presented simplify the interpretability of the analyses but in no way reduce their generality. For further details regarding the spaces see \cite{Susmaga_2023MSD,susmaga2023WMSD}. 

As observed in \cite{susmaga2023WMSD}, TOPSIS may be viewed from a new perspective that is founded on the mean and standard deviation of the $VS$-based vectors representing the alternatives.
In particular, the Euclidean distances between each alternative $a_i$ and the ideal/anti-ideal points may be equivalently expressed with what is referred to as \emph{weight-scaled mean} (WM) and \emph{weight-scaled standard deviation} (WSD) (both of which are functions of a $\mathbf{v} \in VS$) as: 
\begin{align}
 d_I &= \sqrt{(mean(\mathbf{w})-\text{WM})^2+\text{WSD}^2},\\
 d_A &= \sqrt{\text{WM}^2+\text{WSD}^2}. 
\end{align}
Consequently, also the `relative closeness' coefficient may be now re-expressed with WM and WSD as:
\begin{equation}
\mathsf{R}(\text{WM},\text{WSD}) = \frac{\sqrt{\text{WM}^2+\text{WSD}^2}}{\sqrt{\text{WM}^2+\text{WSD}^2}+\sqrt{\big(mean(\mathbf{w})-\text{WM}\big)^2+\text{WSD}^2}}.
\end{equation}


It should be stressed that using the introduced WM and WSD to calculate $\mathsf{R}$ does not influence the method's results.
Instead, it simply lends the method a new, more interpretable perspective. This is because easily comprehensible descriptions like mean and variance of an alternative are better at construing  the inner workings of TOPSIS than the originally calculated Euclidean distances.
Moreover, as shown in ~\cite{Susmaga_2023MSD,susmaga2023WMSD}, WM and WSD can be naturally used to visualize the WMSD-space as a $2$-dimensional area with WM on the x-axis and WSD on the y-axis (Fig.~\ref{fig:all-spaces}E). The space is always maximally $2$-dimensional, regardless of the number of criteria, and as such can always be presented in a plane.
\begin{figure*}[!htb]
\centering
\includegraphics[width=0.5\textwidth]{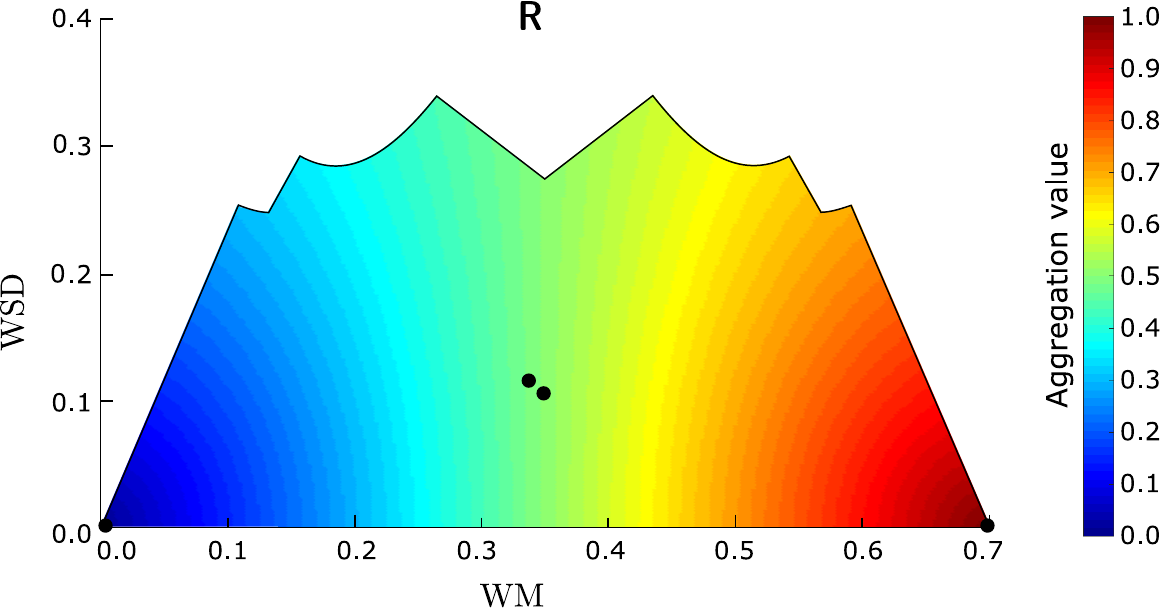}
\caption{Four exemplary alternatives (students) depicted in WMSD-space defined for $\mathbf{w}=[0.5, 0.6, 1.0]$ for aggregation $\mathsf{R}$. Color encodes the aggregation value, with blue representing the least preferred and red the most preferred values.}
\label{fig:colored_R}
\end{figure*}

WMSD-space is thus an effective tool for visualizing whole spaces and particular alternatives. What is more, WMSD-space may also be used to visualize the values of TOPSIS aggregation functions, e.g. $\mathsf{R}$, as by colouring each point one can encode the aggregation value (Fig.~\ref{fig:colored_R}). Moreover, since given a vector of weights WMSD-space presents all obtainable values of WM and WSD for a given number of criteria, it allows for very general, and thus dataset independent analyses. 

The decomposition of distances used in TOPSIS into the WM and WSD components makes it absolutely clear that the method uses WM to construct the ranking as its final result. This fact actually makes it a version of the utility-based methods, i.e. methods that rate and rank alternatives according to weighted versions of utility, e.g. SAW or UTA. However, the same decomposition also reveals WSD, the second component used by the method. Using WSD to construct the ranking may be regarded as the \emph{key feature} of TOPSIS, differentiating it from methods based on WM only. The simultaneous usage of WM and WSD makes TOPSIS a natural generalization of a typical utility-based method ~\cite{susmaga2023WMSD,SS2023}.

Interestingly, even though TOPSIS uses both WM and WSD, the degrees at which WM and WSD influence the rating of an alternative depend on their particular values and cannot be in any way controlled by the decision maker.
In what follows, we show how the relative influence of WSD on the method's result may be gradually increased (effectively promoting WSD over WM) or gradually decreased (effectively promoting WM over WSD). In result, we put forward new versions of TOPSIS. In particular, in one of such versions the influence of WSD is decreased from its original degree down to zero,  effectively producing more and more utility-based-like versions of TOPSIS.

\section{Elliptic generalizations of TOPSIS aggregations}
\label{sec:Elliptic-generalization}

Having laid the foundations in \cite{Susmaga_2023MSD,susmaga2023WMSD}, where it was shown that TOPSIS constitutes a natural generalization of `utility-based methods', in this paper we move one step further and introduce a natural generalization of TOPSIS itself. This generalization will concern the influence of WM and WSD on the final ranking and by using appropriate parametrization it will enable controlling their relative trade-off within this ranking, a feature that is unfortunately absent in the classic version of the method. Because in terms of the ranking producing mechanism TOPSIS is similar, though not identical, to the `utility-based methods'\footnote{Other differences exist, regarding, however, mainly the procedures of generating utility functions. These may be quite complex with the `utility-based methods'; e.g. methods of the UTA family utilize intricate instances of mathematical programming to turn pieces of explicit preference information (provided by the decision maker) into the final form of the utility functions. The simplest of the `utility-based methods' seems to be the~SAW method, in which the utility functions are assumed to be linear (as such, SAW may be thus viewed as the `utility-based method' closest to TOPSIS).}, the generalization will actually allow to shift TOPSIS towards or away from these methods.

\subsection{TOPSIS aggregations}
\label{sec:TOPSIS_aggregations}
The classic TOPSIS `relative closeness' function simultaneously aggregates \emph{two} distances of an alternative to the ideal/anti-ideal points and then uses the aggregation to rate and rank the alternatives. Interestingly, it is fully legitimate to also use any of the individual distances alone for the same purpose. To this end, we define two separate aggregation functions, each based on a \emph{single} distance: aggregation $\mathsf{I}$ based on the distance to the ideal point, and aggregation $\mathsf{A}$ based on the distance to the anti-ideal point. 
Analogously to aggregation $\mathsf{R}
$, these aggregations may also be expressed as functions of WM and WSD:
\begin{align} 
\mathsf{I}(\text{WM},\text{WSD}) &= 1 - \frac{\sqrt{\big(mean(\mathbf{w})-\text{WM}\big)^2+\text{WSD}^2}}{mean(\mathbf{w})}, \\
\mathsf{A}(\text{WM},\text{WSD}) &= \frac{\sqrt{\text{WM}^2+\text{WSD}^2}}{mean(\mathbf{w})}.
\end{align}
The negation (`$1 - ...$') in $\mathsf{I}$ is used only to make it a function to be maximized, as is the case with $\mathsf{A}$ and $\mathsf{R}$. Additionally, the division by $mean(\mathbf{w})$ in $\mathsf{I}$ and $\mathsf{A}$ serves only to unify their counterdomains: from $[0, mean(\mathbf{w})]$ to $[0, 1]$, as is the case with and $\mathsf{R}$.

Because the defined $\mathsf{I}$ and $\mathsf{A}$, similarly to $\mathsf{R}$, are real-valued functions, the resulting ranking of alternatives from $\mathbb{A}$ will have the form of a linear pre-order implied by how the alternatives are rated by the aggregations. As such, the pre-order will consist of a total (in other words: linear) compound relation $\succ \cup \sim$, with $\succ$ denoting preference (formally: a strict order; irreflexive and transitive) and $\sim$ denoting indifference (formally: an equivalence; reflexive, transitive and symmetric). If $G$ denotes any of the three aggregations and values $\text{WM}_i, \text{WSD}_i$ characterize alternative $a_i$, relations $\succ$ and $\sim$ are defined as follows: 
\begin{itemize}
\item $a_1 \succ a_2$ holds when $G(\text{WM}_1,\text{WSD}_1) > G(\text{WM}_2,\text{WSD}_2)$ (higher/lower rating), 
\item $a_1 \sim a_1$ holds otherwise (i.e. when $G(\text{WM}_1,\text{WSD}_1) = G(\text{WM}_2,\text{WSD}_2)$ (equal rating)\footnote{The equality is determined up to the precision of the numerical representation of the results of $G$.}).
\end{itemize}
The resulting ranking is thus $(\mathbb{A},\succ \cup \sim)$.

Now, because the results of ranking methods (e.g. TOPSIS) are naturally viewed as rankings rather than individual relations between pairs of alternatives (as those shown above), rating-related and ranking-related (rather than relation-related) terminology will be used throughout the paper, e.g. `$a_1$ is rated higher than $a_2$' or `$a_1$ is ranked higher than $a_2$' instead of `$a_1$ is preferred to $a_2$' or $a_1 \succ a_2$'.

Although $\mathsf{R}$ is predominant in practical implementations of TOPSIS, any of the three aggregations may equally well be used. Incidentally, the interest in the elementary aggregations grows in recently published papers, e.g. \cite{SPOTIS_2020}, where an analogue of $\mathsf{I}$ is exclusively used.
It should be stressed that $\mathsf{R}$ is in fact a `composite' aggregation of $\mathsf{I}$ and $\mathsf{A}$ (the `elementary' aggregations), and thus inherits its main properties from them. Thus in this paper, all three aggregations are described and examined `in parallel'. 

\begin{figure*}[!htb]
\centering
\includegraphics[width=0.98\textwidth]  {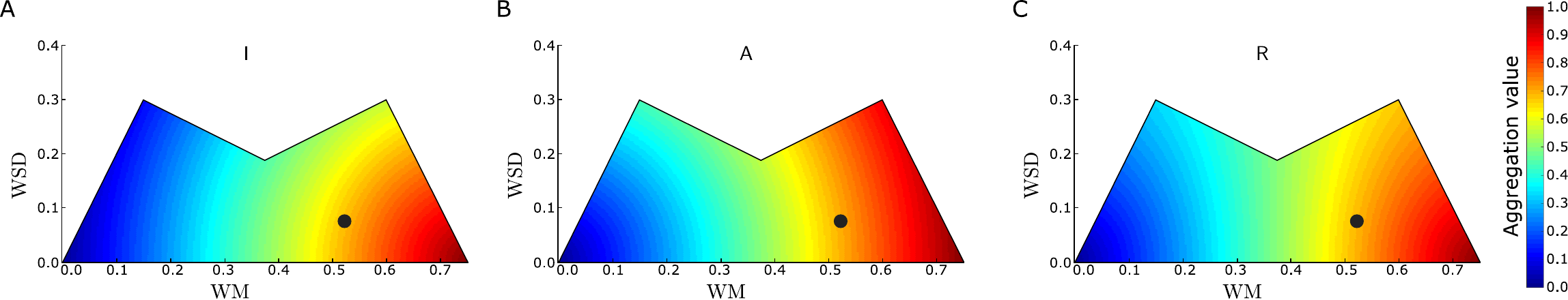}
\caption{An exemplary point $\mathbf{v} = [0.75, 0.25]$ depicted in WMSD-space defined by $\mathbf{w} = [1.0, 0.5]$  for aggregations (A) $\mathsf{I}$, (B) $\mathsf{A}$ and (C) $\mathsf{R}$. Colour encodes the aggregation value, with blue representing the least preferred and red the most preferred values. The isolines of aggregation $\mathsf{I}$ and $\mathsf{A}$ are concentric circles with center in $[mean(\mathbf{w}), 0]$ and $[0, 0]$, respectively. The isolines of $\mathsf{R}$ form two arch-like curves `centred' in $[mean(\mathbf{w}), 0]$ and $[0, 0]$.}
\label{fig:WMSD-colours}
\end{figure*}

Notice that because $\mathsf{I}$ and $\mathsf{A}$ are basically defined as a Euclidean distance from a predefined point, their isolines in $VS$ constitute concentric, $n$-dimensional hyperspheres around $\mathbf{w}$ and $\mathbf{0}$, respectively (none of which could easily be visualized for $n > 3$, incidentally).
Owing to the specific construction of WMSD-space though, the isolines of the aggregations reduce to two-dimensional curves. In particular, the isolines of $\mathsf{I}$ and $\mathsf{A}$ are simply concentric circles centred in $[mean(\mathbf{w}), 0]$ and $[0, 0]$, respectively (see panels A and B of Fig.~\ref{fig:WMSD-colours}), while the isolines of the `composite' $\mathsf{R}$ are two groups of arch-like curves `centred' in $[mean(\mathbf{w}), 0]$ and $[0, 0]$ (see panel C of Fig.~\ref{fig:WMSD-colours}). These curves are circle-like close to their centres, but straighten up towards the middle of the WMSD-space. This results in the `midpoint' isoline being just a vertical segment.

Interestingly, the isolines in WMSD space clearly depict (see Fig.~\ref{fig:WMSD-colours}) the \textit{preference-related interplay} of WM and WSD in the context of the analyzed aggregations~\cite{susmaga2023WMSD}. In particular, assuming constant WSD,
WM is of `gain' type as aggregations $\mathsf{I}$ , $\mathsf{A}$ and $\mathsf{R}$ are always increasing with the increase of WM. On the other hand, under constant WM,  WSD is of `gain' type only for the aggregation $\mathsf{A}$ and for $\mathsf{R}$ when $\text{WM} < \frac{mean(\mathbf{w})}{2}$. It is of `cost' type for the aggregation $\mathsf{I}$ and for $\mathsf{R}$ when $\text{WM} > \frac{mean(\mathbf{w})}{2}$. Then the decrease of WSD results in a increase of the aggregation value. The vertical isoline for $\text{WM} = \frac{mean(\mathbf{w})}{2}$ is the place where the $\mathsf{R}$ aggregation does not depend on WSD at all (i.e. where WSD is neutral).

\subsection{Circular aggregations}
If $i, a, r \in (0,1)$ are to represent some given values of aggregations $\mathsf{I}$, $\mathsf{A}$, $\mathsf{R}$, respectively, then the function $\text{WSD(WM)}$ describing the $2$-dimensional isoline of the given value is in the three cases as follows:
\begin{align}
\text{the case of } \mathsf{I}: \text{WSD(WM)} &= \sqrt{(mean(\mathbf{w})(1-i))^2 - (mean(\mathbf{w}) - \text{WM})^2},\\
\text{the case of } \mathsf{A}: \text{WSD(WM)} &= \sqrt{(mean(\mathbf{w})a)^2 - \text{WM}^2},\\
\text{the case of } \mathsf{R}: \text{WSD(WM)} &= \sqrt{\frac{(mean(\mathbf{w})r - \text{WM})(mean(\mathbf{w})r - (2r - 1)\text{WM})}{2r - 1}}.     
\end{align}
Notice that while the formulae in the case of $\mathsf{I}$ and $\mathsf{A}$ simply express centred circles (precisely: `positive' semicircles, because the square root is non-negative), which are defined for every $i, a \in (0,1)$, the formula in the case of `composite' aggregation $\mathsf{R}$ expresses a more sophisticated curve that is defined for every $r \in (0,\frac{1}{2}) \cup (\frac{1}{2},1)$. However, both for $r \rightarrow \frac{1}{2}^{-}$ as well as $r \rightarrow \frac{1}{2}^{+}$ the shape of the curve converges to a vertical line $\text{WM} = \frac{mean(\mathbf{w})}{2}$, which constitutes the isoline of $\mathsf{R}$ for $r = \frac{1}{2}$, where the WSD is neutral (see panel C of Fig.~\ref{fig:WMSD-colours}). 
For obvious reasons, the name \emph{circular} will be applied to aggregations $\mathsf{I}$, $\mathsf{A}$ and, for consistency, also to aggregation $\mathsf{R}$, even though its isolines are different from circles (in this case the name will be quoted).

\subsection{Elliptic aggregations}
\label{sec:Elliptic}
Since the `composite' aggregation $\mathsf{R}$ is based on $\mathsf{I}$ and $\mathsf{A}$, let us initially focus on the two latter.
Isolines of these two aggregations have the form of (differently centred) circles. Now, because a natural generalization of a circle is an ellipse, the undertaken approach to generalizing $\mathsf{I}$ and $\mathsf{A}$ was to redesign their isolines from circles to ellipses. This was implemented by introducing into their $\text{WSD(WM)}$ formulae a scaling coefficient, denoted by $\epsilon$. Analogous modification was applied to the $\text{WSD(WM)}$ formula of $\mathsf{R}$. 

The resulting generalized formulae are as follows:
\begin{align}
\text{the case of }\mathsf{I}: \text{WSD(WM)} &= \epsilon \cdot \sqrt{(mean(\mathbf{w})(1-i))^2 - (mean(\mathbf{w}) - \text{WM})^2},\\
\text{the case of } \mathsf{A}: \text{WSD(WM)} &= \epsilon \cdot  \sqrt{(mean(\mathbf{w})a)^2 - \text{WM}^2},\\
\text{the case of } \mathsf{R}: \text{WSD(WM)} &= \epsilon \cdot \sqrt{\frac{(mean(\mathbf{w})r - \text{WM})(mean(\mathbf{w})r - (2r - 1)\text{WM})}{2r - 1}}.   
\end{align}
Generalizations of aggregations $\mathsf{I}$ and $\mathsf{A}$, denoted as $\mathsf{I}^\epsilon$ and $\mathsf{A}^\epsilon$, will have isolines in the form of ellipses (precisely: `positive' semi-ellipses) and thus will be referred to as \emph{elliptic}. For consistency, the same name will be applied to the generalization of $\mathsf{R}$, denoted as $\mathsf{R}^\epsilon$, even though its isolines are different from ellipses (in this case the name will be quoted). For the sake of distinction, aggregations $\mathsf{A}$, $\mathsf{I}$ and $\mathsf{R}$ will be from now on referred to as \emph{classic}.

The allowed range of $\epsilon$ is $(0, +\infty)$ for $\mathsf{R}^\epsilon$, while $(E, +\infty)$ for $\mathsf{I}^\epsilon$ and $\mathsf{A}^\epsilon$, where $E > 0$ is a lower limit (see the Appendix~\ref{app:operational_ranges} for the justification and derivation of this limit). The`neutral' value of $\epsilon$ is $1$, with $\epsilon > 1$ promoting WM over WSD, and $\epsilon < 1$ promoting WSD over WM (`promoting' in the sense: `increasing its influence above the original level').
As expected, when $\epsilon = 1$ (the `neutral' value), all the elliptic aggregations, i.e. $\mathsf{I}^\epsilon$, $\mathsf{A}^\epsilon$ and $\mathsf{R}^\epsilon$, reduce to their classic counterparts, i.e. $\mathsf{I}$, $\mathsf{A}$ and $\mathsf{R}$, respectively (as presented in Fig.~\ref{fig:aggregations-for-theta-050}).
When WM is to be promoted over WSD, the resulting isolines become elongated vertically, as depicted in Fig.~\ref{fig:aggregations-for-theta-065}. On the other hand, when WSD is to be promoted over WM, they become elongated horizontally, as in Fig.~\ref{fig:aggregations-for-theta-041}.

Applying $\epsilon \neq 1$ not only affects the level of (either horizontal or vertical) elongation of the aggregations' isolines, but may also affect the counterdomains of the aggregations. Fortunately, this does not happen when $\epsilon$ is maintained with its allowed ranges, i.e. when $\epsilon \in (0, +\infty)$ for $\mathsf{R}^\epsilon$ and $\epsilon \in (E, +\infty)$ for $\mathsf{I}^\epsilon$ and $\mathsf{A}^\epsilon$.

It must be kept in mind, though, that although introducing $\epsilon \neq 1$ always affects the values of the aggregations, it need not always affect the resulting rankings. In particular, given any aggregation, the ordering of any set of alternatives characterized by common WSD (images of such alternatives are arranged horizontally in the WMSD-space) or by common WM (images of such alternatives are arranged vertically in the WMSD-space) will remain the same for every allowed value of $\epsilon$.

\subsection{Special cases of the elliptic aggregations}

\begin{figure*}[!htb]
\centering
\includegraphics[width=0.98\textwidth]  {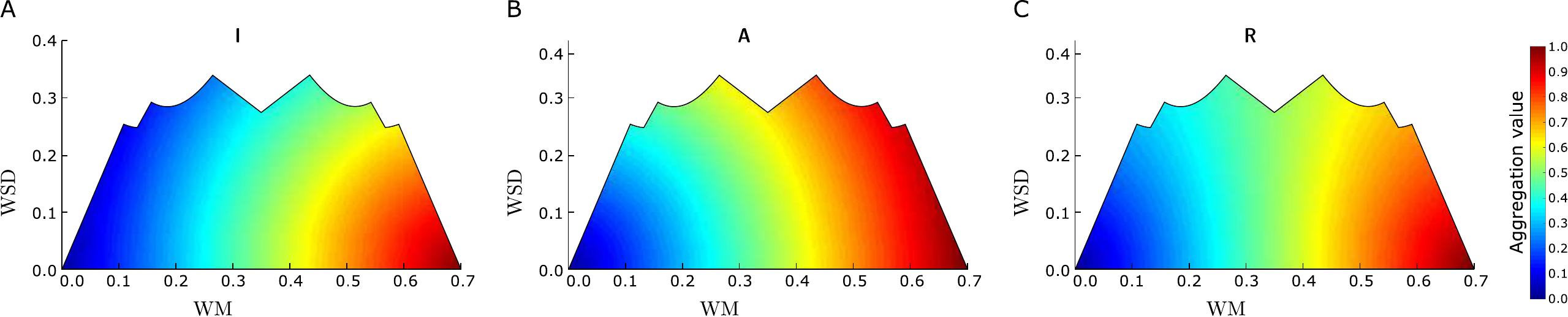}
\caption{WMSD-space defined by $\mathbf{w} = [1.0, 0.6, 0.5]$ depicted against circular aggregations: (A) $\mathsf{I}$, (B) $\mathsf{A}$ (C) $\mathsf{R}$ (equivalent to the corresponding elliptic aggregations for $\epsilon = 1$).}
\label{fig:aggregations-for-theta-050}
\end{figure*}

\begin{figure*}[!htb]
\centering
\includegraphics[width=0.98\textwidth]  {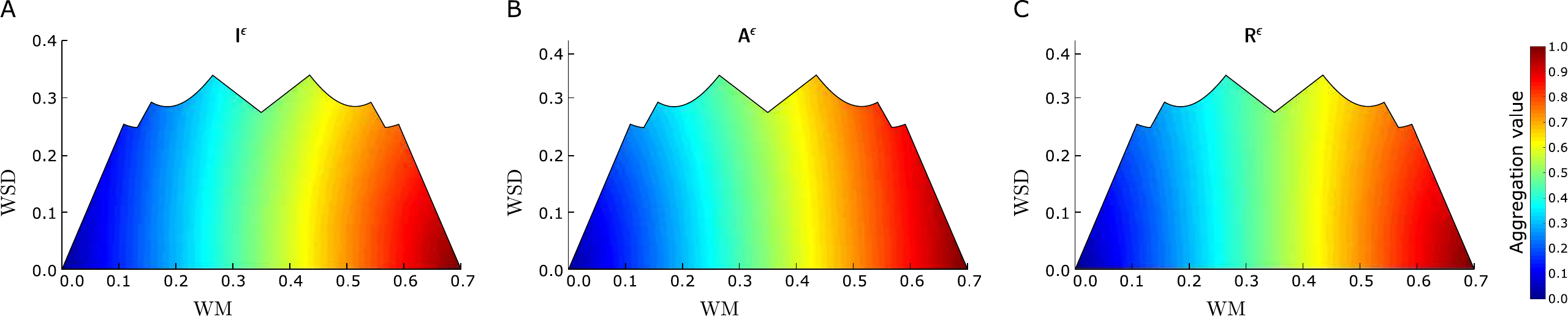}
\caption{WMSD-space defined by $\mathbf{w} = [1.0, 0.6, 0.5]$ depicted against elliptic aggregations for $1 < \epsilon = 1.86 < +\infty$: (A) $\mathsf{I}^\epsilon$, (B) $\mathsf{A}^\epsilon$ (C) $\mathsf{R}^\epsilon$.}
\label{fig:aggregations-for-theta-065}
\end{figure*}

\begin{figure*}[!htb]
\centering
\includegraphics[width=0.98\textwidth]  {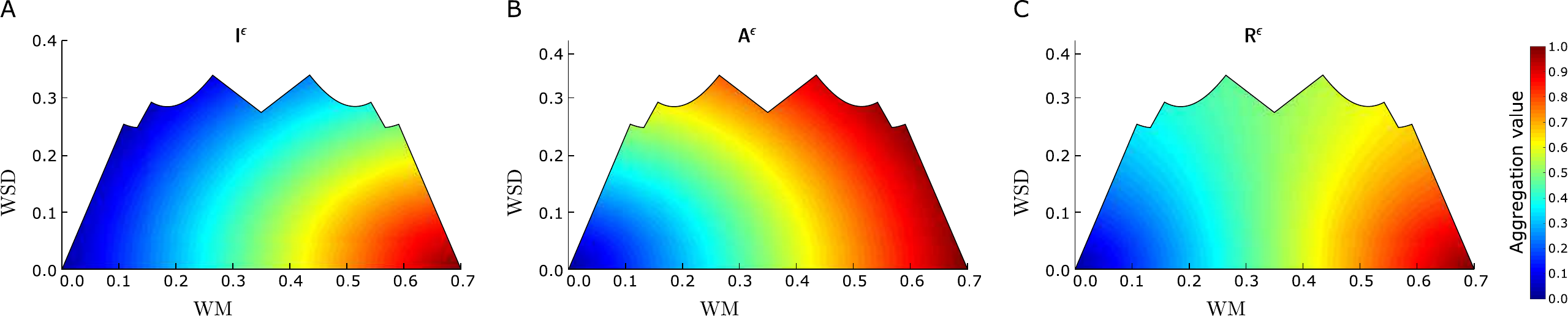}
\caption{WMSD-space defined by $\mathbf{w} = [1.0, 0.6, 0.5]$ depicted against elliptic aggregations for $E = 0.68 < \epsilon = 0.69 < 1$: (A) $\mathsf{I}^\epsilon$, (B) $\mathsf{A}^\epsilon$ (C) $\mathsf{R}^\epsilon$.}
\label{fig:aggregations-for-theta-041}
\end{figure*}

As far as the $\mathsf{I}$ aggregation is concerned, an alternative's rating (expressed by the value of the aggregation function) increases with the decrease of the WSD when WM is kept unchanged~\cite{susmaga2023WMSD}. This is clear e.g. with a WMSD-based visualization presented in Fig.~\ref{fig:aggregations-for-theta-050}, where 
moving any point vertically downwards (which corresponds to decreasing the WSD of the corresponding alternative) invariably makes this point coincide with isolines characterized by higher values of the aggregation.
Now, making the isolines of the aggregation more and more vertical, e.g. when $\epsilon$ grows from $1.00$ (see Fig.~\ref{fig:aggregations-for-theta-050}) to $1.86$ (see Fig.~\ref{fig:aggregations-for-theta-065}), decreases the influence of WSD (after an identical vertical move the same point will experience a smaller change in the aggregation value). In result it becomes easier to affect the rating of alternatives by changing the WM (which corresponds to a horizontal move) rather than WSD. A reverse situation can be of course observed when the isolines become more and more horizontal, e.g. when $\epsilon$ drops from $1.00$ (see Fig.~\ref{fig:aggregations-for-theta-050}) to $0.69$ (see Fig.~\ref{fig:aggregations-for-theta-041}).

\begin{figure}[!htb]
\centerline{
\includegraphics[width=0.35\textwidth]  {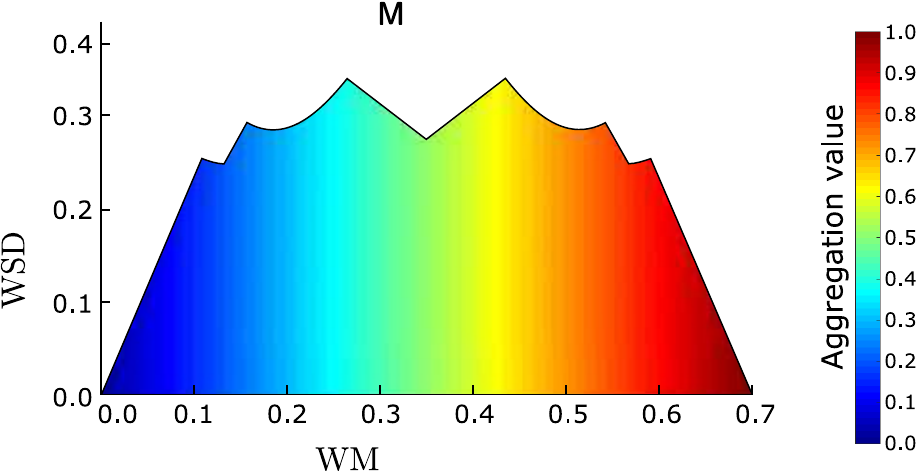}
}
\caption{WMSD-space defined by $\mathbf{w} = [1.0, 0.6, 0.5]$ depicted against aggregation $\mathsf{M} = \text{WM}$. Notice its full independence of WSD.}
\label{fig:aggregations-for-theta-100}
\end{figure}

Summarizing, by incorporating $\epsilon$ in their formulae, the new, elliptic aggregations produce natural generalizations of TOPSIS. These generalizations have the following special cases. 
\begin{itemize}
\item Under $\epsilon = 1$, all the elliptic aggregations reduce to circular aggregations (with $\mathsf{I}^\epsilon$ and $\mathsf{A}^\epsilon$ producing circles instead of ellipses, as in Fig.~\ref{fig:aggregations-for-theta-050}). This simply illustrates the fact that $\mathsf{I}$, $\mathsf{A}$ and $\mathsf{R}$ constitute special cases of $\mathsf{I}^\epsilon$, $\mathsf{A}^\epsilon$ and $\mathsf{R}^\epsilon$, respectively.
\item Under $\epsilon$ minimal (i.e. $\epsilon = E$ for $\mathsf{I}^\epsilon$ and $\mathsf{A}^\epsilon$; and $\epsilon \rightarrow 0$ for $\mathsf{R}^\epsilon$) all the elliptic aggregations maximally promote WSD over WM (ellipses of $\mathsf{I}^\epsilon$ and $\mathsf{A}^\epsilon$ are elongated horizontally to their extremes; the same effect concerns the shapes of $\mathsf{R}^\epsilon$), producing `possibly horizontal' isolines (see Fig.~\ref{fig:aggregations-for-theta-041}).
\item Under $\epsilon$ maximal, i.e. $\epsilon \rightarrow +\infty$, all the elliptic aggregations converge to one that in practice considers only WM (ellipses of $\mathsf{I}^\epsilon$ and $\mathsf{A}^\epsilon$ are elongated vertically to their extremes; the same effect concerns the shapes of $\mathsf{R}^\epsilon$), producing `increasingly vertical' isolines. In limit ($\epsilon = +\infty$) this is equivalent to employing a very specific, new aggregation: $\mathsf{M} = \text{WM}$, which is depicted in Fig.~\ref{fig:aggregations-for-theta-100}. Observe that $\mathsf{M}$ constitutes a common special case of $\mathsf{I}^\epsilon$, $\mathsf{A}^\epsilon$ and $\mathsf{R}^\epsilon$.
\end{itemize}

As it turns out, $\mathsf{M}$ is additionally completely equivalent to a version of `relative closeness' in which the Euclidean distance measure is substituted with the Manhattan distance measure. The `in limit' situation of $\epsilon = +\infty$ is especially interesting because TOPSIS with this particular aggregation loses its key feature (i.e. taking into account the standard deviation of the utilities) and, as far as the ranking producing mechanisms is concerned, starts behaving exactly like the `utility-based methods'. 
Otherwise, i.e. for $\epsilon \in (E,  +\infty)$ with $\mathsf{A}^\epsilon$ and $\mathsf{I}^\epsilon$, and for $\epsilon \in (0,  +\infty)$ with $\mathsf{R}^\epsilon$, TOPSIS equipped with the generalized aggregations allows for a precisely controlled trade-off between WM and WSD in the resulting rankings. 

\section{Lexicographic generalizations of TOPSIS aggregations}
\label{sec:Lexicographic-generalization}
Consider any generally understood aggregation $G: \mathbb{A} \rightarrow C$ that operates by aggregating elements of a set $\mathbb{A}$ of alternatives to a counter-domain $C$ in order to define on $\mathbb{A}$ a preferential linear pre-order $(\mathbb{A},\succ \cup \sim)$, a union of preference and indifference, where $\succ$ denotes preference, $\sim$ denotes indifference and the union $\succ \cup \sim$ is total.

Let $a_1, a_2 \in \mathbb{A}$ be two alternatives and let $rep(a_1), rep(a_2)$ denote some real-valued representations of these alternatives. In this paper the typical representations are vectors from $VS$.
If:
\begin{itemize}
\item $C \subseteq (-\infty,+\infty)^1 = (-\infty,+\infty)$, then the aggregation will be referred to as \emph{one-dimensional} ($G$ generates a single value). The relations $\succ$ and $\sim$ are then defined for the $a_1, a_2$ as follows: 
	\begin{itemize}
	\item $a_1 \succ a_2$ holds when $G(rep(a_1)) > G(rep(a_2))$, 
	\item $a_1 \sim a_2$ holds otherwise (i.e. when $G(rep(a_1)) = G(rep(a_2))$).
	\end{itemize}
\item $C \subseteq (-\infty,+\infty)^2 = (-\infty,+\infty) \times (-\infty,+\infty)$, then the aggregation will be referred to as \emph{two-dimensional lexicographic} ($G$ generates a pair of values). The relations $\succ$ and  $\sim$ are then defined for the $a_1, a_2$ as follows: 
	\begin{itemize}
	\item $a_1 \succ a_2$ holds when $G_1(rep(a_1)) > G_1(rep(a_2))$ or ($G_1(rep(a_1)) = G_1(rep(a_2))$ and $G_2(rep(a_1)) > G_2(rep(a_2))$),
	\item $a_1 \sim a_2$ holds otherwise (i.e. when $G_1(rep(a_1)) = G_1(rep(a_2))$ and $G_2(rep(a_1)) = G_2(rep(a_2))$).
	\end{itemize}
\item Etc. (These can easily be extended to any $c > 2$).
\end{itemize}
Given $c \geq 2$, it is thus possible to define $c$-dimensional lexicographic aggregation. However, also a one-dimensional aggregation might be formally referred to as lexicographic (namely: \emph{one-dimensional lexicographic}) for the sake of completeness. 
Consequently, an $c$-dimensional lexicographic aggregation may be defined for any $c \geq 1$.
The distinctive characteristics of an $c$-dimensional lexicographic aggregation is that it generates an $c$-tuple of real values. Elements of this $c$-tuple are values that will be further referred to as components.

The `the aggregation mechanism' built into $c$-dimensional lexicographic aggregations is naturally of `lexicographic' character (it adheres strictly to the order of the components): 
\begin{itemize}
\item A one-dimensional aggregation (generates one component). Arguments assigned higher values of the component are preferred over arguments assigned lower values of the component. Otherwise, the arguments are indifferent. Shortly: the aggregation decides about the preference by acting with the component.
\item A two-dimensional aggregation (generates two components). Arguments assigned higher values of the first component or equal values of the 1-st component but higher values of the 2-nd component are preferred over arguments assigned lower values of the 1-st component or equal values of the 1-st component but lower values of the 2-nd component. Otherwise, the arguments are indifferent. Shortly: the aggregation decides about the preference by acting with the 1-st component first and, if necessary, with the 2-nd component next (i.e. only when the action of the first component is not conclusive).
\item Etc. (These can easily be extended to any $c > 2$).
\end{itemize}
Notice that without the `lexicographic' mechanism, instead of generating a single (aggregated) linear pre-order, every $c$-dimensional lexicographic aggregation would for $c > 1$ generate $c$ separate (individual) linear pre-orders corresponding to the individual components. These pre-orders would be identical for fully ordinally equivalent components and thus would in practice reduce the dimensionality of the aggregation to $p \in \{1, 2, ..., c-1\}$. In result, it is further assumed that no two components of such and aggregation are fully ordinally equivalent.

Of course, the aggregated linear pre-order implied by the lexicographic aggregations may be influenced by changing the order of components or by modifying the components themselves (which influences the individual linear pre-orders).
Let $G$ be an $n$-dimensional lexicographic aggregation that operates by aggregating elements of $\mathbb{A}$, $i \in \{1, 2, ..., c\}$ and let $f(x)$ be a real-valued, strictly increasing, while $h(x)$ a real-valued, strictly decreasing function of a real-valued argument $x \in C_i$, where $C_i$ is the counter-domain of component $G_i$. By definition, the linear pre-order implied by $G_i$:
\begin{itemize}
\item will not be changed, if $f(G_i(a))$ is generated instead of $G_i(a)$ for every $a \in \mathbb{A}$, 
\item will become reversed, if $h(G_i(a))$ is generated instead of $G_i(a)$ for every $a \in \mathbb{A}$. 
\end{itemize}
Additionally, the individual linear pre-order implied by component $G_i(a)$ will become trivial (all elements indifferent from one another), if any real-valued constant is generated instead of $G_i(a)$ for every $a \in \mathbb{A}$. If $c > 1$ such a trivialization of an individual component of the $c$-tuple practically reduces $G$ to $(c-1)$-dimensional lexicographic (because the constant component $G_i$ may be removed without influencing the aggregated linear pre-order). Analogous practical reduction occurs when two components are fully correlated. 

On the other hand, `unreducible' lexicographic aggregation of high dimensions will tend to imply many cases of preference and few cases of indifference. In result, such high-dimensional lexicographic aggregations will tend to imply linear orders ($\succ$) instead of linear pre-orders ($\succ \cup \sim$).

Needless to say, an $c$-dimensional lexicographic aggregation possesses also $c$ separate collections of isolines that will be identifiable by the components: the first component isolines, the second component isolines, etc. Notice that for $c > 1$ the entirety of isolines cannot be naturally visualized with colours, i.e. with a \emph{single} representation that uses a regular colour map. However, they may be visualized using $c$ separate representations (that use either different or identical colour maps) corresponding to successive components.

In the subsequent subsections, it will be naturally assumed that vectors from $VS$ represent alternatives from $\mathbb{A}$, i.e. if $a \in \mathbb{A}$, then $rep(a) \in VS$.

\subsection{One-dimensional lexicographic aggregations in TOPSIS}

All the aggregations considered up to this point, namely, the three classic ones: $\mathsf{I}$, $\mathsf{A}$ and  $\mathsf{R}$, as well as the three elliptic ones: $\mathsf{I}^\epsilon$, $\mathsf{A}^\epsilon$, $\mathsf{R}^\epsilon$, are all of the form $G: VS \rightarrow [0,1]$. Because $[0,1] \subset (-\infty,+\infty)$, these aggregations are all one-dimensional, which means that they may be referred to as one-dimensional lexicographic. Apart from implying a linear pre-order on the vectors from $VS$, the one-dimensionality allows to naturally render their values with colours (i.e. with regular colour maps) against the visualization of WMSD-space.

An inherent feature of the elliptic aggregations is that, depending on their parameter $\epsilon$, they may become reduced to the classic aggregations ($\epsilon = 1$), with the default influence of WM and WSD on the result, or they may start functioning differently from the classic aggregations, explicitly changing this influence of WM and WSD on the result. In the latter case they may either promote WSD over WM ($\epsilon < 1$) or promote WM over WSD ($\epsilon > 1$). 

As far as $\mathsf{I}^\epsilon$, $\mathsf{A}^\epsilon$, $\mathsf{R}^\epsilon$  are concerned, every increase of $\epsilon$ results in the relative influence of WM becoming higher and that of WSD becoming lower. Because $\epsilon$ is bounded from below in the case of $\mathsf{I}^\epsilon$ and $\mathsf{A}^\epsilon$, but in no case from above, by unrestrained increase of $\epsilon$ the influence of WSD in these three aggregations may be made arbitrarily low (though always non-zero). 
Furthermore, $\mathsf{I}^\epsilon$, $\mathsf{A}^\epsilon$, $\mathsf{R}^\epsilon$ may be theoretically applied with with $\epsilon = +\infty$, in which case all three aggregations stop being elliptic, and all reduce to a common, single aggregation, further referred to as $\mathsf{M}$ and defined as $\mathsf{M} = \text{WM}$. In this aggregation, the result is by definition influenced only by the value of WM (which also means that the influence of WSD is zero). Of course TOPSIS equipped with this aggregation behaves like the `utility-based methods' and thus loses its key feature. For this reason it should no longer be treated as a TOPSIS-like method.

\subsection{Two-dimensional lexicographic aggregations of TOPSIS}

Interestingly, it is possible to construct equivalents of $\mathsf{I}^\epsilon$, $\mathsf{A}^\epsilon$ and  $\mathsf{R}^\epsilon$, in which the influence of WSD is smaller than in $\mathsf{I}^\epsilon$, $\mathsf{A}^\epsilon$ or $\mathsf{R}^\epsilon$ for $\epsilon \rightarrow +\infty$, but as opposed to $\mathsf{M}$, still non-zero. In this sense they may be treated as being `positioned between' $\mathsf{I}^\epsilon$, $\mathsf{A}^\epsilon$ or $\mathsf{R}^\epsilon$ for $\epsilon \rightarrow +\infty$ and
$\mathsf{I}^\epsilon$, $\mathsf{A}^\epsilon$ or $\mathsf{R}^\epsilon$ for $\epsilon = +\infty$ (or $\mathsf{M}$), respectively. 
Three such equivalents, denoted as $\mathsf{I}^L$, $\mathsf{A}^L$ and $\mathsf{R}^L$, are defined as follows: 
\begin{itemize}
\item $\mathsf{I}^L = (\text{WM},-\text{WSD})$,
\item $\mathsf{A}^L = (\text{WM},+\text{WSD})$,
\item $\mathsf{R}^L = (\text{WM},z)$, where $z =$
\begin{itemize}[label=$\circ$]
\item[$\diamond$] $+\text{WSD}$ when $\text{WM} < \frac{mean(\mathbf{w})}{2}$,
\item[$\diamond$] 0 when $\text{WM} = \frac{mean(\mathbf{w})}{2}$,
\item[$\diamond$] $-\text{WSD}$ when $\text{WM} > \frac{mean(\mathbf{w})}{2}$.
\end{itemize}
\end{itemize}
All the equivalents generate pairs of values, which makes them two-dimensional lexicographic aggregations. Notice that because $\text{WM}$ and $\text{WSD}$ are `locally independent', the components of these aggregations are not ordinally equivalent, thus satisfying the corresponding assumptions. As opposed to the elliptic aggregations, however, the lexicographic ones admit no parameters. 

The `the ranking mechanism' built into $\mathsf{I}^L$, $\mathsf{A}^L$ and  $\mathsf{R}^L$ naturally follows the ordering of WM and WSD in the generated tuples, as well as their specific values. In particular:
\begin{itemize}
\item $\mathsf{I}^L$ decides about the preference/indifference by acting with the regular value of WM first and, if necessary, with the negated value of WSD next.
\item $\mathsf{A}^L$ decides about the preference/indifference by acting with the regular value of WM first and, if necessary, with the regular value of WSD next.
\item $\mathsf{R}^L$ decides about the preference/indifference by acting with the regular value of WM first and, next if necessary, when WM $\neq \frac{mean(\mathbf{w})}{2}$ with
\begin{itemize}
\item the regular value of WSD for WM $< \frac{mean(\mathbf{w})}{2}$,
\item the negated value of WSD for WM $> \frac{mean(\mathbf{w})}{2}$.
\end{itemize}
\end{itemize}
Notice that by definition aggregations $\mathsf{I}^L$, $\mathsf{A}^L$ and $\mathsf{R}^L$ treat WM and WSD exactly as their respective elliptic equivalents do.
Recall that the preference-related interplay of WM and WSD (Section~\ref{sec:TOPSIS_aggregations}) in the context of all the analyzed elliptic aggregations meant that the aggregations are always increasing with the increase of WM (under constant WSD). 
This is imitated by the lexicographic aggregations $\mathsf{I}^L$, $\mathsf{A}^L$ and $\mathsf{R}^L$ as they decide about the preference/indifference by acting with the value of WM first (i.e. higher values of WM imply higher values of the lexicographic aggregations).
In the context of WSD, when WM is constant the preference-related interplay came down to WSD being:
\begin{itemize}
    \item of type `cost' for the aggregation $\mathsf{I}^\epsilon$ and for $\mathsf{R}^\epsilon$ when $\text{WM} > \frac{mean(\mathbf{w})}{2}$,
    \item of type `gain' for the aggregation $\mathsf{A}^\epsilon$ and for $\mathsf{R}^\epsilon$ when $\text{WM} < \frac{mean(\mathbf{w})}{2}$,
    \item neutral for the aggregation $\mathsf{R}^\epsilon$ when $\text{WM} = \frac{mean(\mathbf{w})}{2}$.
\end{itemize}
In the same manner, when WSD is taken into account, $\mathsf{I}^L$ and $\mathsf{R}^L$ for $\text{WM} > \frac{mean(\mathbf{w})}{2}$ decides about the preference/indifference by acting with the negated values of WSD (i.e. the smaller WSD, the higher the aggregation value). Analogously, $\mathsf{A}^L$ and $\mathsf{R}^L$ for $\text{WM} < \frac{mean(\mathbf{w})}{2}$ decides about the preference/indifference by acting with the regular values of WSD (i.e. the higher WSD, the higher the aggregation value).
Finally, when $\text{WM} = \frac{mean(\mathbf{w})}{2}$, aggregation $\mathsf{R}^L$ decides by acting only with the regular value of WM.

Additionally, $\mathsf{I}^L$, $\mathsf{A}^L$ and  $\mathsf{R}^L$ share selected properties with all the other aggregations. In particular, they:
\begin{itemize}
\item may be expressed both with WM and WSD,
\item retain a non-zero influence of WSD on the result (the key feature of TOPSIS),
\item may be visualized against the WMSD-space (using two representations corresponding to the two components).
\end{itemize}
What differs $\mathsf{I}^L$, $\mathsf{A}^L$ and $\mathsf{R}^L$ from other aggregations is their `positioning between' $\mathsf{I}^\epsilon$, $\mathsf{A}^\epsilon$ and $\mathsf{R}^\epsilon$ for $\epsilon \rightarrow +\infty$, and $\mathsf{I}^\epsilon$, $\mathsf{A}^\epsilon$ and $\mathsf{R}^\epsilon$ for $\epsilon = +\infty$ (or $\mathsf{M}$), respectively. In terms of isolines, this results from the fact that when $\epsilon$ grows, the isolines of $\mathsf{I}^\epsilon$, $\mathsf{A}^\epsilon$ and $\mathsf{R}^\epsilon$  become more and more (although never fully) vertical. In result, the influence of WSD on the final ranking becomes smaller and smaller, but always non-zero. On the other hand, the isolines of $\mathsf{I}^\epsilon$, $\mathsf{A}^\epsilon$ and $\mathsf{R}^\epsilon$ for $\epsilon = +\infty$ are fully vertical, reducing all these aggregations to $\mathsf{M}$, which results in the influence of WSD on the result becoming exactly zero. As far as $\mathsf{I}^L$, $\mathsf{A}^L$ and $\mathsf{R}^L$ are concerned, their first component isolines (i.e. the isolines of WM) are fully vertical, while their second component isolines (i.e. the isolines of WSD) are fully horizontal. Thanks to the existence of the second (non-constant) component, $\mathsf{I}^L$, $\mathsf{A}^L$ and $\mathsf{R}^L$ do not reduce to $\mathsf{M}$ and thus manifest a non-zero influence of WSD on the result. This means that TOPSIS equipped with these three aggregations does not lose its key feature and may still be treated as a TOPSIS-like method.

Even though the two-dimensional lexicographic aggregations defined above are all non-parametric, it is possible to define parametric two-dimensional lexicographic aggregations. 

The first example here is a parameterized variant of $\mathsf{R}^L$, denoted as $\mathsf{R}^{L_{\pm p}}$, where $p \in \{-1, +1\}$: $\mathsf{R}^{L_{\pm p}} = (\text{WM},z)$, where $z =$
\begin{itemize}
\item $-p \cdot \text{WSD}$ when $\text{WM} < \frac{mean(\mathbf{w})}{2}$,
\item 0 when $\text{WM} = \frac{mean(\mathbf{w})}{2}$,
\item $+p \cdot \text{WSD}$ when $\text{WM} > \frac{mean(\mathbf{w})}{2}$.
\end{itemize}
For $p = +1$ aggregation $\mathsf{R}^{L_{\pm p}}$ reduces to $\mathsf{R}^L$. For $p = -1$ it acts as $\mathsf{R}^L$ but with the role of $\text{WSD}$ `reversed'. In result, $\mathsf{R}^{L_{\pm p}}$ decides about the preference by acting with the regular value of WM first and, if necessary, when WM $\neq \frac{mean(\mathbf{w})}{2}$ with
\begin{itemize}
\item the negated value of WSD for WM $< \frac{mean(\mathbf{w})}{2}$,
\item the regular value of WSD for WM $> \frac{mean(\mathbf{w})}{2}$
\end{itemize}
next.

The second example is a parameterized two-dimensional lexicographic aggregation, denoted as $\mathsf{X}_\mathbf{w}^{L_{\pm p}}$, where $p \in \{-1, +1\}$, in which the order of components is exchanged:
\begin{itemize}
\item $\mathsf{X}_\mathbf{w}^{L_{\pm p}} =$
    \begin{itemize}
    \item $(\mathsf{I}^\epsilon,\mathsf{A}^\epsilon)$ for $p = -1$,
    \item $(\mathsf{A}^\epsilon,\mathsf{I}^\epsilon)$ for $p = +1$.
    \end{itemize}
\end{itemize}
This aggregation basically imitates aggregations $\mathsf{I}^\epsilon$ and $\mathsf{A}^\epsilon$ for $p = -1$ and $p = +1$, respectively, but (owing to its second component) may generate fewer cases of indifference than the imitated aggregations. This is especially useful in situations in which abundant cases of indifference are implied by $\mathsf{I}^\epsilon$ and $\mathsf{A}^\epsilon$.

Of course it is not only possible to define parameterized two-dimensional lexicographic aggregations, but also other, e.g. more-dimensional lexicographic aggregations. As already stated, such more-dimensional aggregations are useful against abundant cases of indifference. An example of a three-dimensional lexicographic aggregations is: 
$\mathsf{R}_\mathbf{w}^{L_3} = (\text{WM},max(\mathbf{v}),-min(\mathbf{v}))$.
This aggregation clearly imitates $\mathsf{M}$. However, if necessary, it acts with the maximal value of $\mathbf{v}$. This means that e.g. out of two alternatives characterized in $VS$ by two vectors that are different but have the same WM, the aggregation will rate higher this alternative whose maximal value is better. Finally, if still necessary, it acts with the negated minimal value of $\mathbf{v}$.

\section{Case Study}
\label{sec:Case-Study}
To show the usability of the proposed generalizations of TOPSIS and illustrate their practical implications, let us consider three simple case studies based on the real-world dataset used in~\cite{GSZ13,ZIE17,Susmaga_2023MSD}. 
The original data describes the technical condition of 32 buses, however, for the sake of brevity, in this paper we shall focus only on a subset of ten buses (Table~\ref{tab:exemplary-alternatives-buses-CS}). The denotation of the alternatives is kept as in the full dataset in~\cite{Susmaga_2023MSD} to facilitate comparison with the other papers. Each alternative is described by eight numeric criteria referring to its technical condition. All the criteria are assumed to be equally important ($\mathbf{w} = \mathbf{1} = [1, 1, 1, 1, 1, 1, 1, 1]$), with four of them being of type `gain' (`Speed', `Pressure', `Torque', `Horsepower') and four of type `cost' (`Blacking in exhaust gas', `Summer/Winter fuel consumption', `Oil consumption'). Notice that all the chosen alternatives happen to have identical descriptions in terms of `Pressure', which also means that this particular criterion (despite its non-zero weight) has no influence on the generated rankings.
For more detailed description of the dataset see~\cite{Susmaga_2023MSD}.

\begin{figure*}[!htb]
\centering
\includegraphics[trim = 0mm 25mm 0mm 22mm, clip, width=0.95\textwidth]  {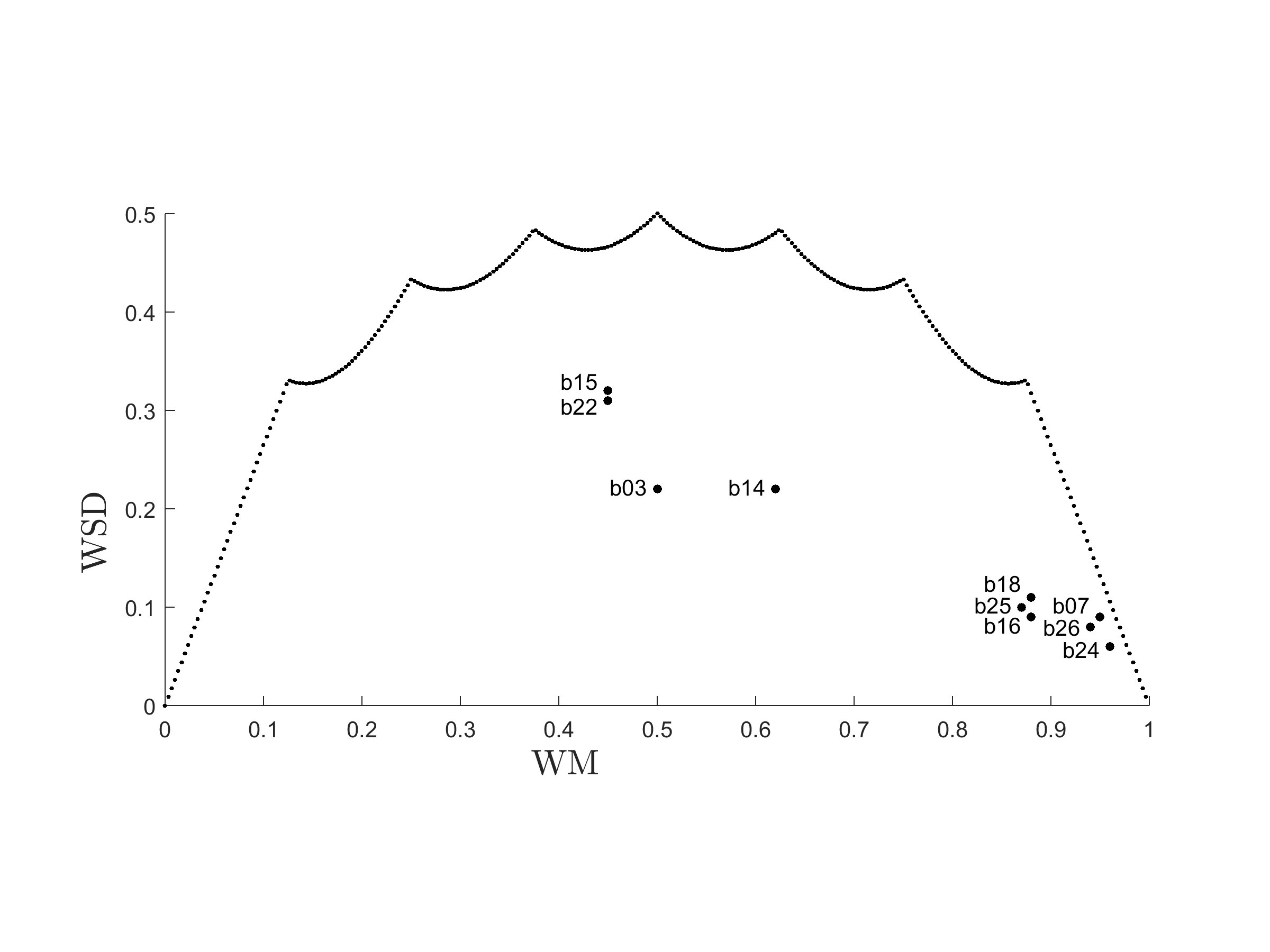}
\caption{The WMSD-space defined by $\mathbf{w} = \mathbf{1}$ with points representing the alternatives.}
\label{fig:figure-1}
\end{figure*}

Table~\ref{tab:exemplary-alternatives-buses-CS} presents the description of the chosen alternatives in terms of criteria (i.e. elements of $CS$), whereas Table~\ref{tab:exemplary-alternatives-buses-VS} exhibits them in terms of weighted utilities (i.e. elements of $VS$), as well as in terms of WM and WSD (i.e. elements of WMSD-space). 
Notice that $US = VS$ owing to $\mathbf{w} = \mathbf{1}$, so Table~\ref{tab:exemplary-alternatives-buses-VS} actually presents alternatives' representations both in the utility space and in its weighted counterpart. 
The WM and WSD values are used to depict the alternatives as points in the WMSD-space (Fig.~\ref{fig:figure-1}).
The shape of the WMSD-space as well as the position of the points (alternatives) within the space 
will not change under different aggregations considered in the subsequent subsection. It is due to the fact that these aspects are dependent on the particular weights of the criteria and the particular descriptions of alternatives, but independent of the aggregations used.
\begin{table*}[!ht]
\footnotesize
  \centering
  \caption{Description of chosen alternatives in terms of criteria (elements of $CS$)}
    \begin{tabular}{@{}l|cccccccr@{}}
\toprule
\qquad & \multicolumn{8}{c}{Specifications in $CS$} \\
Bus & Speed & Pressure & Blacking & Torque & Summer & Winter & Oil & HP 
    \\ \midrule 
$\mathbf{b}_{03}$ & 72 & 2 & 73 & 425 & 23 & 27 & 2 & 112\\
$\mathbf{b}_{07}$ & 90 & 2 & 26 & 482 & 22 & 24 & 0 & 148\\
$\mathbf{b}_{14}$ & 75 & 2 & 64 & 432 & 22 & 25 & 1 & 114\\
$\mathbf{b}_{15}$ & 68 & 2 & 70 & 400 & 22 & 26 & 2 & 100\\
$\mathbf{b}_{16}$ & 88 & 2 & 44 & 478 & 21 & 25 & 0 & 138\\
$\mathbf{b}_{18}$ & 90 & 2 & 40 & 480 & 22 & 25 & 0 & 139\\
$\mathbf{b}_{22}$ & 68 & 2 & 88 & 422 & 22 & 25 & 3 & 108\\
$\mathbf{b}_{24}$ & 90 & 2 & 38 & 482 & 20 & 24 & 0 & 146\\
$\mathbf{b}_{25}$ & 90 & 2 & 45 & 479 & 21 & 25 & 1 & 145\\
$\mathbf{b}_{26}$ & 90 & 2 & 34 & 486 & 21 & 25 & 0 & 148\\
\bottomrule
    \end{tabular}%
\label{tab:exemplary-alternatives-buses-CS}
\end{table*}%

\begin{table*}[htb]
\footnotesize
  \centering
  \caption{Description of chosen alternatives in terms of weighted utilities (elements of $VS$) and in terms of WM and WSD (elements of WMSD-space)}
    \begin{tabular}{@{}l|cccccccc|rr@{}}
\toprule
\qquad & \multicolumn{8}{c}{Specifications in $VS$}  & \multicolumn{2}{c}{WMSD-space}\\
Bus & Speed & Pressure & Blacking & Torque & Summer & Winter & Oil & HP & WM & WSD
    \\ \midrule 
$\mathbf{b}_{03}$ & 0.40 & 1.00 & 0.32 & 0.29 & 0.57 & 0.60 & 0.50 & 0.31& 0.50 & 0.22 \\
$\mathbf{b}_{07}$ & 1.00 & 1.00 & 1.00 & 0.95 & 0.71 & 0.90 & 1.00 & 1.00& 0.95 & 0.09 \\
$\mathbf{b}_{14}$ & 0.50 & 1.00 & 0.45 & 0.37 & 0.71 & 0.80 & 0.75 & 0.35& 0.62 & 0.22 \\
$\mathbf{b}_{15}$ & 0.27 & 1.00 & 0.36 & 0.00 & 0.71 & 0.70 & 0.50 & 0.08& 0.45 & 0.32 \\
$\mathbf{b}_{16}$ & 0.93 & 1.00 & 0.74 & 0.91 & 0.86 & 0.80 & 1.00 & 0.81& 0.88 & 0.09 \\
$\mathbf{b}_{18}$ & 1.00 & 1.00 & 0.80 & 0.93 & 0.71 & 0.80 & 1.00 & 0.83& 0.88 & 0.11 \\
$\mathbf{b}_{22}$ & 0.27 & 1.00 & 0.10 & 0.26 & 0.71 & 0.80 & 0.25 & 0.23& 0.45 & 0.31 \\
$\mathbf{b}_{24}$ & 1.00 & 1.00 & 0.83 & 0.95 & 1.00 & 0.90 & 1.00 & 0.96& 0.96 & 0.06 \\
$\mathbf{b}_{25}$ & 1.00 & 1.00 & 0.72 & 0.92 & 0.86 & 0.80 & 0.75 & 0.94& 0.87 & 0.10 \\
$\mathbf{b}_{26}$ & 1.00 & 1.00 & 0.88 & 1.00 & 0.86 & 0.80 & 1.00 & 1.00& 0.94 & 0.08 \\
\bottomrule
    \end{tabular}%
\label{tab:exemplary-alternatives-buses-VS}
\end{table*}%

\subsection{Classic and Elliptic instances of $\mathsf{R}$}

\begin{figure*}[!htb]
\centering
\begin{tabular}{cc}
\multicolumn{1}{l}{\quad A} & \multicolumn{1}{l}{\quad B}\\
\includegraphics[trim = 0mm 20mm 0mm 10mm, clip, width=0.45\textwidth]  {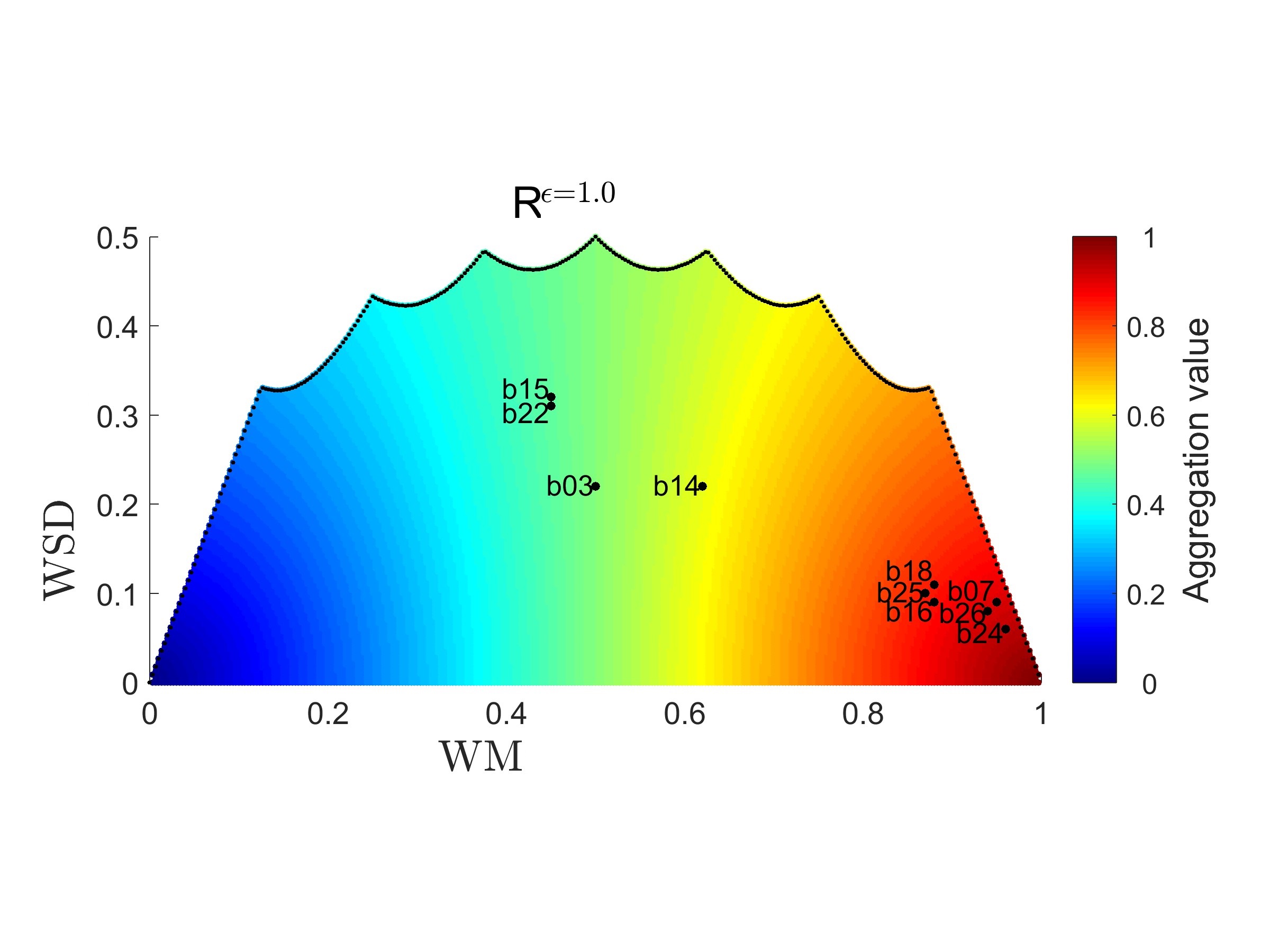} &
\includegraphics[trim = 0mm 20mm 0mm 10mm, clip, width=0.45\textwidth]  {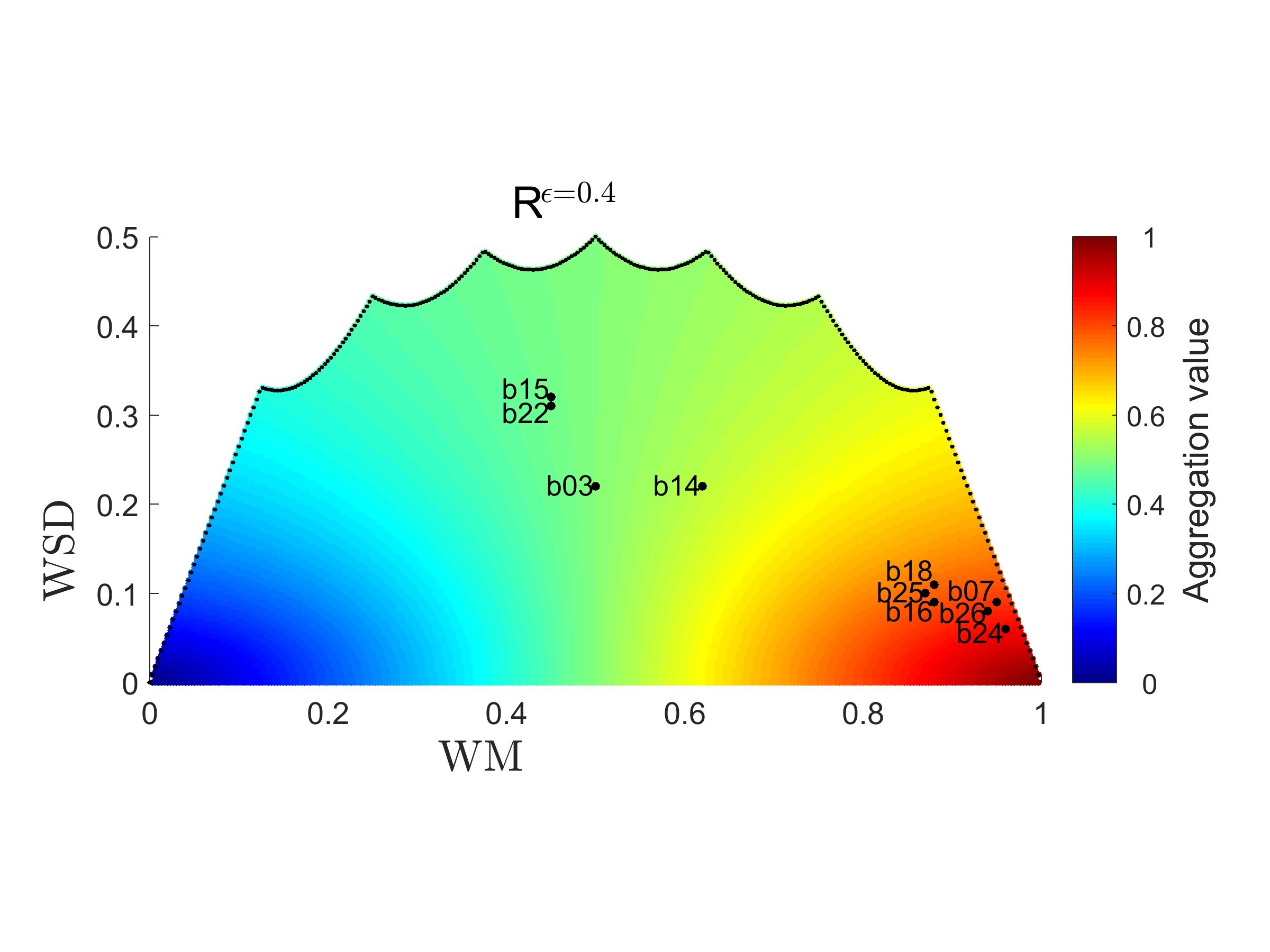}\\
\multicolumn{1}{l}{\quad C} & \multicolumn{1}{l}{\quad D}\\
\includegraphics[trim = 0mm 20mm 0mm 10mm, clip, width=0.45\textwidth]  {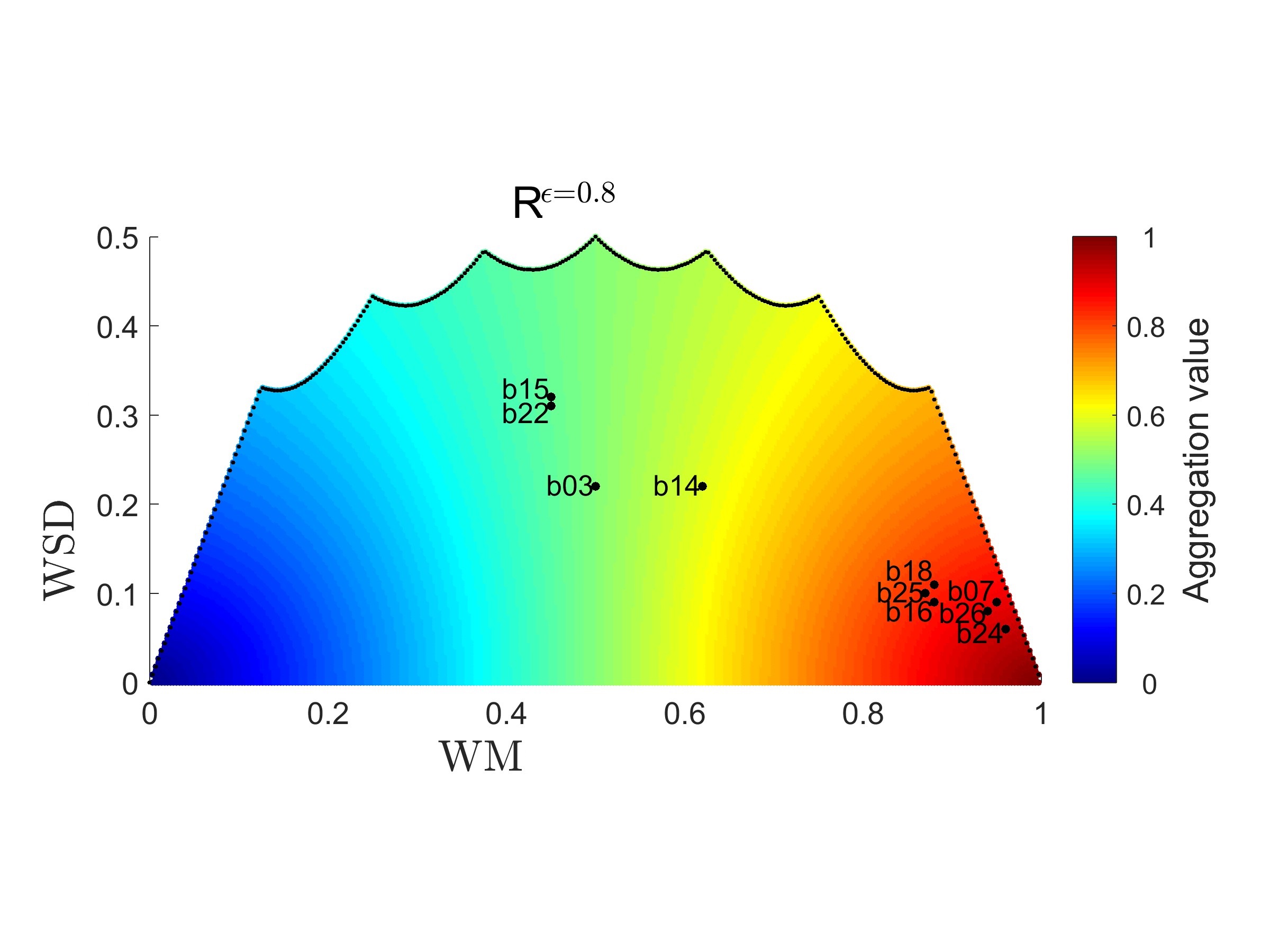} &
\includegraphics[trim = 0mm 20mm 0mm 10mm, clip, width=0.45\textwidth]  {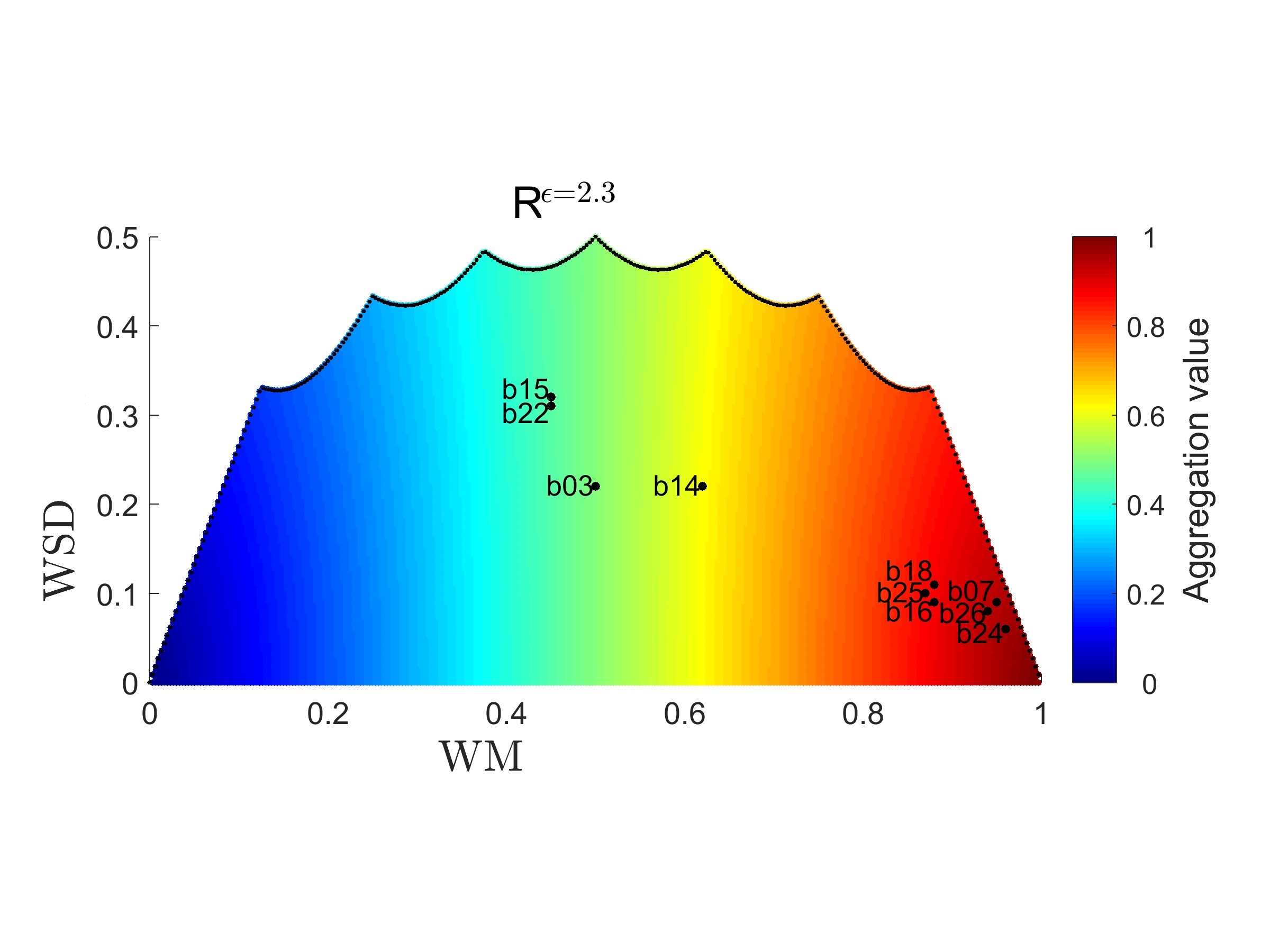}\\
\end{tabular}

\caption{The WMSD-space defined by $\mathbf{w} = \mathbf{1}$ with points representing the alternatives\\ against elliptic aggregation $\mathsf{R}^\epsilon$ for: (A) $\epsilon = 1$ -- equivalent to circular aggregation $\mathsf{R}$, (B) and (C) $\epsilon = 0.4$ and $\epsilon = 0.8$ -- aggregation promoting WSD over WM, (D) $\epsilon = 2.3$ -- aggregation promoting WM over WSD.}
\label{fig:figure-3-R}
\end{figure*}

\begin{figure*}[!htb]
\centering
\includegraphics[trim = 0mm 30mm 0mm 22mm, clip, width=0.65\textwidth]  {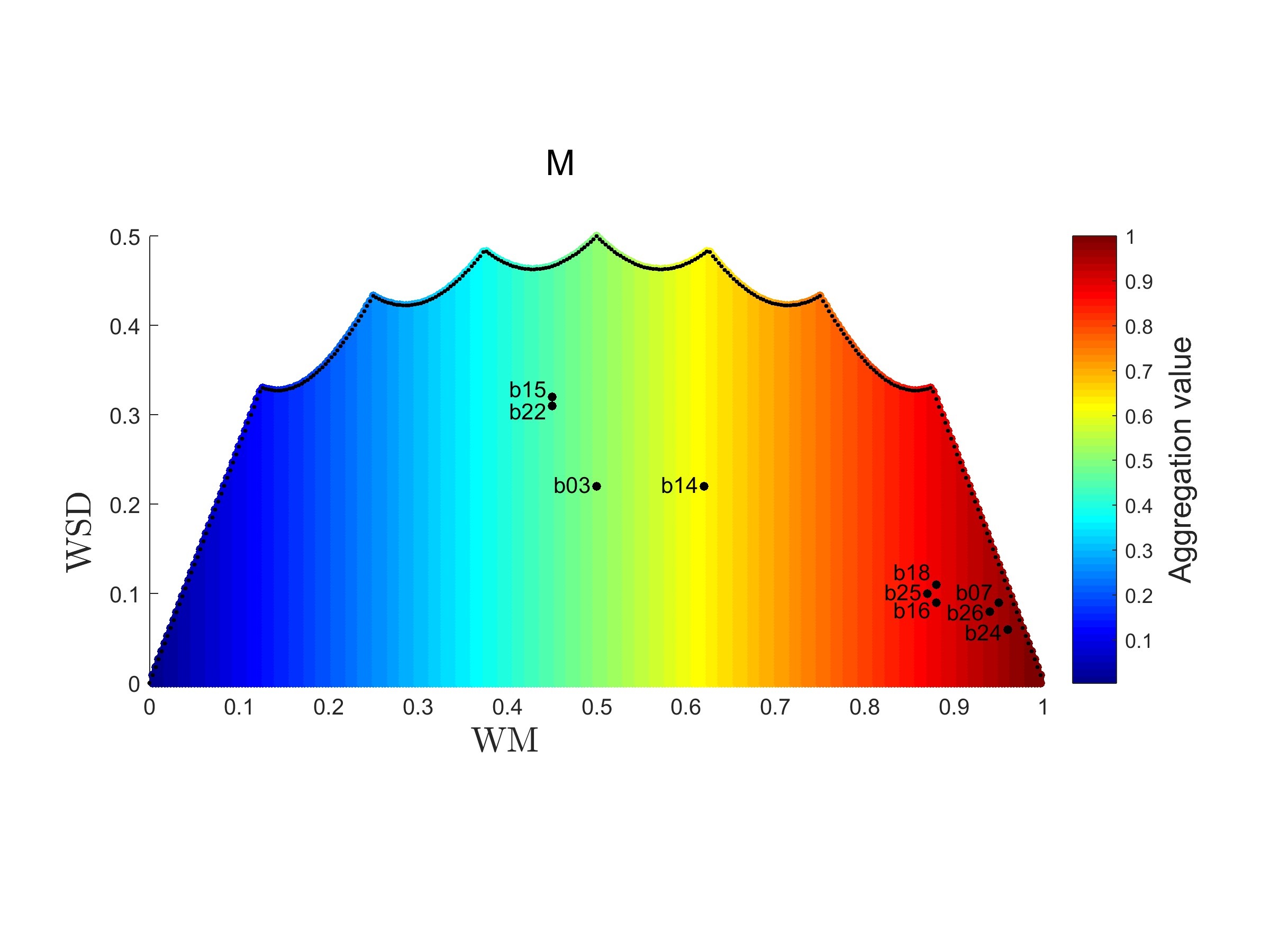}

\caption{The WMSD-space defined by $\mathbf{w} = \mathbf{1}$ with points representing the alternatives\\ against aggregation $\mathsf{M}$  (i.e. aggregation fully independent of WSD).}
\label{fig:figure-5-R-1.00}
\end{figure*}

In the conducted case study the classic TOPSIS aggregation $\mathsf{R}$, with `circular' isolines (equivalent to $\mathsf{R}^{\epsilon=1}$, i.e. an elliptic aggregation characterized by $\epsilon=1$; see Fig.~\ref{fig:figure-3-R}A), was compared with:
\begin{itemize}
    \item two elliptic aggregations promoting WSD over WM, i.e. characterized by $\epsilon = 0.4 < 1$ and $\epsilon = 0.8 < 1$ (see Figs.~\ref{fig:figure-3-R}B~and~C),
    \item an elliptic aggregation promoting WM over WSD, i.e. characterized by $\epsilon = 2.3 > 1$ (see Fig.~\ref{fig:figure-3-R}D),
    \item the $\mathsf{M}$ aggregation, in which only WM is taken into account, characterized by the `in limit' situation: $\epsilon = +\infty$ (see Fig.~\ref{fig:figure-5-R-1.00}).
\end{itemize}
It should be noted that as far as $\mathsf{R}^\epsilon$ is concerned, the value of $\epsilon$ is by no means limited, so it could have taken any value from $(0, +\infty)$.

Table~\ref{tab:exemplary-alternatives-buses-WMSD-R_aggregs} provides the representation of the considered alternatives in terms of WM and WMSD first, and in terms of the values of five compared aggregations next. Additionally, the ranking position of an alternative under particular aggregation is given as the bracketed upper index. 

\begin{table*}[htb]
\footnotesize
  \centering
  \caption{Description of chosen alternatives in terms of WM and WSD, and in terms of five $\mathsf{R}^\epsilon$ aggregations; bracketed upper indices show their positions in the ranking.}
    \begin{tabular}{@{}l|cc|ccccc@{}}
\toprule
\qquad & \multicolumn{2}{c}{WMSD-space} & \multicolumn{5}{c}{Aggregations} \\
Bus & WM & WSD 
		  &$\mathsf{R}^{\epsilon=1}$
        &$\mathsf{R}^{\epsilon=0.4}$
        &$\mathsf{R}^{\epsilon=0.8}$
        &$\mathsf{R}^{\epsilon=2.3}$
        &$\mathsf{R}^{\epsilon=\infty} = \mathsf{M}$
    \\ \midrule 
$\mathbf{b}_{03}$ & $0.50$ & $0.22$ & $0.500^{(8)}$ & $0.500^{(8)}$ & $0.500^{(7)}$ & $0.500^{(7)}$ & $0.500^{(7)}$ \\ 
$\mathbf{b}_{07}$ & $0.95$ & $0.09$ & $0.903^{(3)}$ & $0.818^{(3)}$ & $0.886^{(3)}$ & $0.932^{(2)}$ & $0.950^{(2)}$ \\ 
$\mathbf{b}_{14}$ & $0.62$ & $0.22$ & $0.600^{(7)}$ & $0.558^{(7)}$ & $0.591^{(6)}$ & $0.616^{(6)}$ & $0.620^{(6)}$ \\ 
$\mathbf{b}_{15}$ & $0.45$ & $0.32$ & $0.465^{(9)}$ & $0.485^{(9)}$ & $0.470^{(8)}$ & $0.454^{(8)}$ & $0.450^{(8)}$ \\ 
$\mathbf{b}_{16}$ & $0.88$ & $0.09$ & $0.855^{(4)}$ & $0.789^{(4)}$ & $0.844^{(4)}$ & $0.875^{(3)}$ & $0.880^{(4)}$ \\ 
$\mathbf{b}_{18}$ & $0.88$ & $0.11$ & $0.845^{(5)}$ & $0.764^{(6)}$ & $0.830^{(5)}$ & $0.872^{(4)}$ & $0.880^{(4)}$ \\ 
$\mathbf{b}_{22}$ & $0.45$ & $0.31$ & $0.464^{(10)}$ & $0.484^{(10)}$ & $0.469^{(9)}$ & $0.453^{(9)}$ & $0.450^{(8)}$ \\ 
$\mathbf{b}_{24}$ & $0.96$ & $0.06$ & $0.930^{(1)}$ & $0.870^{(1)}$ & $0.919^{(1)}$ & $0.953^{(1)}$ & $0.960^{(1)}$ \\ 
$\mathbf{b}_{25}$ & $0.87$ & $0.10$ & $0.842^{(6)}$ & $0.771^{(5)}$ & $0.830^{(5)}$ & $0.864^{(5)}$ & $0.870^{(5)}$ \\ 
$\mathbf{b}_{26}$ & $0.94$ & $0.08$ & $0.904^{(2)}$ & $0.830^{(2)}$ & $0.890^{(2)}$ & $0.932^{(2)}$ & $0.940^{(3)}$ \\ 
\bottomrule
    \end{tabular}%
\label{tab:exemplary-alternatives-buses-WMSD-R_aggregs}
\end{table*}%

\noindent
The $\epsilon$ parameter influences the aggregations or, more precisely, shape of their isolines, changing them from `circular' ones (Fig.~\ref{fig:figure-3-R}A) to horizontally elongated `elliptical' ones (Figs.~\ref{fig:figure-3-R}B and \ref{fig:figure-3-R}C) on one hand, or from `circular' ones (Fig.~\ref{fig:figure-3-R}A) through vertically elongated `elliptical' ones (Fig.~\ref{fig:figure-3-R}D) all the way to straight vertical lines (Fig.~\ref{fig:figure-5-R-1.00}), on the other. 
Because the final rating of an alternative is directly influenced by the shape of these isolines, the $\epsilon$ parameter has a direct impact on the position of the alternative in the final ranking.

Let us now have a closer look at how the ratings of the considered alternatives react to the different versions of the considered aggregation.
First of all, $\mathbf{b}_{24}$ is the undisputed winner under all four considered aggregations, as it has unrivaled values of WM (almost maximum) and WSD (low enough).

Next, consider alternatives $\mathbf{b}_{07}$ and $\mathbf{b}_{26}$, that lie close to each other within the WMSD-space. This is clear after the weighted utilities (Table~\ref{tab:exemplary-alternatives-buses-VS}) of both alternatives are compared, because they are fairly similar. The small differences lead to $\mathbf{b}_{26}$ having slightly smaller WM, but slightly larger WSD (this results from the fact that the weighted utilities of $\mathbf{b}_{26}$ are more `dispersed' than those of $\mathbf{b}_{07}$). As such, $\mathbf{b}_{26}$ would be ranked below $\mathbf{b}_{07}$ by all the `utility-based methods' (which take only WM into account).
This, however, is not the case with TOPSIS, which does not rate alternatives by their WM, but by their distances to the ideal and anti-ideal points, making the rating dependent not only on WM, but also on WSD (the `key feature' of TOPSIS).
In result, TOPSIS in its classic version (i.e. under aggregation $\mathsf{R}$) ranks $\mathbf{b}_{26}$ higher than $\mathbf{b}_{07}$. The generalized version of TOPSIS (i.e. under aggregation $\mathsf{R}^\epsilon$) described in this paper allows to control the influence of WM and WSD, effectively shifting the behaviour of TOPSIS away or towards that of the `utility-based methods' (with full agreement achieved under $\mathsf{R}^{\epsilon}$ for $\epsilon \rightarrow +\infty$).

As already stated, in the case of $\mathsf{R}$ (circular aggregation) 
$\mathbf{b}_{26}$ is rated higher than $\mathbf{b}_{07}$ (resulting in the higher rank of $\mathbf{b}_{26}$). This advantage of $\mathbf{b}_{26}$ clearly intensifies in the case of $\mathsf{R}^{\epsilon=0.4}$ and $\mathsf{R}^{\epsilon=0.8}$ (elliptic aggregations promoting WSD), in which the influence of WSD on the result is further increased.
Of course, the situation gradually reverses with aggregations promoting WM over WSD. In particular, in the case of $\mathsf{R}^{\epsilon=2.3}$ (elliptic aggregation, promoting WM) $\mathbf{b}_{07}$ and $\mathbf{b}_{26}$ are ranked equal, while in the case of $\mathsf{M}$ (WM only) it is $\mathbf{b}_{07}$ that is higher in the ranking than $\mathbf{b}_{26}$. 

A very similar situation concerns alternatives $\mathbf{b}_{18}$ and $\mathbf{b}_{25}$, which differ in both WM and WSD by 0.01 (exactly the same holds for $\mathbf{b}_{07}$ and $\mathbf{b}_{26}$), but the considered aggregations rank alternatives $\mathbf{b}_{18}$ and $\mathbf{b}_{25}$ differently than alternatives $\mathbf{b}_{07}$ and $\mathbf{b}_{26}$: $\mathbf{b}_{18}$ is ranked lower than $\mathbf{b}_{25}$ by $\mathsf{R}^{\epsilon=0.4}$, both are ranked the same by $\mathsf{R}^{\epsilon=0.8}$, and finally, $\mathbf{b}_{18}$ is ranked higher by $\mathsf{R}^{\epsilon=1.0}$, $\mathsf{R}^{\epsilon=2.3}$ and $\mathsf{M}$.
This is due to the fact that the influence of WM and WSD differs in different regions of WMSD-space and pair $\mathbf{b}_{18}$ and $\mathbf{b}_{25}$ is located in a different region of WMSD-space than pair $\mathbf{b}_{07}$ and $\mathbf{b}_{26}$.

Differences in how the aggregations rank alternatives occur also when the values of WM or the values of WSD of these alternatives are identical. This again emphasizes the difference between  classic TOPSIS and the `utility-based methods' (as well as the generalized TOPSIS, although in generalized TOPSIS the difference may be arbitrarily diminished).
Consider alternatives $\mathbf{b}_{16}$ and $\mathbf{b}_{18}$, which are characterized by the same $WM=0.88$ and different WSD. 
Since $\text{WM}=0.88>\frac{mean(\mathbf{w})}{2}=0.5$, the higher the value of WSD, the lower the ranking of an alternative under $\mathsf{R}^{\epsilon=1}$ (recall the preference-related-interplay between WM and WSD in Section~\ref{sec:TOPSIS_aggregations} and in~\cite{susmaga2023WMSD}). Thus, $\mathbf{b}_{16}$ is ranked higher than $\mathbf{b}_{18}$. Furthermore, the fact that the alternatives have the same WM also implies that there is no such $\epsilon$ that could place $\mathbf{b}_{18}$ higher in the ranking than $\mathbf{b}_{16}$. Nonetheless, $\mathbf{b}_{18}$ and $\mathbf{b}_{16}$ can be ranked equally under $\mathsf{M}$, as it is independent of WSD and ranks according to WM only.

Analogous consideration can be made for alternatives
$\mathbf{b}_{15}$ and $\mathbf{b}_{22}$, which are characterized by $\text{WM}=0.45<0.5$. In this case, however, the higher the WSD, the higher the ranking position under $\mathsf{R}^{\epsilon=1}$. Thus, $\mathbf{b}_{15}$ is ranked higher than $\mathbf{b}_{22}$. Choosing any $\epsilon$ cannot reverse this ranking. However, for $\epsilon = \infty$ (i.e. for $\mathsf{M}$), the two alternatives can be ranked equally, which constitutes a result consistent with that of the `utility-based methods'.

Now, let us consider two alternatives that have the same WSD, but different WM: $\mathbf{b}_{03}$ and $\mathbf{b}_{14}$. As shown in~\cite{susmaga2023WMSD}, the higher the WM, the better ranking position under the $\mathsf{R}$. Since $\mathbf{b}_{14}$ has higher WM than $\mathbf{b}_{03}$, it is higher in the ranking. This does not change under different considered values of $\epsilon$, 
see Figs.~\ref{fig:figure-3-R} and \ref{fig:figure-5-R-1.00}, and again constitutes a ranking consistent with that of the `utility-based methods'.

Finally, consider alternative $\mathbf{b}_{03}$, which lies in the middle of the WSD-space, on the vertical isoline of $\mathsf{R}$ and of $\mathsf{R}^{\epsilon}$ (Fig.~\ref{fig:figure-3-R}). This isoline is the only one that does not change its shape under any change to $\epsilon$ in $\mathsf{R}^{\epsilon}$ and always remains vertical. In the case of all alternatives lying in the middle of the WMSD-space, TOPSIS (both classic, as well as elliptic) behaves exactly as the `utility-based methods', ranking them always as equal.

\subsection{Classic and Elliptic instances of $\mathsf{I}$ and $\mathsf{A}$}

\begin{figure*}[!htb]
\centering
\begin{tabular}{cc}
\multicolumn{1}{l}{\quad A} & \multicolumn{1}{l}{\quad B}\\
\includegraphics[trim = 0mm 20mm 0mm 10mm, clip, width=0.475\textwidth]  {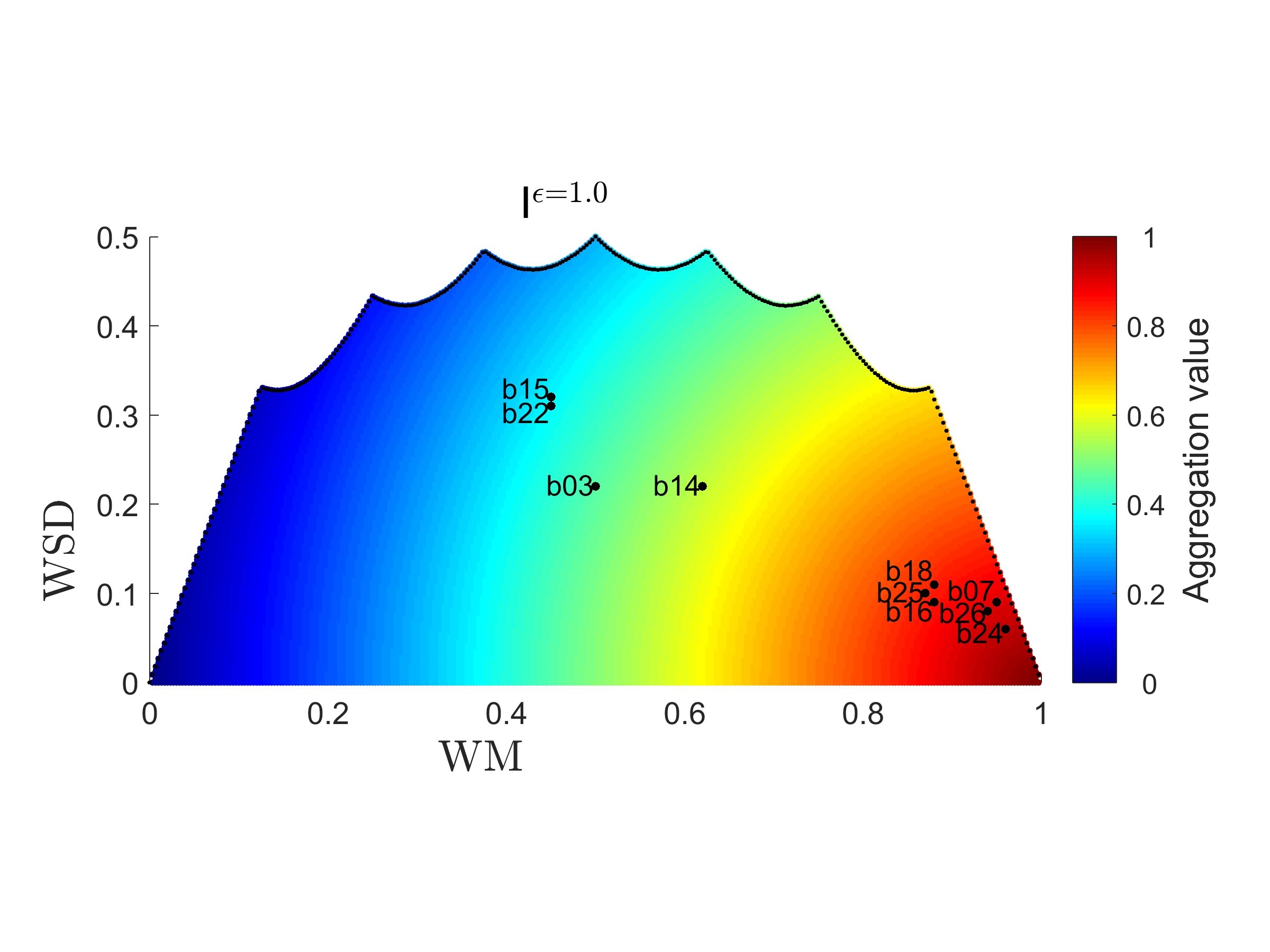} &
\includegraphics[trim = 0mm 20mm 0mm 10mm, clip, width=0.475\textwidth]  {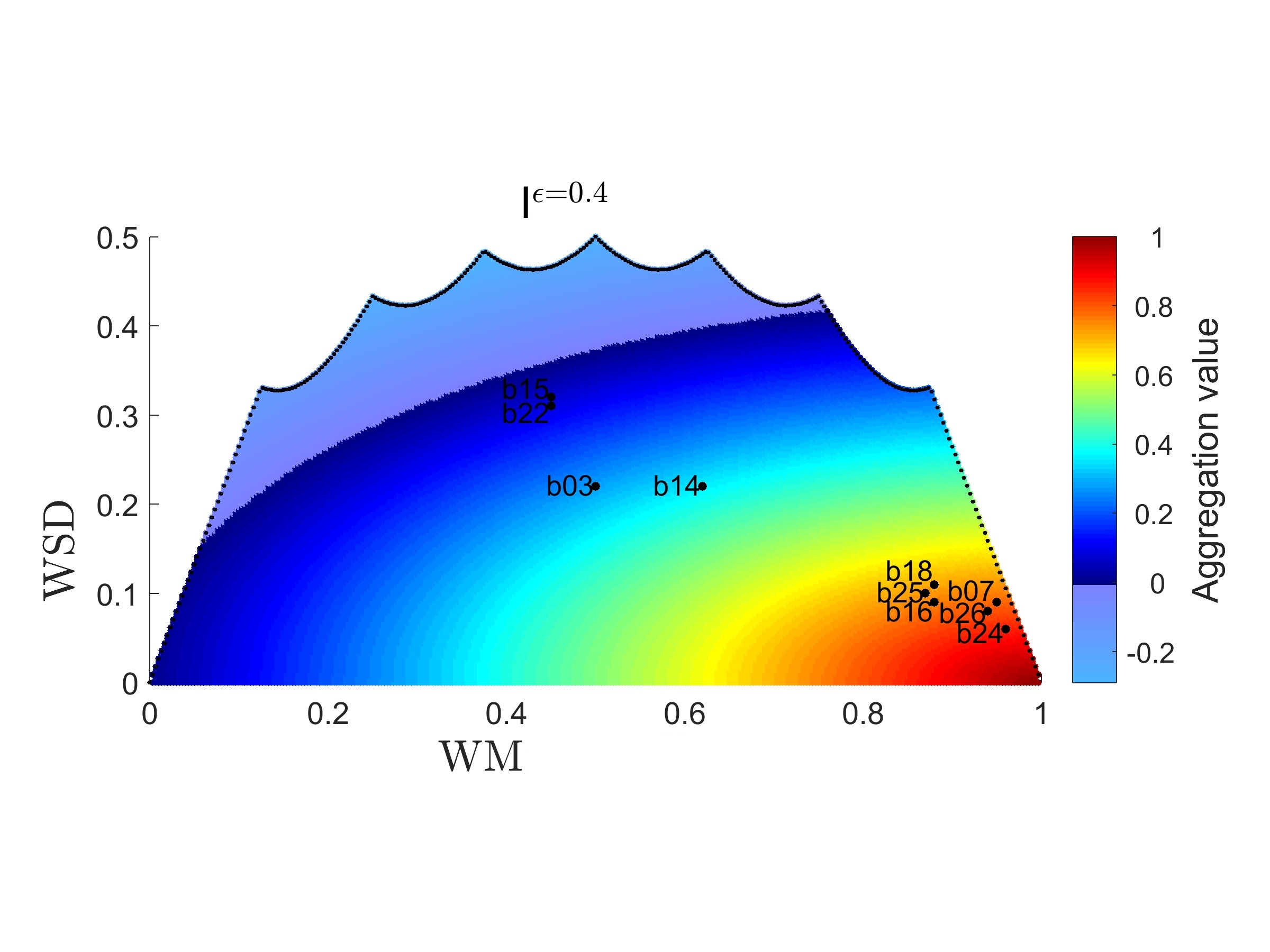}\\
\multicolumn{1}{l}{\quad C} & \multicolumn{1}{l}{\quad D}\\
\includegraphics[trim = 0mm 20mm 0mm 10mm, clip, width=0.475\textwidth]  {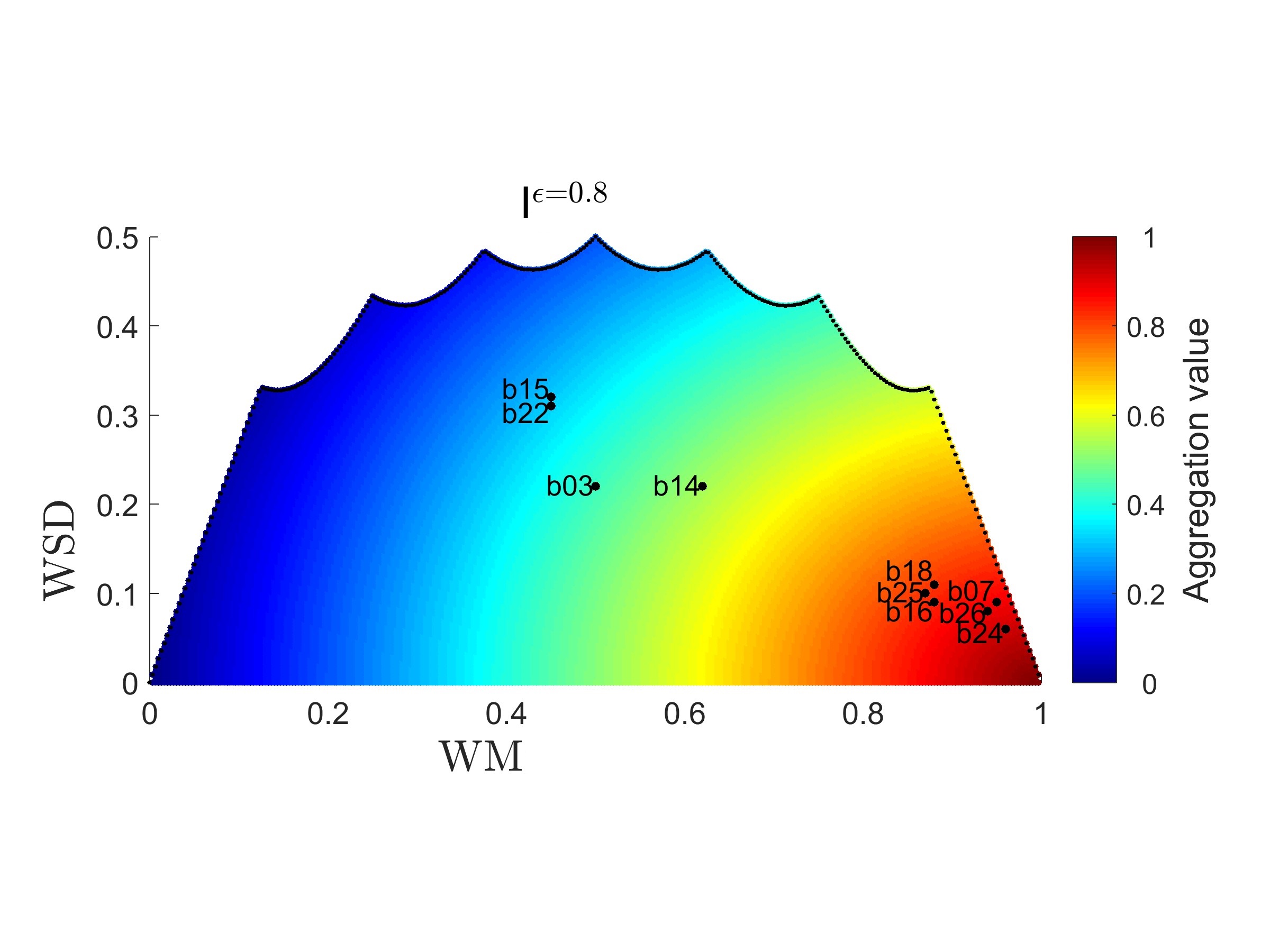} &
\includegraphics[trim = 0mm 20mm 0mm 10mm, clip, width=0.475\textwidth]  {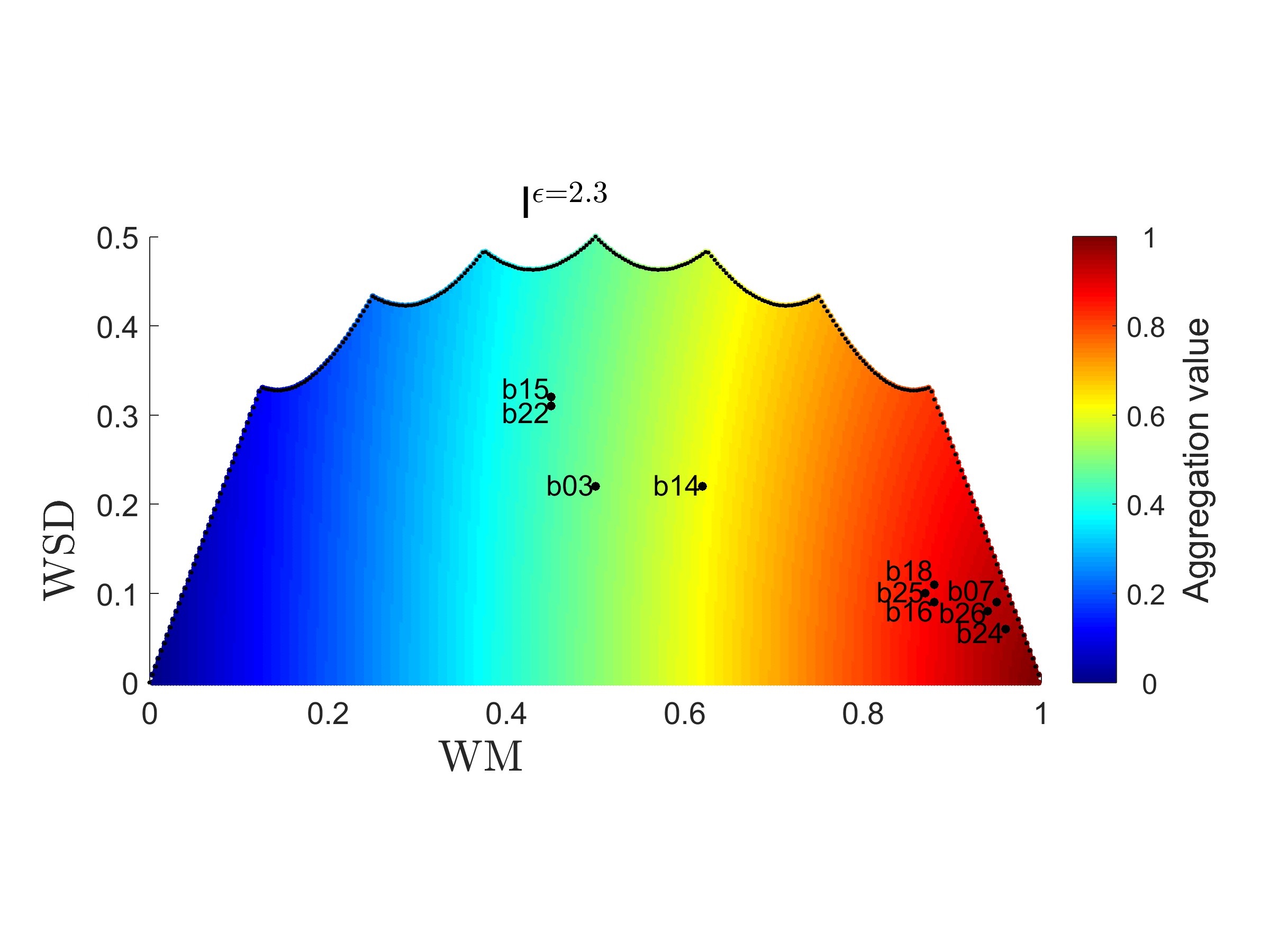}\\
\end{tabular}

\caption{The WMSD-space defined by $\mathbf{w} = \mathbf{1}$ with points representing the alternatives\\ against elliptic aggregation $\mathsf{I}^\epsilon$ for: (A) $\epsilon = 1$ -- equivalent to circular aggregation $\mathsf{I}$, (B) and (C) $\epsilon = 0.4$ and $\epsilon = 0.8$ -- aggregation promoting WSD over WM, (D) $\epsilon = 2.3$ -- aggregation promoting WM over WSD.}
\label{fig:figure-3-I}
\end{figure*}

In this case study the aggregation $\mathsf{I}$, with circular isolines (equivalent to $\mathsf{I}^{\epsilon=1} $ -- elliptic aggregation characterized by $\epsilon=1$; see Fig.~\ref{fig:figure-3-I}A) was compared with:
\begin{itemize}
    \item two elliptic aggregations promoting WSD over WM, i.e. characterized by $\epsilon = 0.4 < 1$ and $\epsilon = 0.8 < 1$ (see Figs.~\ref{fig:figure-3-I}B~and~C),
    \item an elliptic aggregation promoting WM over WSD, i.e. characterized by $\epsilon = 2.3 > 1$ (see Fig.~\ref{fig:figure-3-I}D),
    \item the $\mathsf{M}$ aggregation, in which only WM is taken into account, characterized by the `in limit' situation: $\epsilon = +\infty$ (see Fig.~\ref{fig:figure-5-R-1.00}). Note that the $\mathsf{M}$ aggregation is actually the `in limit' situation for all the considered elliptic aggregations: 
$\mathsf{R}^{\epsilon=\infty} 
=\mathsf{I}^{\epsilon=\infty} 
=\mathsf{A}^{\epsilon=\infty} 
= \mathsf{M}$.
\end{itemize}
Table~\ref{tab:exemplary-alternatives-buses-WMSD-I_aggregs} provides the representation of the considered busses in terms of WM and WSD first, and in terms of the values of the five considered $\mathsf{I}^\epsilon$ aggregations next.

It should be noted that with $\mathsf{I}^\epsilon$ 
the value of $\epsilon$ is limited by 
$E(\mathsf{I}^\epsilon,\mathbf{1}) = 0.683$ (for details on establishing the operational ranges of $\epsilon$ see Appendix~\ref{app:operational_ranges}). 
As a result, only aggregations with $\epsilon \in (0.683, +\infty)$ should be considered. Clearly, $\epsilon = 0.4$ (see Fig.~\ref{fig:figure-3-I}B) does not belong to this interval, which means that this particular result should not be treated as valid and is discussed here with the sole purpose of illustrating the consequences of violating the ‘maximality/minimality property’ in the case of
$\mathsf{I}^\epsilon$. As defined in~\cite{SS2023}
`maximality/minimality property' requires that within WMSD:
\begin{itemize}
    \item the minimal value of the aggregation is achieved only for $[0, 0]$,
    \item the maximal value of the aggregation is achieved only for $[mean(\mathbf{w}), 0]$.
\end{itemize}
Notice that such requirement is absolutely crucial in the aggregation context because aggregations that do not satisfy `maximality/minimality' may be shown to cease being dominance-compliant.

\begin{table*}[htb]
\footnotesize
  \centering
  \caption{Description of chosen alternatives in terms of WM and WSD, and in terms of five $\mathsf{I}^\epsilon$ aggregations; bracketed upper indices show their positions in the ranking.}
    \begin{tabular}{@{}l|cc|ccccc@{}}
\toprule
\qquad & \multicolumn{2}{c}{WMSD-space} & \multicolumn{5}{c}{Aggregations} \\
Bus & WM & WSD 
		  &$\mathsf{I}^{\epsilon=1}$
        &$\mathsf{I}^{\epsilon=0.4}$
        &$\mathsf{I}^{\epsilon=0.8}$
        &$\mathsf{I}^{\epsilon=2.3}$
        &$\mathsf{I}^{\epsilon=\infty} = \mathsf{M}$
    \\ \midrule
$\mathbf{b}_{03}$ & $0.50$ & $0.22$ & $0.454^{(8)}$ & $0.283^{(8)}$ & $0.429^{(8)}$ & $0.491^{(8)}$ & $0.500^{(7)}$ \\ 
$\mathbf{b}_{07}$ & $0.95$ & $0.09$ & $0.897^{(3)}$ & $0.784^{(3)}$ & $0.877^{(3)}$ & $0.937^{(2)}$ & $0.950^{(2)}$ \\ 
$\mathbf{b}_{14}$ & $0.62$ & $0.22$ & $0.561^{(7)}$ & $0.361^{(7)}$ & $0.531^{(7)}$ & $0.608^{(7)}$ & $0.620^{(6)}$ \\ 
$\mathbf{b}_{15}$ & $0.45$ & $0.32$ & $0.364^{(10)}$ & $0.073^{(10)}$ & $0.320^{(10)}$ & $0.433^{(10)}$ & $0.450^{(8)}$ \\ 
$\mathbf{b}_{16}$ & $0.88$ & $0.09$ & $0.850^{(4)}$ & $0.758^{(4)}$ & $0.836^{(4)}$ & $0.874^{(4)}$ & $0.880^{(4)}$ \\ 
$\mathbf{b}_{18}$ & $0.88$ & $0.11$ & $0.837^{(5)}$ & $0.717^{(6)}$ & $0.818^{(6)}$ & $0.871^{(5)}$ & $0.880^{(4)}$ \\ 
$\mathbf{b}_{22}$ & $0.45$ & $0.31$ & $0.369^{(9)}$ & $0.091^{(9)}$ & $0.327^{(9)}$ & $0.434^{(9)}$ & $0.450^{(8)}$ \\ 
$\mathbf{b}_{24}$ & $0.96$ & $0.06$ & $0.928^{(1)}$ & $0.854^{(1)}$ & $0.915^{(1)}$ & $0.952^{(1)}$ & $0.960^{(1)}$ \\ 
$\mathbf{b}_{25}$ & $0.87$ & $0.10$ & $0.836^{(6)}$ & $0.733^{(5)}$ & $0.820^{(5)}$ & $0.863^{(6)}$ & $0.870^{(5)}$ \\ 
$\mathbf{b}_{26}$ & $0.94$ & $0.08$ & $0.900^{(2)}$ & $0.804^{(2)}$ & $0.883^{(2)}$ & $0.931^{(3)}$ & $0.940^{(3)}$ \\ 
\bottomrule
    \end{tabular}%
\label{tab:exemplary-alternatives-buses-WMSD-I_aggregs}
\end{table*}%

\noindent
As was the case with aggregation $\mathsf{R}^\epsilon$, the $\epsilon$ parameter influences the shape of the isolines of aggregation $\mathsf{I}^\epsilon$, changing them from circular ones (Fig.~\ref{fig:figure-3-I}A) to horizontally elongated elliptical ones (Figs.~\ref{fig:figure-3-I}B and C) on one hand, or from circular ones (Fig.~\ref{fig:figure-3-I}A) through vertically elongated elliptical ones (Fig.~\ref{fig:figure-3-I}D) all the way to straight vertical lines (Fig.~\ref{fig:figure-5-R-1.00}), on the other. 
In the same manner, because the final rating of an alternative is directly influenced by the shape of these isolines, the $\epsilon$ parameter has a direct impact on the position of the alternative in the final ranking.

Since the discussion about alternatives' ratings under $\mathsf{I}^\epsilon$ for different values of $\epsilon$ is very much analogous to that for aggregation $\mathsf{R}^\epsilon$, let us focus only on the case when $\epsilon = 0.4$ (i.e., the value that falls out of the feasible limit: $\epsilon \in (0.683, +\infty)$). 
First of all, notice the importance of the general, dataset-independent analyses conducted using WMSD-space, as only such reveal that  $\mathsf{I}^{\epsilon=0.4}$ violates the ‘maximality/minimality property’. Figure~\ref{fig:figure-3-I}B exposes that the isolines are so horizontally elongated that within the WMSD-space there occur points characterized by not-allowed, negative values of the aggregation function (marked with shades of grey-blue). The $\mathsf{I}^{\epsilon=0.4}$ aggregation thus ceases to be dominance-compliant as the lowest values are no longer related to the anti-ideal point. This is counter-intuitive for decision makers and, as such, ‘too horizontal’ aggregations should not be used. 
Let us additonally stress that looking at a particular dataset only (as in $\mathsf{I}^{\epsilon=0.4}$ column in Table~\ref{tab:exemplary-alternatives-buses-WMSD-I_aggregs}), a decision maker might not see anything worrisome, as the particular values of the aggregation might not exceed the $(0, 1)$ range. A user unaware of the limit: $\epsilon \in (0.683, +\infty)$, could thus proceed with the faulty aggregation in his application and make decisions based on it.

\begin{figure*}[!htb]
\centering
\begin{tabular}{cc}
\multicolumn{1}{l}{\quad A} & \multicolumn{1}{l}{\quad B}\\
\includegraphics[trim = 0mm 20mm 0mm 10mm, clip, width=0.475\textwidth]  {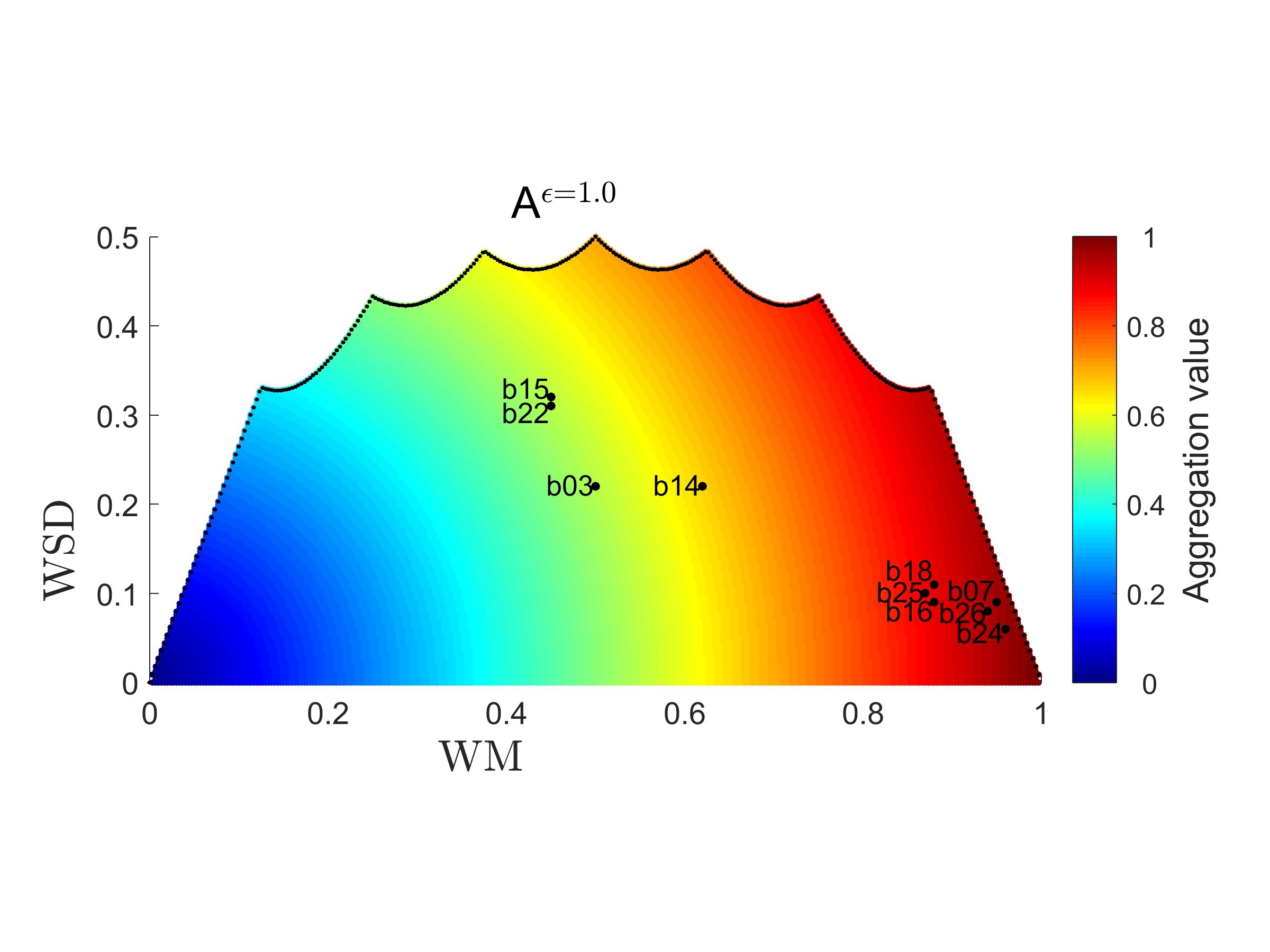} &
\includegraphics[trim = 0mm 20mm 0mm 10mm, clip, width=0.475\textwidth]  {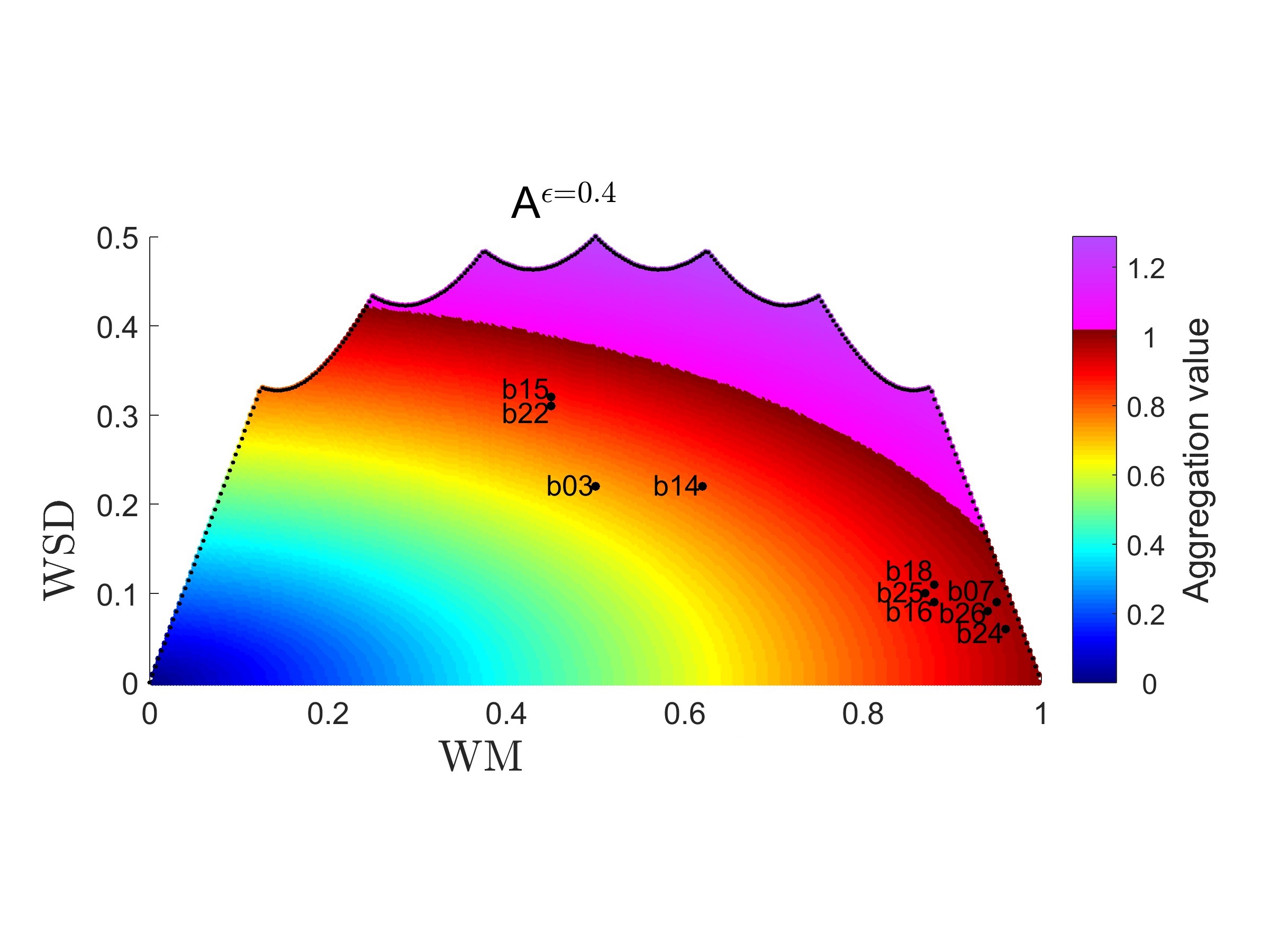}\\
\multicolumn{1}{l}{\quad C} & \multicolumn{1}{l}{\quad D}\\
\includegraphics[trim = 0mm 20mm 0mm 10mm, clip, width=0.475\textwidth]  {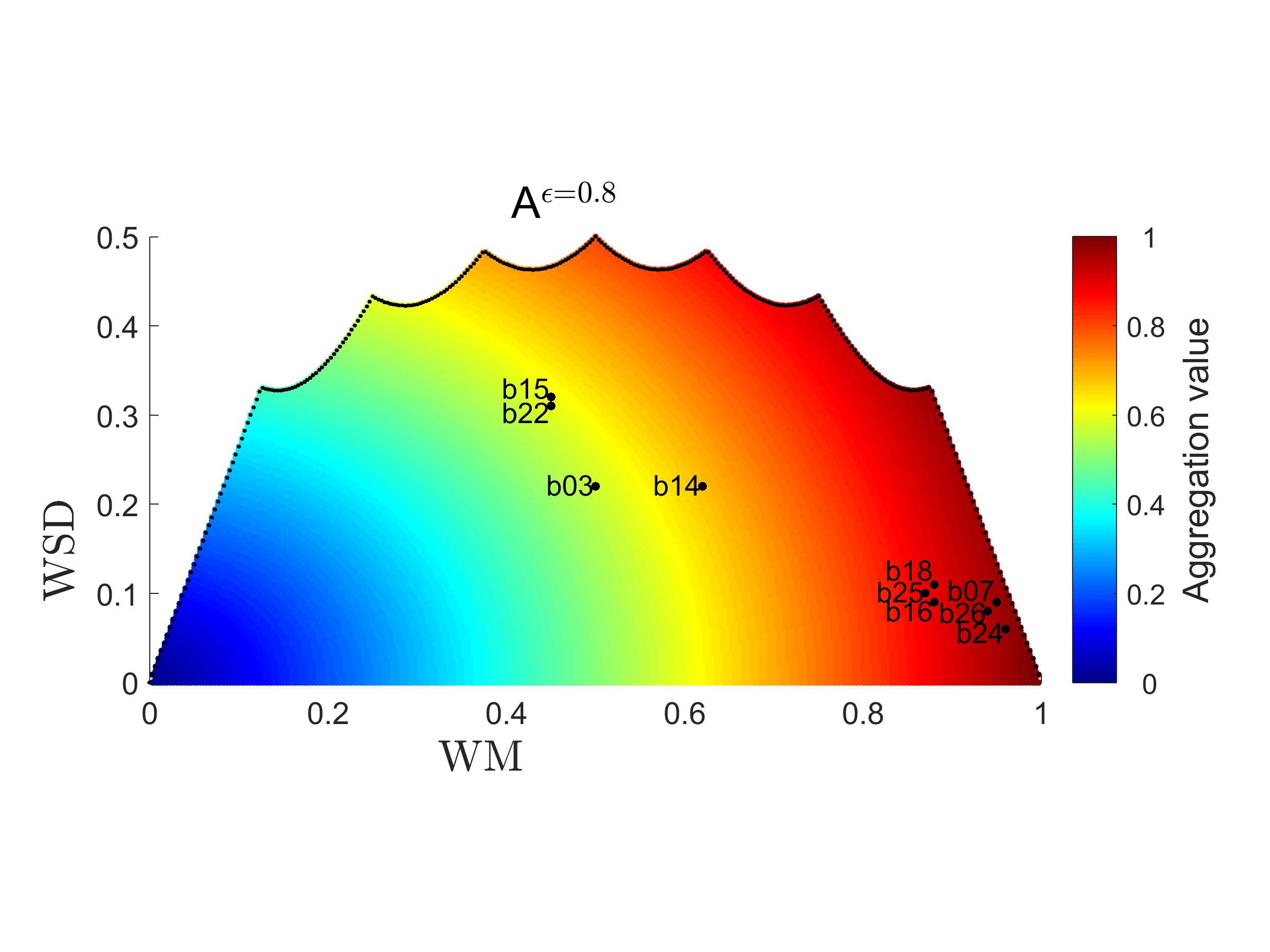} &
\includegraphics[trim = 0mm 20mm 0mm 10mm, clip, width=0.475\textwidth]  {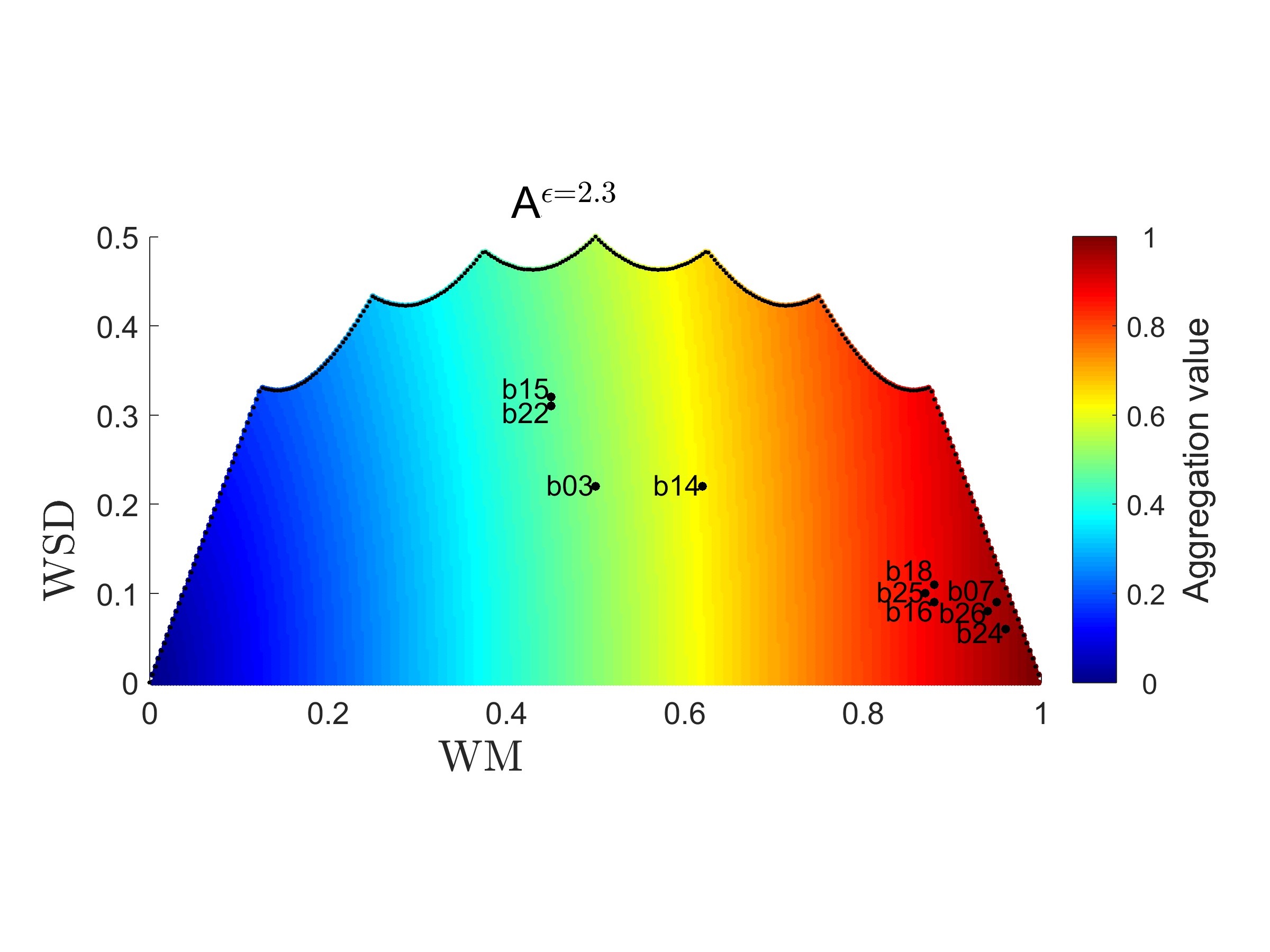}\\
\end{tabular}

\caption{The WMSD-space defined by $\mathbf{w} = \mathbf{1}$ with points representing the alternatives\\ against elliptic aggregation $\mathsf{A}^\epsilon$ for: (A) $\epsilon = 1$ -- equivalent to circular aggregation $\mathsf{A}$, (B) and (C) $\epsilon = 0.4$ and $\epsilon = 0.8$ -- aggregation promoting WSD over WM, (D) $\epsilon = 2.3$ -- aggregation promoting WM over WSD.}
\label{fig:figure-3-A}
\end{figure*}

Fully analogous considerations hold for the $\mathsf{A}$ and $\mathsf{A}^\epsilon$ aggregations, which are depicted in Fig.~\ref{fig:figure-3-A}, while their application to buses dataset is presented in Table~\ref{tab:exemplary-alternatives-buses-WMSD-A_aggregs}.

\begin{table*}[htb]
\footnotesize
  \centering
  \caption{Description of chosen alternatives in terms of WM and WSD, and in terms of five $\mathsf{A}^\epsilon$ aggregations; bracketed upper indices show their positions in the ranking.}
    \begin{tabular}{@{}l|cc|ccccc@{}}
\toprule
\qquad & \multicolumn{2}{c}{WMSD-space} & \multicolumn{5}{c}{Aggregations} \\
Bus & WM & WSD 
		  &$\mathsf{A}^{\epsilon=1}$
        &$\mathsf{A}^{\epsilon=0.4}$
        &$\mathsf{A}^{\epsilon=0.8}$
        &$\mathsf{A}^{\epsilon=2.3}$
        &$\mathsf{A}^{\epsilon=\infty} = \mathsf{M}$
    \\ \midrule
$\mathbf{b}_{03}$ & $0.50$ & $0.22$ & $0.546^{(9)}$ & $0.717^{(10)}$ & $0.571^{(10)}$ & $0.509^{(7)}$ & $0.500^{(7)}$ \\ 
$\mathbf{b}_{07}$ & $0.95$ & $0.09$ & $0.954^{(2)}$ & $0.973^{(1)}$ & $0.957^{(2)}$ & $0.951^{(2)}$ & $0.950^{(2)}$ \\ 
$\mathbf{b}_{14}$ & $0.62$ & $0.22$ & $0.658^{(7)}$ & $0.805^{(9)}$ & $0.678^{(7)}$ & $0.627^{(6)}$ & $0.620^{(6)}$ \\ 
$\mathbf{b}_{15}$ & $0.45$ & $0.32$ & $0.552^{(8)}$ & $0.872^{(7)}$ & $0.602^{(8)}$ & $0.470^{(8)}$ & $0.450^{(8)}$ \\ 
$\mathbf{b}_{16}$ & $0.88$ & $0.09$ & $0.885^{(5)}$ & $0.905^{(5)}$ & $0.887^{(5)}$ & $0.881^{(4)}$ & $0.880^{(4)}$ \\ 
$\mathbf{b}_{18}$ & $0.88$ & $0.11$ & $0.887^{(4)}$ & $0.917^{(4)}$ & $0.891^{(4)}$ & $0.881^{(4)}$ & $0.880^{(4)}$ \\ 
$\mathbf{b}_{22}$ & $0.45$ & $0.31$ & $0.546^{(9)}$ & $0.852^{(8)}$ & $0.594^{(9)}$ & $0.469^{(9)}$ & $0.450^{(8)}$ \\ 
$\mathbf{b}_{24}$ & $0.96$ & $0.06$ & $0.962^{(1)}$ & $0.970^{(2)}$ & $0.963^{(1)}$ & $0.960^{(1)}$ & $0.960^{(1)}$ \\ 
$\mathbf{b}_{25}$ & $0.87$ & $0.10$ & $0.876^{(6)}$ & $0.901^{(6)}$ & $0.879^{(6)}$ & $0.871^{(5)}$ & $0.870^{(5)}$ \\ 
$\mathbf{b}_{26}$ & $0.94$ & $0.08$ & $0.943^{(3)}$ & $0.958^{(3)}$ & $0.945^{(3)}$ & $0.941^{(3)}$ & $0.940^{(3)}$ \\ 
\bottomrule
    \end{tabular}%
\label{tab:exemplary-alternatives-buses-WMSD-A_aggregs}
\end{table*}%

\subsection{Lexicographic aggregations}
\begin{figure*}[!htb]
\centering
\includegraphics[trim = 0mm 30mm 0mm 22mm, clip, width=0.45\textwidth]  {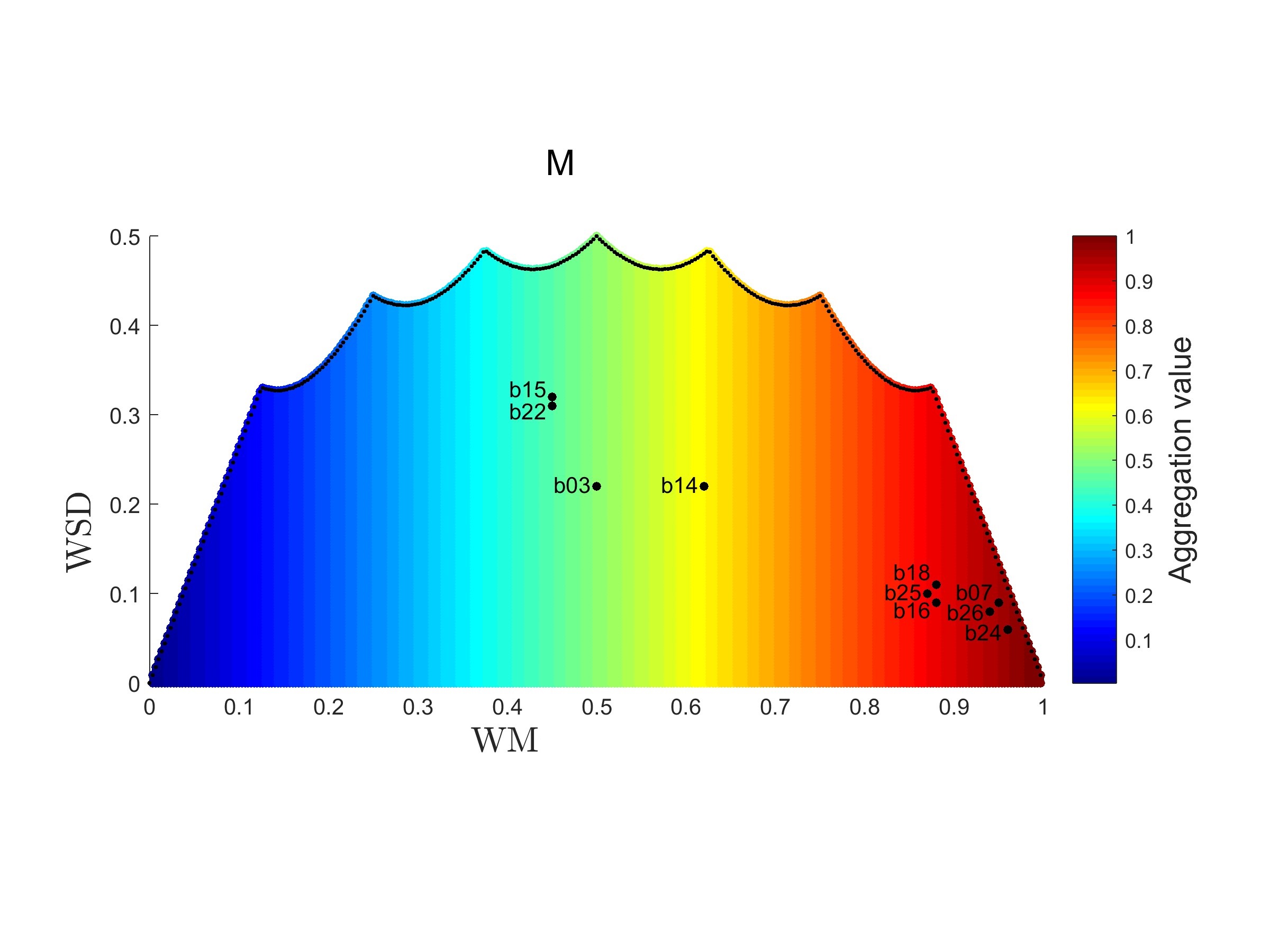}
\includegraphics[trim = 0mm 30mm 0mm 22mm, clip, width=0.45\textwidth]  {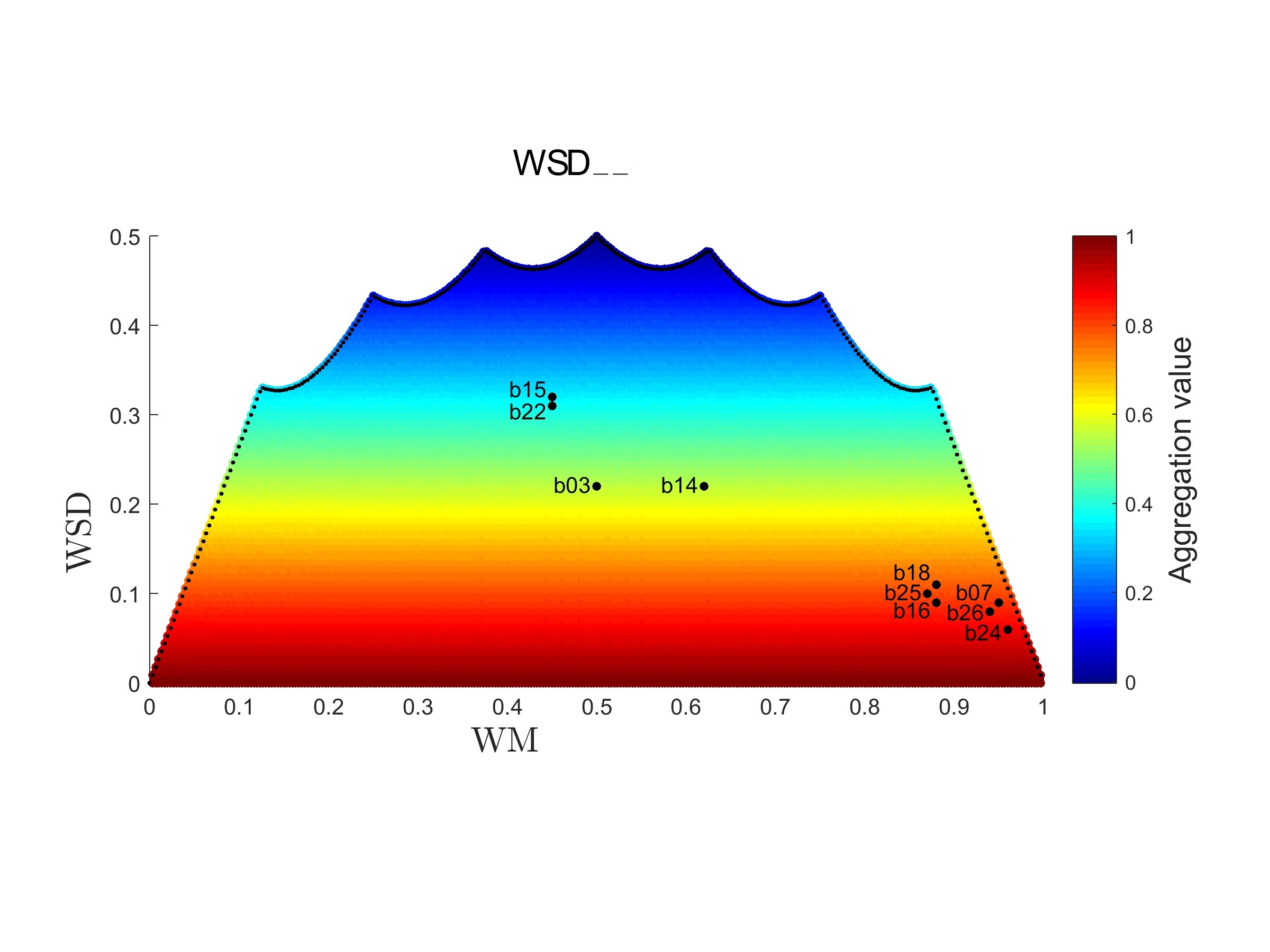}
\captionsetup{justification=centering}
\caption{The WMSD-space defined by $\mathbf{w} = \mathbf{1}$ with points representing the alternatives\\ against the two components of the two-dimensional lexicographic aggregation $\mathsf{I}^L$: component WM that comes down to aggregation M, and component WSD.}
\label{fig:figure-14-lexi-I}
\end{figure*}

\begin{figure*}[!htb]
\centering
\includegraphics[trim = 0mm 30mm 0mm 22mm, clip, width=0.45\textwidth]  {figure-5-I-1.00.jpg}
\includegraphics[trim = 0mm 30mm 0mm 22mm, clip, width=0.45\textwidth]  {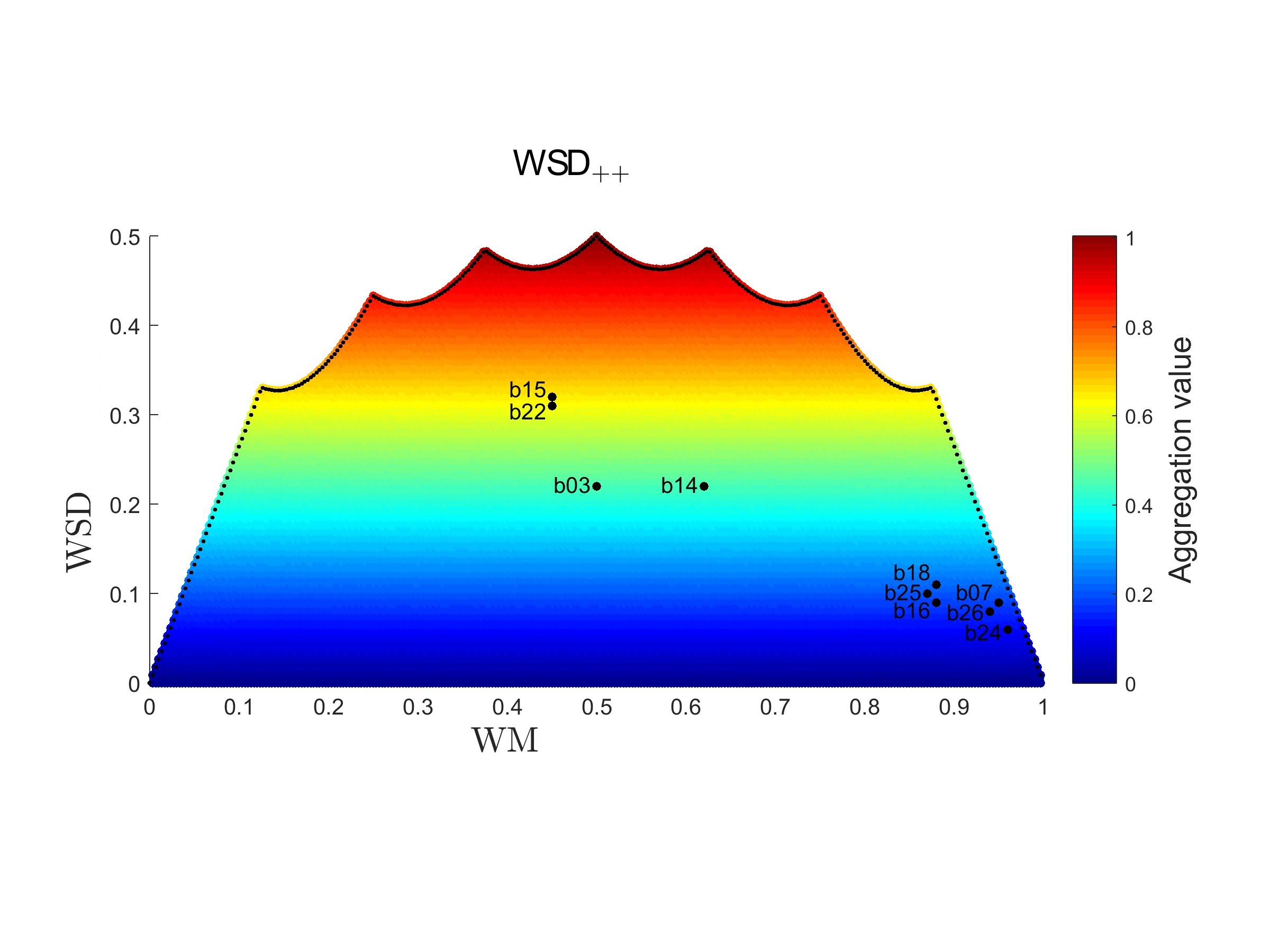}
\captionsetup{justification=centering}
\caption{The WMSD-space defined by $\mathbf{w} = \mathbf{1}$ with points representing the alternatives\\ against the two components of the two-dimensional lexicographic aggregation $\mathsf{A}^L$: component WM that comes down to aggregation M, and component WSD.}
\label{fig:figure-13-lexi-A}
\end{figure*}

\begin{figure*}[!htb]
\centering
\includegraphics[trim = 0mm 30mm 0mm 22mm, clip, width=0.45\textwidth]  {figure-5-I-1.00.jpg}
\includegraphics[trim = 0mm 30mm 0mm 22mm, clip, width=0.45\textwidth]  {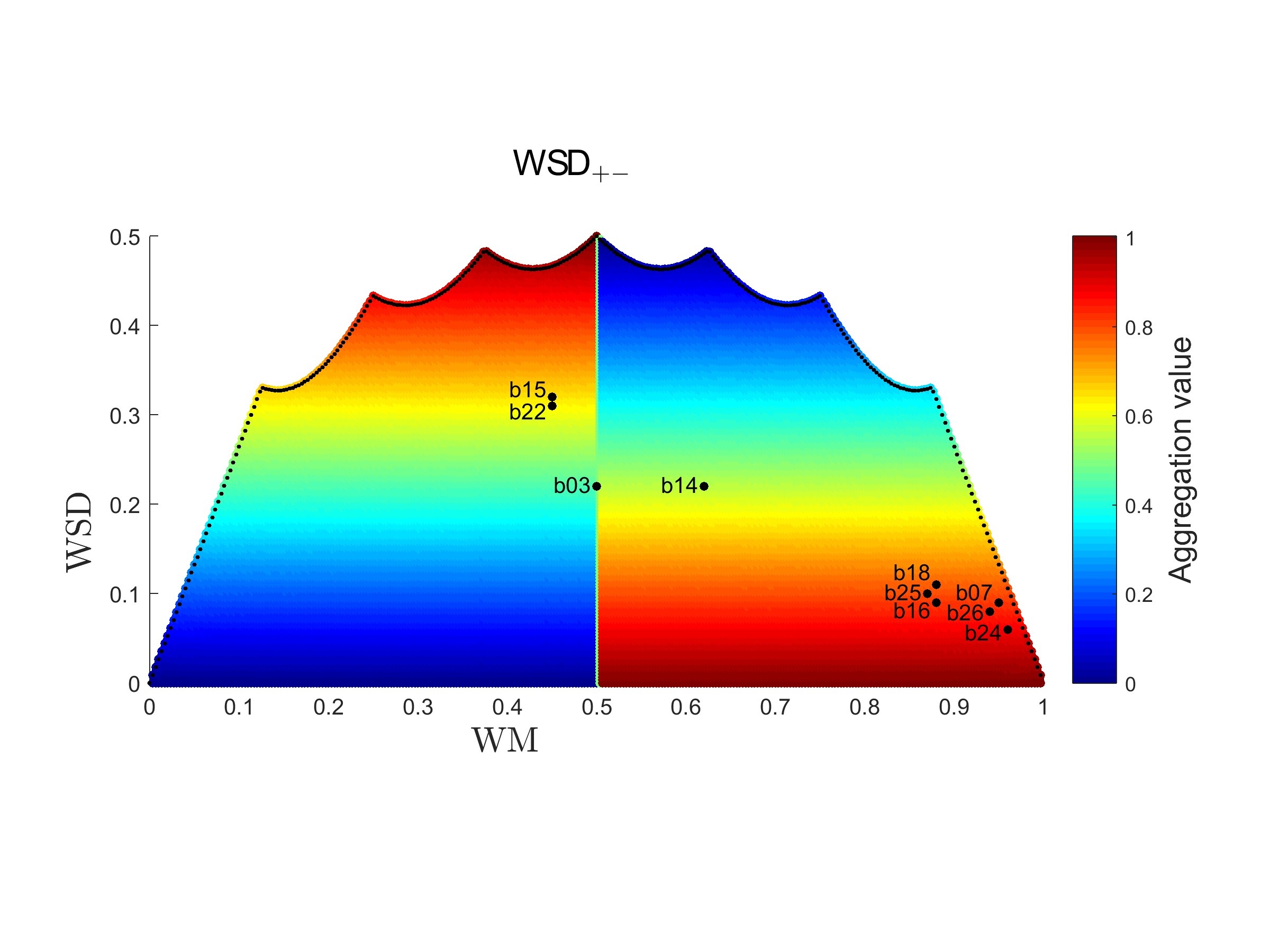}
\captionsetup{justification=centering}
\caption{The WMSD-space defined by $\mathbf{w} = \mathbf{1}$ with points representing the alternatives\\ against the two components of the two-dimensional lexicographic aggregation $\mathsf{R}^L$: component WM that comes down to aggregation M, and component WSD.}
\label{fig:figure-15-lexi-R}
\end{figure*}

Now, let us compares the three aggregations: $\mathsf{I}^L$, $\mathsf{A}^L$ and $\mathsf{R}^L$ in the context of the dataset describing the technical condition of buses. Since the discussed aggregations are two-dimensional lexicographic, their values for each alternative are given as pairs, representing the two components related to WM and WSD (Table~\ref{tab:exemplary-alternatives-buses-WMSD-Lex_aggregs}). Additionally, the table shows the ranking positions of alternatives as bracketed upper indices. The visualization of each of the considered aggregations requires two WMSD-spaces: one for each component constituting the aggregation (see Figs.~\ref{fig:figure-14-lexi-I}--\ref{fig:figure-15-lexi-R} with their left panels corresponding to the WM component and their right panels corresponding to WSD component of the respective aggregations). Recall that the influence of both components is designed to resemble the gain/cost type of relationship between WM and WSD in classic TOPSIS (see Section~\ref{sec:TOPSIS_aggregations} and~\cite{susmaga2023WMSD}). 
\begin{table*}[htb]
\footnotesize
  \centering
  \caption{Description of chosen alternatives in terms of WM and WSD and in terms of three two-dimensional lexicographic aggregations; bracketed upper indices show their positions in the ranking.}
    \begin{tabular}{@{}l|cc|ccc@{}}
\toprule
\qquad & \multicolumn{2}{c}{WMSD-space} & \multicolumn{3}{c}{Aggregations} \\
Bus & WM & WSD 
		  &$\mathsf{I}^L$
        &$\mathsf{A}^L$
        &$\mathsf{R}^L$
    \\ \midrule
$\mathbf{b}_{03}$  &  0.50  &  0.22   & $(0.50,-0.22)^{(8)}$ & $(0.50,+0.22)^{(8)}$  & $(0.50,\,\,\,0.00)^{(8)}$\\
$\mathbf{b}_{07}$  &  0.95  &  0.09  & $(0.95,-0.09)^{(2)}$ & $(0.95,+0.09)^{(2)}$   & $(0.95,-0.09)^{(2)}$  \\
$\mathbf{b}_{14}$  &  0.62  &  0.22  & $(0.62,-0.22)^{(7)}$ & $(0.62,+0.22)^{(7)}$   & $(0.62,-0.22)^{(7)}$  \\
$\mathbf{b}_{15}$  &  0.45  &  0.32  & $(0.45,-0.32)^{(10)}$ & $(0.45,+0.32)^{(9)}$  & $(0.45,+0.32)^{(9)}$  \\
$\mathbf{b}_{16}$  &  0.88  &  0.09  & $(0.88,-0.09)^{(4)}$ & $(0.88,+0.09)^{(5)}$   & $(0.88,-0.09)^{(4)}$  \\
$\mathbf{b}_{18}$  &  0.88  &  0.11  & $(0.88,-0.11)^{(5)}$ & $(0.88,+0.11)^{(4)}$   & $(0.88,-0.11)^{(5)}$  \\
$\mathbf{b}_{22}$  &  0.45  &  0.31   & $(0.45,-0.31)^{(9)}$ & $(0.45,+0.31)^{(10)}$ & $(0.45,+0.31)^{(10)}$ \\
$\mathbf{b}_{24}$  &  0.96  &  0.06  & $(0.96,-0.06)^{(1)}$ & $(0.96,+0.06)^{(1)}$   & $(0.96,-0.06)^{(1)}$  \\
$\mathbf{b}_{25}$  &  0.87  &  0.10  & $(0.87,-0.10)^{(6)}$ & $(0.87,+0.10)^{(6)}$   & $(0.87,-0.10)^{(6)}$  \\
$\mathbf{b}_{26}$  &  0.94  &  0.08  & $(0.94,-0.08)^{(3)}$ & $(0.94,+0.08)^{(3)}$   & $(0.94,-0.08)^{(3)}$  \\
\bottomrule
    \end{tabular}%
\label{tab:exemplary-alternatives-buses-WMSD-Lex_aggregs}
\end{table*}%

Clearly, the impact of the WM component in all the lexicographic aggregations is the same as all the left panels on Figs.~\ref{fig:figure-14-lexi-I}--\ref{fig:figure-15-lexi-R} are the same. The increase of WM always results in a higher aggregation value, which is visualized as vertical lines i.e. the aggregation $\mathsf{M}$. Even slight differences in WM translate to different ranking positions of the alternatives. This can be observed for e.g. buses $\mathbf{b}_{07}$ and $\mathbf{b}_{26}$: the WM of $\mathbf{b}_{07}$ is only 0.01 higher than WM of $\mathbf{b}_{26}$, but it results in $\mathbf{b}_{07}$ being positioned higher than $\mathbf{b}_{26}$ by $\mathsf{I}^L$, $\mathsf{A}^L$ and $\mathsf{R}^L$. 

By the definition of two-dimensional lexicographic aggregations, only when the first component is not able to differentiate the alternatives, is the second component taken into account. In case of $\mathsf{I}^L$, $\mathsf{A}^L$ and $\mathsf{R}^L$ the second components is related to WSD (see the right panels of Figs.~\ref{fig:figure-14-lexi-I}--\ref{fig:figure-15-lexi-R}). Though all the considered right panels have horizontal iso-lines, their colours change in a different manner: from small values at the top to high values at the bottom for $\mathsf{I}^L$ and for the part of $\mathsf{R}^L$ where $\text{WM}>\frac{mean(\mathbf{w})}{2}$, and from small values at the bottom to high values at the top for $\mathsf{A}^L$ and for the part of $\mathsf{R}^L$ where $\text{WM}<\frac{mean(\mathbf{w})}{2}$. It is actually the only feature that makes the considered aggregations different from one another. For aggregation $\mathsf{I}^L$, when alternatives are characterized by the same WM, the smaller values of WSD translate directly to higher ranking positions. This can be observed in Table~\ref{tab:exemplary-alternatives-buses-WMSD-Lex_aggregs} for e.g. bus $\mathbf{b}_{22}$, which is ranked higher than $\mathbf{b}_{15}$ due to its smaller WSD; or, analogously, for $\mathbf{b}_{16}$, which is ranked higher than $\mathbf{b}_{18}$. The opposite happens under $\mathsf{A}^L$, as higher values of WSD result in higher ranking positions: bus $\mathbf{b}_{15}$ is thus ranked higher than $\mathbf{b}_{22}$ (analogously $\mathbf{b}_{18}$ is ranked higher than $\mathbf{b}_{16}$).
Finally, when considering the WSD component, the $\mathsf{R}^L$ aggregation, behaves like $\mathsf{A}^L$ when $\text{WM}<\frac{mean(\mathbf{w})}{2}$ and like $\mathsf{I}^L$ when $\text{WM}>\frac{mean(\mathbf{w})}{2}$. Thus, bus $\mathbf{b}_{15}$ is positioned higher than $\mathbf{b}_{22}$, but $\mathbf{b}_{16}$ higher than $\mathbf{b}_{18}$.

It is worth stressing that the proposed two-dimensional lexicographic aggregations retain the key feature of TOPSIS, since they are in some way dependent on WSD. The influence of WSD becomes evident only when the first component (i.e. WM) is not enough to differentiate between two alternatives; nevertheless, in general, the influence of WSD remains non-zero.

Among all the aggregations considered in this paper, the influence of WSD is the greatest (in the sense that it is never a zero influence) for elliptic aggregations $\mathsf{I}^\epsilon$, $\mathsf{A}^\epsilon$ and $\mathsf{R}^\epsilon$ with $\epsilon \rightarrow +\infty$. This includes the classic TOPSIS aggregations as the special cases of the elliptic ones.
Then, the WSD influence is smaller for the lexicographic aggregations ($\mathsf{I}^L$, $\mathsf{A}^L$ and $\mathsf{R}^L$) as WSD may not have any impact on the ratings provided all alternatives have different values of WM.
Finally, the $\mathsf{M}$ aggregation, equivalent to 
$\mathsf{I}^\epsilon$, $\mathsf{A}^\epsilon$ or $\mathsf{R}^\epsilon$ for $\epsilon = +\infty$ is fully independent of WSD, and as such does not lend TOPSIS its key feature, making it behave like a typical `utility-based methods'.

\section{Discussion}

\begin{figure*}[!h]
\centering
\includegraphics[width=0.9\textwidth]{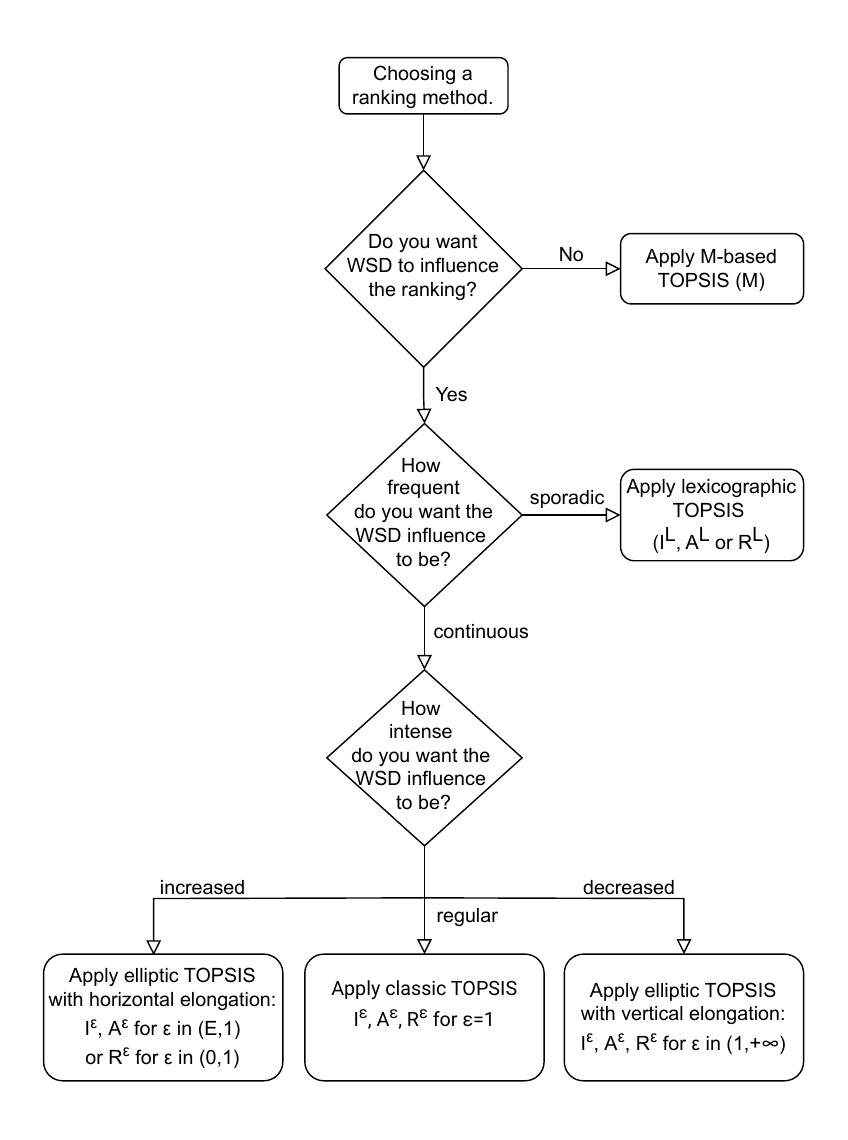}
\caption{The flowchart presenting the process of choosing the ranking method.}
\label{fig:flowchart}
\end{figure*}

In decision making related processes what often appears in a variety of places and contexts is the notion of a risk, which must be then modeled and accordingly handled. For example in finance, specifically in stock portfolio analyses, the mean of some stock prices is treated as the expectation of its return, while the variance of those prices as the indicator of its risk \cite{Mark1952}. 

Remembering that WM is a form of a weighted mean, while WSD is a form of a weighted standard deviation, WM and WSD used here actually represent the same descriptions (with standard deviation being a form of variance). The basic interpretation of these descriptors is as follows: while WM reports the mean of $VS$-based values describing an alternative, WSD reports the dispersion of these values. 
In result, while there exist basically no doubt that WM should be of type `gain', the desired type of WSD is not so obvious. Following the risk-related interpretation, WSD should be of type `cost' (as in finance risk is something to be minimized). This need not be always the case, however, since WSD may be equally well treated as `cost' or some combination of `cost' and `gain', even though such a `double treatment' might seem apparently incoherent.

The explanation is as follows.
As implied by its definition, higher WSD value designates more dispersion of the processed values. The crucial observation now is that among more dispersed values the minimum tends to be lower, while the maximum higher than those among less dispersed ones. And this is how the apparent incoherence of the two opposing treatments is consistently explained: treating WSD as `cost' represents a form of pessimistic approach, while treating WSD as `gain' represents a form of optimistic approach. This is because in the former the potentially lower minima, while in the latter the potentially higher maxima, are kept in mind.

Incidentally, this is what implicitly happens in classic TOPSIS (i.e. with aggregation $\mathsf{R}$), where WSD is given the `double treatment', as it is sometimes assigned type `cost', and sometimes type `gain'. 

More generally (with all three aggregations considered), the WSD treatment depends on the aggregation in the following way:
\begin{itemize}
\item $\mathsf{I}$ always favours low values of WSD (WSD is `cost'),
\item $\mathsf{A}$ always favours high values of WSD (WSD is `gain'),
\item $\mathsf{R}$ favours:
    \begin{itemize}
    \item low values of WSD for WM > $\frac{mean(\mathbf{w})}{2}$ (WSD is `cost').
    \item high values of WSD for WM < $\frac{mean(\mathbf{w})}{2}$ (WSD is `gain'),
    \end{itemize}
\end{itemize}
Notice that for WM = $\frac{mean(\mathbf{w})}{2}$ aggregation $\mathsf{R}$ does not depend on WSD, so WSD is in this particular case `neutral'.

In result, DMs who are inclined to always prefer cases of low dispersion among the values characterizing the alternatives are advised to apply aggregation $\mathsf{I}$. On the other hand, DMs always preferring higher dispersion should choose aggregation $\mathsf{A}$. Interestingly, DMs applying aggregation $\mathsf{R}$ must agree with its `double treatment' of WSD, clearly resulting from the fact that $\mathsf{R}$ is in a way composed of $\mathsf{I}$ and $\mathsf{A}$, which gives it the more sophisticated structure.

By introducing versions of TOPSIS in which the WM and WSD are explicitly used (instead of the distances), this paper allows the DM to gain control over the relative level of influence that WSD exerts on the final result (this, in particular, may prove useful for DMs who expect an unconventional level of this influence). More precisely, the paper shows how the influence of WSD on the method's result (independently of the chosen aggregation) may be increased (effectively promoting WSD over WM) or decreased (effectively promoting WM over WSD). In the latter case, when the influence of WSD is gradually reduced, less and less TOPSIS-like versions of the method are produced.

Summing up, the venues available to the DM are as follows (see~Fig.~\ref{fig:flowchart}):
\begin{itemize}
\item First, the DM must choose whether WSD is to have any influence at all. If no, then aggregation $\mathsf{M}$ should be used, since it uses WM only (alternatively, any of the classic `utility-based methods', e.g. SAW, may be applied.).
\item If yes, the influence of WSD may be chosen to be sporadic or continuous. If sporadic, then any of the lexicographic aggregations ($\mathsf{I}^L$, $\mathsf{A}^L$ or $\mathsf{R}^L$) should be used, since they use WSD only for those alternatives as to which WM was not sufficient to differentiate.
\item If continuous, the influence of WSD may be chosen to be unchanged, decreased or increased, as compared to its standard levels. In the first case any of the classic aggregations ($\mathsf{I}$, $\mathsf{A}$ or $\mathsf{R}$) should be used. Otherwise, the influence may be decreased or increased, in which case elliptic aggregations ($\mathsf{I}^\epsilon$, $\mathsf{A}^\epsilon$ or $\mathsf{R}^\epsilon$) with $\epsilon > 1$ or $\epsilon < 1$, respectively, should be used. This is because these aggregations either decrease or increase the relative influence of WSD below or above the level characteristic to the classic aggregations.
\end{itemize}

\section{Conclusions and Future Works}

Although numerous modifications and adaptations of TOPSIS had been proposed, none of them fully explains the intricate relations between this method and the `utility-based methods' in terms of their ranking generating mechanisms. This was only done in \cite{susmaga2023WMSD} (preceded by \cite{Susmaga_2023MSD}), where distances to the ideal and anti-ideal points were decomposed into what is referred to as the weight-scaled mean (WM) and weight-scaled standard deviation (WSD) of utilities. In result, under the assumption of linearity of utility functions, TOPSIS may be treated as a generalization of `utility-based methods', because while these methods use only WM to generate rankings, TOPSIS uses both WM and WSD (this fact is best illustrated with the WMSD-space visualizations).

This paper continues the task by exploiting the WMSD-space and observing that in this space the standard aggregations are `circular', which means that they possess natural, `elliptic' generalizations. Introduction of such elliptic aggregations, with parametrisation that controls their actual elongation, allows to alter the relative influence that WM and WSD exert on the generated rankings. 
The parametrisation demonstrates that TOPSIS equipped with the parametrised elliptic aggregations becomes a proper generalization of classic TOPSIS and `utility-based methods'. This is because by changing the elongation of the isolines the behaviour of TOPSIS may remain typical for this method or be shifted towards or away from `utility-based methods'. In all those situations, however, the key feature of TOPSIS, i.e. non-zero influence of WSD on the results, is retained.

When the changing the influence of WSD is concerned, the following two theoretical results have been arrived at:
\begin{itemize}
\item for the effect of WSD being increased, it was shown that there exist an upper limit to the allowed increase, which was then identified and described, 
\item for the effect of WSD being decreased, it was shown that the ultimate lower limit to the allowed decrease is zero, which means that the generated rankings depend then on WM only. 
\end{itemize}
As far as the ranking generating mechanism is concerned, the second scenario renders TOPSIS fully equivalent to the `utility-based methods', because then (independently of which of the three aggregations is initially considered) a WM-only aggregation finally arises. This is the situation in which the key feature of TOPSIS is irrevocably lost.

Apart from the three above-mentioned elliptic generalizations of the three standard aggregations, 
this paper puts forward three lexicographic generalizations of the WM-only aggregation.
Although these lexicographic aggregations may be viewed as even less influenced by WSD than any of the elliptic ones, its influence remains non-zero, which means that all these aggregations also retain the key feature of TOPSIS.

It should be stressed that comprehension of the properties of all the introduced aggregations is made especially easy with the WMSD-space visualizations of the isolines of these aggregations. In result, decision-makers equipped with such a tool can swiftly observe and compare the considered aggregations and choose those that best suit their preference. In a truly informed manner they can also use the tool to design and construct new aggregations.

Further investigations regarding TOPSIS may include combining the introduced elliptic generalizations of the method with its other generalizations and extensions, concerning e.g. fuzzy data on input, explicit preferential information, alternate distance measures or the problem of sorting instead of ranking. Although TOPSIS has already been adapted to all those problems, the methodology used in those adaptations strictly follows that of classic TOPSIS, which means that the adaptations equipped with the generalized aggregations might have considerable novelty value.

\bmhead{Acknowledgments}
This research was partly funded by the National Science Centre, Poland, grant number: 2022/47/D/ST6/01770. For the purpose of Open Access, the author has applied a CC-BY public copyright licence to any Author Accepted Manuscript (AAM) version arising from this submission.

\section*{Declarations}
\bmhead{Data availability statement} The data on technical condition of buses analyzed in Section~\ref{sec:Case-Study} is a part of a dataset   from paper~\cite{GSZ13}. To make our study self-contained, our subset is
presented in Table~\ref{tab:exemplary-alternatives-buses-CS} of this article.

\bmhead{Conflict of interest} The authors declare no  conflict of interest in this paper.

\bibliography{ecai}

\newpage
\section*{Appendix}
\appendix

\setcounter{figure}{0}
 \renewcommand{\thefigure}{A\arabic{figure}}

\section{Operational ranges of $\epsilon$ parameter for aggregations $\mathsf{I}^\epsilon$, $\mathsf{A}^\epsilon$ and $\mathsf{R}^\epsilon$}
\label{app:operational_ranges}

\subsection{The `maximality/minimality' property}

To retain non-negative values of aggregations, $\epsilon$ is assumed to satisfy $\epsilon \in [0, +\infty)$. Nonetheless, $\epsilon = 0$ cannot be used with and non-trivial aggregations because it would simply collapse all values to zero, rendering every aggregation a constant (and thus trivial) function. It might just seem that the allowed range of $\epsilon$ is $(0, +\infty)$. However, for aggregations $\mathsf{I}^\epsilon$ and $\mathsf{A}^\epsilon$ there exist a limit, $E > 0$, which is needed to guarantee that (for $\epsilon > E$) the following `maximality/minimality property' holds: within WMSD 
\begin{itemize}
    \item the minimal value of the aggregation is achieved only for $[0, 0]$,
    \item the maximal value of the aggregation is achieved only for $[mean(\mathbf{w}), 0]$.
\end{itemize}
This means that for aggregations $\mathsf{I}^\epsilon$ and $\mathsf{A}^\epsilon$ the allowed range for $\epsilon$ is $(E, +\infty)$. 

Notice that the above-mentioned `maximality/minimality property' of aggregations is fully analogous to the so-called `maximality/minimality property' of confirmation measures introduced in \cite{G_13}, elaborated in \cite{GSS_12} and visualized in \cite{Susmaga2014}. An important difference is that while the property is merely useful in the confirmation context, it is absolutely crucial in the aggregation context. This is because aggregations that do not satisfy `maximality/minimality' may be shown to cease being dominance-compliant.

\begin{figure*}[t!]
\centering
\begin{tabular}{rl}
A & \vspace{-6mm} \\
& \includegraphics[trim = 10mm 42mm 0mm 42mm, clip, width=0.85\textwidth]  {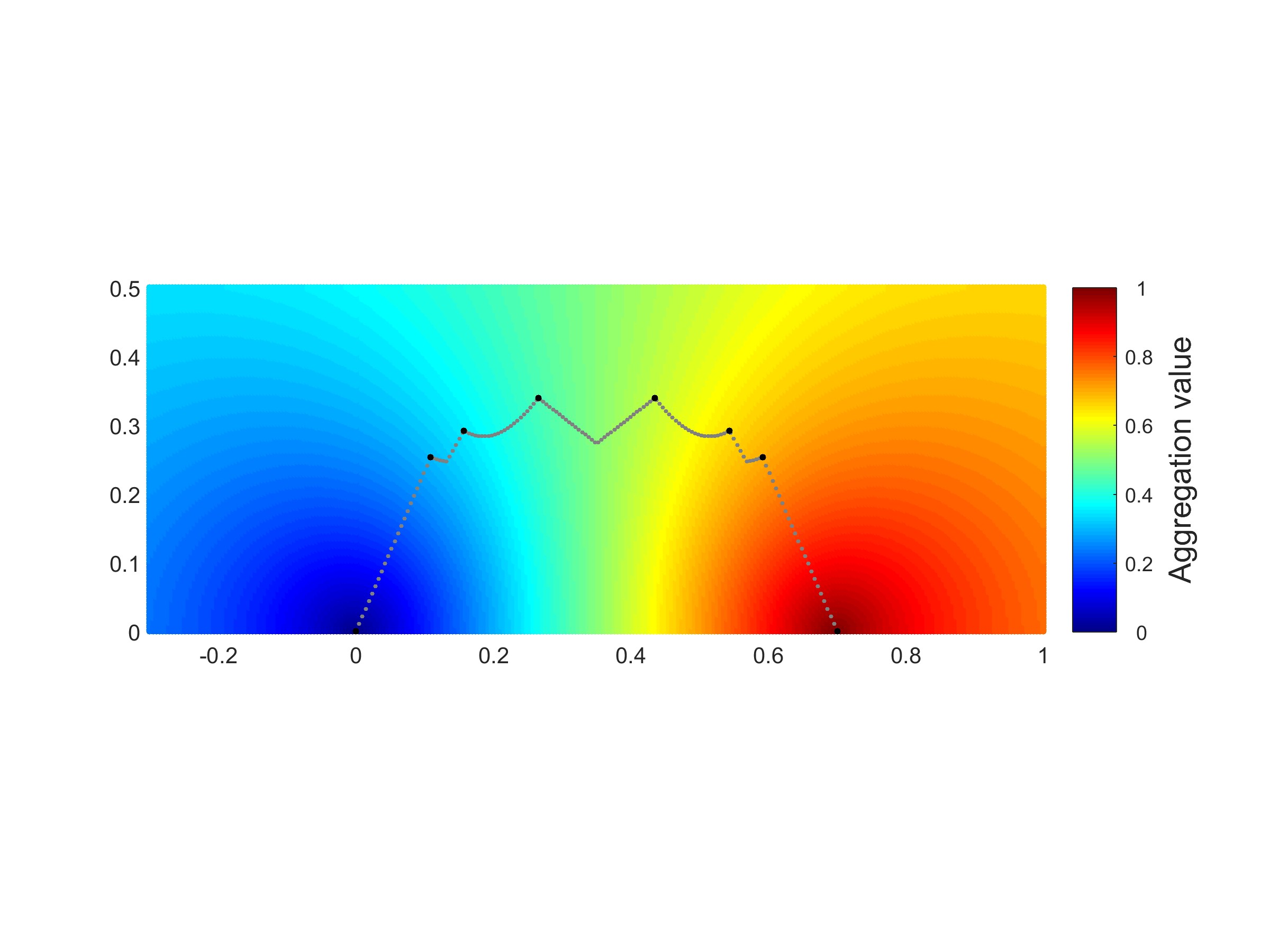} \\
B & \vspace{-6mm} \\
& \includegraphics[trim = 10mm 42mm 0mm 42mm, clip, width=0.85\textwidth]  {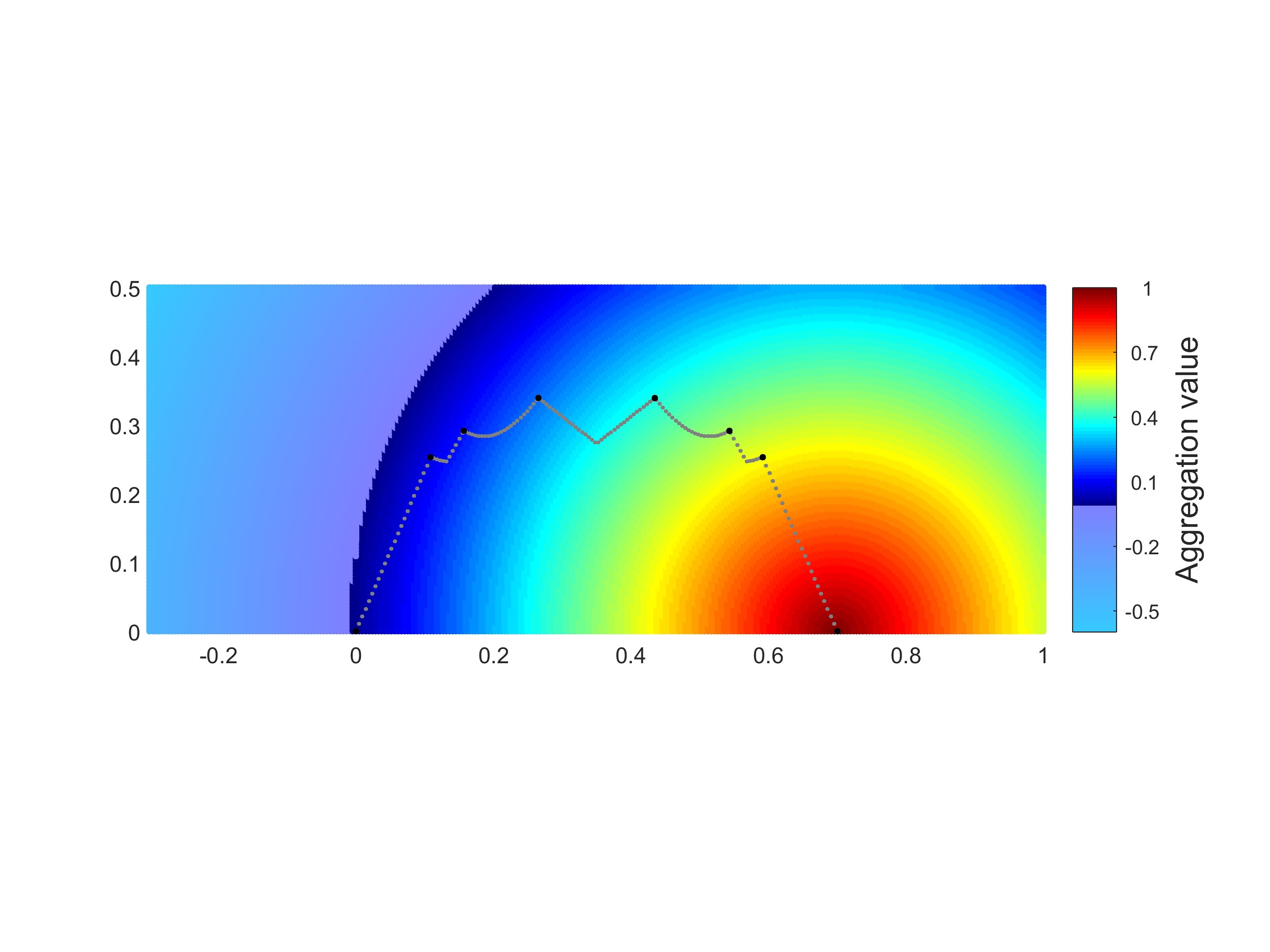} \\
C & \vspace{-6mm} \\
& \includegraphics[trim = 10mm 42mm 0mm 42mm, clip, width=0.85\textwidth]  {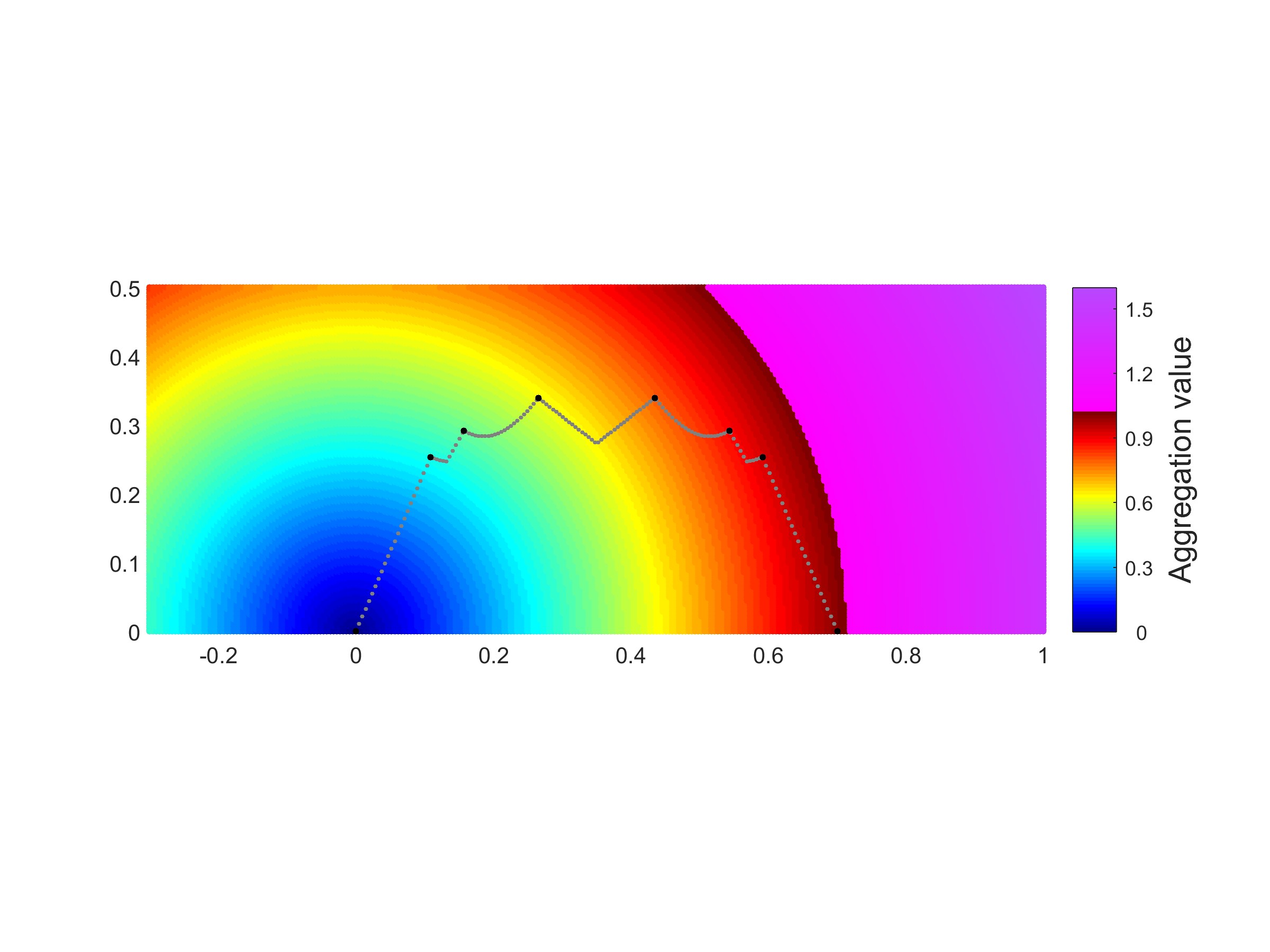} \\
\end{tabular}
\caption{Aggregations: (A) $\mathsf{R}$, (B) $\mathsf{I}$ and (C) $\mathsf{A}$ depicted both within as well as beyond the WMSD-space (defined by $\mathbf{w} = [1.0, 0.6, 0.5]$). Notice the presence of isolines external to $[0, 1]$ in $\mathsf{I}$ ($v$-isolines, where $v < 0$; shaded grey-blue) and in $\mathsf{A}$ ($v$-isolines, where $v > 1$; shaded grey-magenta).}
\label{fig:weightcastle-050610-beyond[01]}
\end{figure*}

\subsection{The case of $\mathsf{R}^\epsilon$}

As opposed to $\mathsf{I}^\epsilon$ and $\mathsf{A}^\epsilon$, aggregation $\mathsf{R}^\epsilon$ is free from the risk of not satisfying the maximality/minimality'. 
This is well illustrated in a WMSD-space based visualization of $\mathsf{R}^\epsilon$, where $\epsilon = 1$ (or, simply, $\mathsf{R}$), as depicted e.g. in Fig.~\ref{fig:weightcastle-050610-beyond[01]}A. Notice that each $v$-isoline of $\mathsf{R}^\epsilon$, where $v \in [0,1]$), coincides (in at least 1 point) with the WMSD-space. This means that the counterdomain of $\mathsf{R}^\epsilon$ contains interval $[0,1]$. Simultaneously, no other isolines (i.e. no $v$-isoline with $v \notin [0,1]$) exist to coincide with WMSD-space. Thus the counterdomain of $\mathsf{R}^\epsilon$ equals $[0,1]$. 
Additionally, $0$ is the minimum of the aggregation (and is attained only in $[0, 0] \in $WMSD), while $1$ is the maximum of the aggregation (and is attained only in $[mean(\mathbf{w}), 0] \in $WMSD).
The `maximality/minimality property' is thus satisfied by $\mathsf{R}^\epsilon$ for $\epsilon = 1$.

Now, consider $\epsilon \neq 1$. After observing how the application of $\epsilon$ influences the shape of the isolines, it is clear that no $\epsilon \in (0, +\infty)$ has any chance of changing the above situation. This is because no $v$-isoline, where $v \in [0,1]$, will stop coinciding with WMSD-space, and no $v$-isoline, where $v \notin [0,1]$, will start coinciding with WMSD-space. In result, for every $\epsilon \in (0, +\infty)$, the counterdomain of $\mathsf{R}^\epsilon$ remains equal to $[0,1]$, with the value and position of its minimum/maximum unchanged.
The `maximality/minimality property' is thus satisfied by $\mathsf{R}^\epsilon$ for $\epsilon \in (0, +\infty)$.

\subsection{The case of $\mathsf{I}^\epsilon$ and $\mathsf{A}^\epsilon$}

As far as aggregations $\mathsf{I}^\epsilon$ and $\mathsf{A}^\epsilon$ are concerned, the values of $\epsilon$ must belong to $(E, +\infty)$. The specific value of $E$ depends on the relation of their isolines and the shape of the WMSD-space which, in turn, depends on $\mathbf{w}$ (see Fig.~\ref{fig:weightcastle-050610-beyond[01]}B and \ref{fig:weightcastle-050610-beyond[01]}C). $E$ is thus a function $E(\mathit{G},\mathbf{w})$ of the aggregation $\mathit{G}$ and the weights $\mathbf{w}$. 

For instance, $E(\mathsf{A}^\epsilon,[1.0, 0.5]) = 0.6667$ (see the WMSD-space depicted in \ref{fig:WMSD-colours}). This is because for $\epsilon \leq 0.6667$ the `maximality/minimality property' is not satisfied with $\mathsf{A}^\epsilon$ -- its maximal value is attained in two points: $[mean([1.0, 0.5]), 0.00] = [0.75, 0.00]$ and $[0.60, 0.30]$ for $\epsilon = 0.6667$, and in even more points for $\epsilon < 0.6667$. Similarly, $E(\mathsf{A}^\epsilon,[1.0, 0.6, 0.5]) = 0.6767$ (see the WMSD-space depicted in Figs.~\ref{fig:aggregations-for-theta-050}--\ref{fig:aggregations-for-theta-100}).

\begin{figure*}[!htb]
\centering
\begin{tabular}{cc}
\multicolumn{1}{l}{\quad A} & \multicolumn{1}{l}{\quad B}\\
\includegraphics[trim = 0mm 10mm 0mm 10mm, clip, width=0.475\textwidth]  {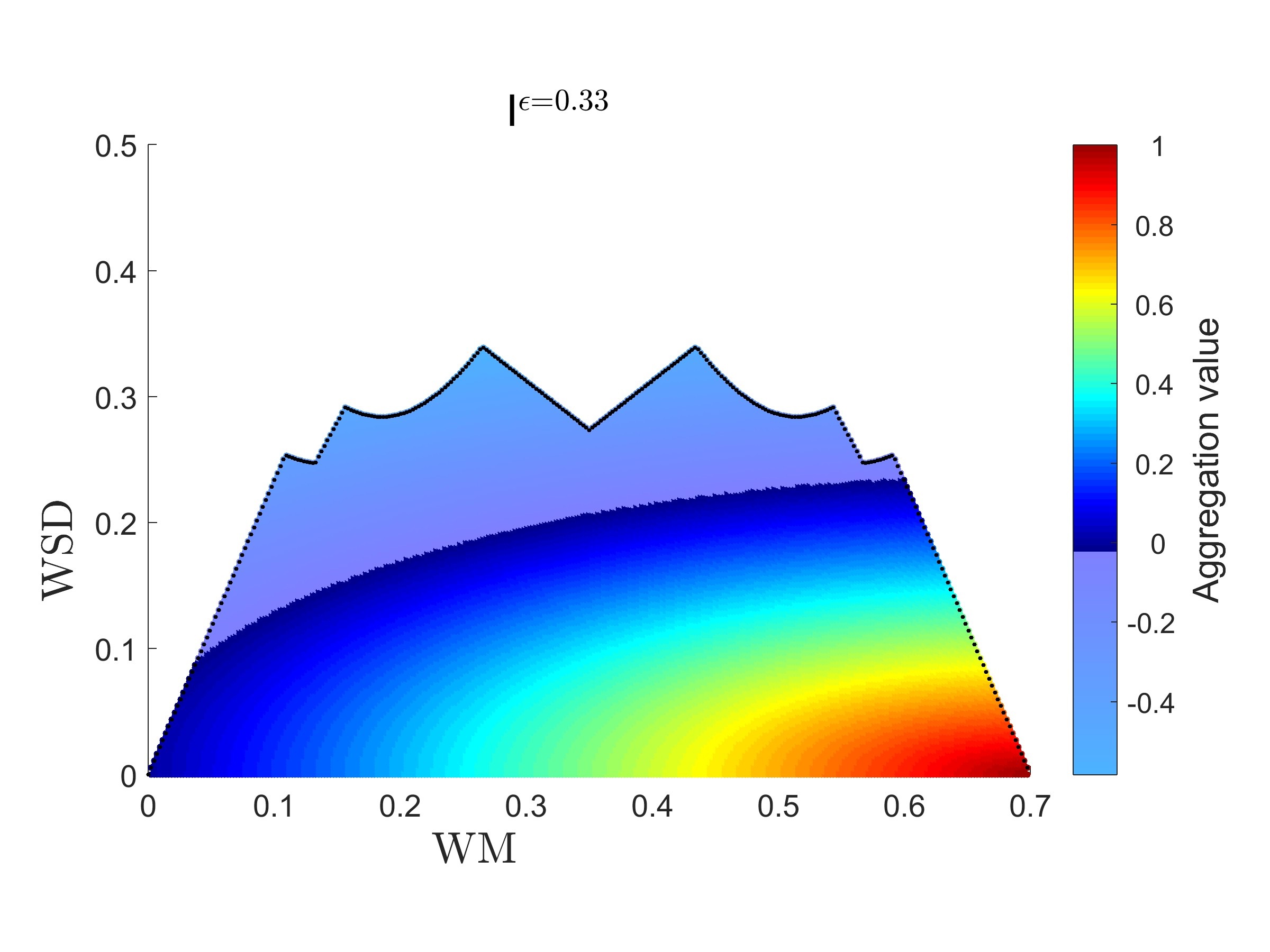} &
\includegraphics[trim = 0mm 10mm 0mm 10mm, clip, width=0.475\textwidth]  {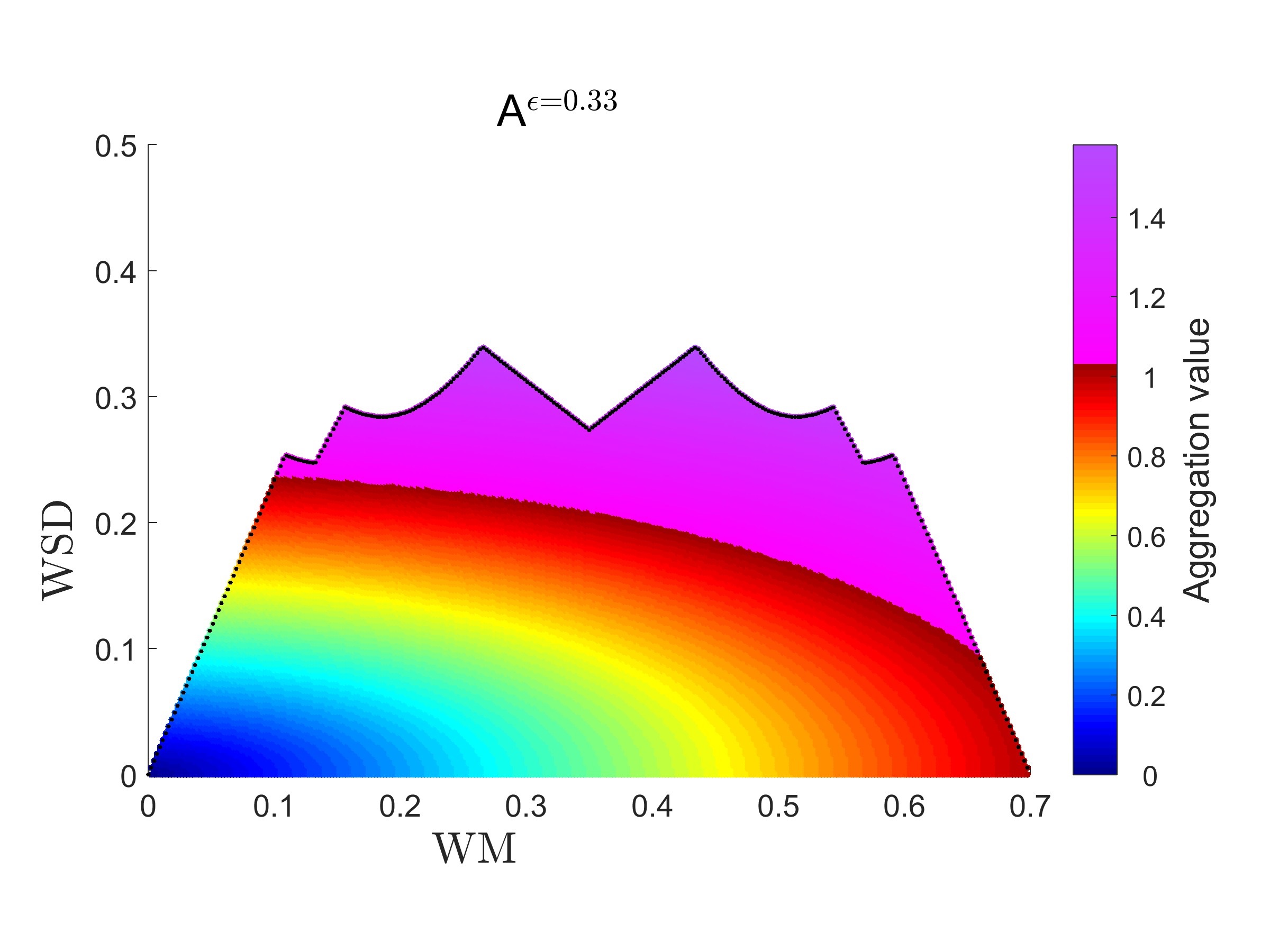}\\
\end{tabular}
\caption{WMSD-space defined by $\mathbf{w} = [1.0, 0.6, 0.5]$ depicted against elliptic aggregations for $0 < \epsilon = 0.3333 < E$, which constitutes clear violation of the `maximality/minimality property: (A) $\mathsf{I}^\epsilon$ with minimum of $-0.58 < 0 $ at $[0.27,0.35]$ instead of $[0.0,0.0]$, (B) $\mathsf{A}^\epsilon$ with maximum of $1.58 > 1$ at $[0.43,0.35]$ instead of $[0.7,0.0]$. Notice that the counterdomains of both aggregations ($[-0.58,1]$ and $[0,1.58]$, respectively, as expressed by the colour bars) are both different from the expected $[0, 1]$.}
\label{fig:aggregations-for-theta-025}
\end{figure*}

As a consequence of violating the `maximality/minimality property' in the case of $\mathsf{I}^\epsilon$ and $\mathsf{A}^\epsilon$, these aggregations cease to be dominance-compliant and one gets seemingly displaced reference points (see Fig.~\ref{fig:aggregations-for-theta-025}), which are counter-intuitive for decision makers.  
To see this, consider the first reference point, i.e. the ideal point (fully analogous reasoning concerns the anti-ideal point),
defined as the vertex $[v_1^*, v_2^*, ..., v_n^*] \in CS$ and characterized by maximal value of WM~$= mean(\mathbf{w})$ and by minimal value of WSD~$ = 0$ in WMSD-space. In classic TOPSIS this point also happens to be characterized by the \emph{only} maximum value of every aggregation. Now, modifying an aggregation to exhibit maxima in any other points seemingly displaces the ideal point and thus severely undermines the method's interpretability and explainability. A case of this troublesome phenomenon is exemplified in Fig.~\ref{fig:aggregations-for-theta-025}B where, as a result of using $\epsilon \leq E$ (which caused the isolines to be `too horizontal'), the maximum of $\mathsf{A}^\epsilon$ appeared also at the top of the WMSD-space, seemingly displacing the ideal point.

\subsection{Derivation of $E(\mathit{G},\mathbf{w})$ for aggregations $\mathsf{I}^\epsilon$ and $\mathsf{A}^\epsilon$}

Let $V_j$, for $j \in \{0, 1, ..., n_p\}$, denote a vertex of the WMSD-space, and $x(V_j)$, $y(V_j)$ its two respective coordinates. 

\begin{figure*}[!h]
\centering
\includegraphics[trim = 25mm 10mm 20mm 8mm, clip, width=0.7\textwidth]{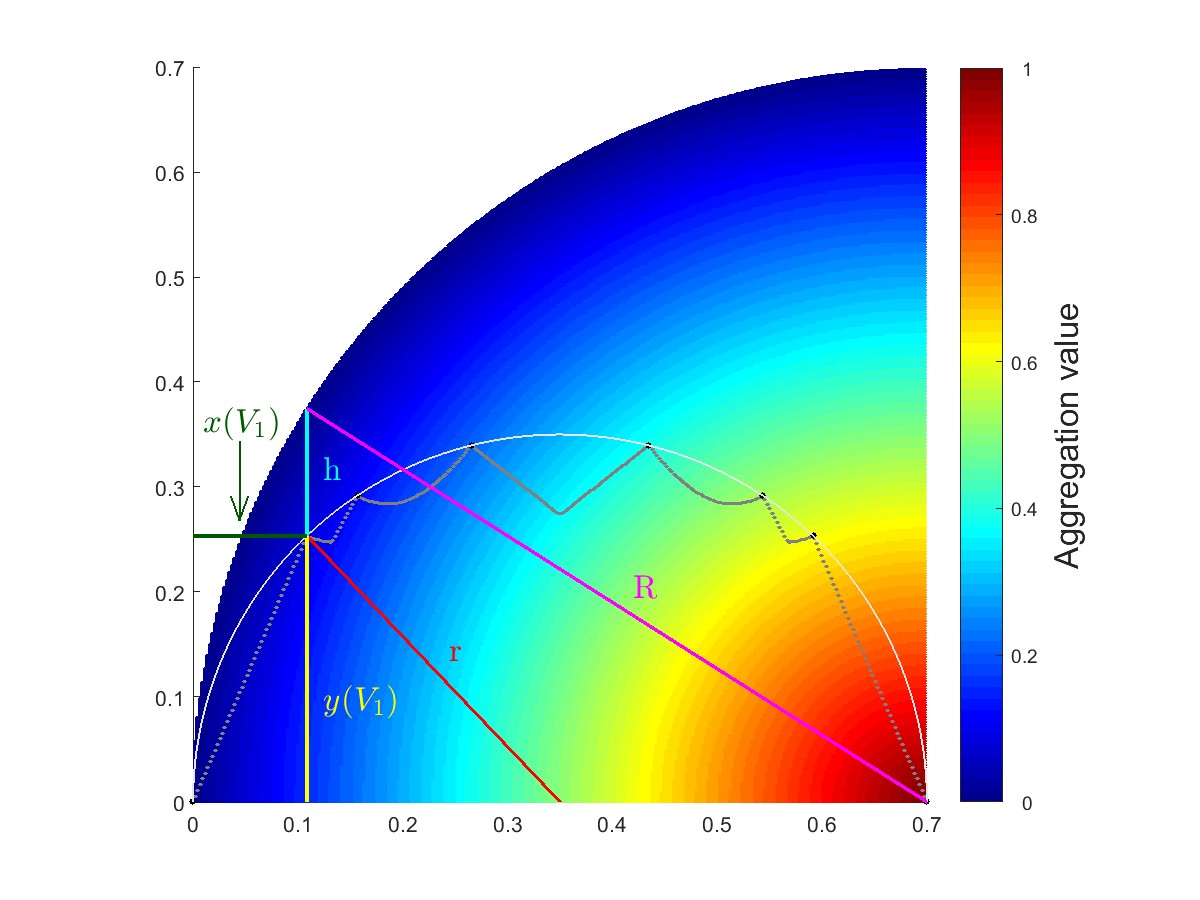}
\caption{An illustration of the $\mathsf{I}$ aggregation depicted both within as well as above the WMSD-space (defined by $\mathbf{w} = [1.0, 0.6, 0.5]$). The named segments are instrumental in deriving the value of $E(\mathsf{I},[1.0, 0.6, 0.5])$.}
\label{fig:weightcastle-050610-derivation}
\end{figure*}

Given $\mathbf{w}$,
it is a feature of the WMSD-space that all its vertices align along a circle of radius $\frac{mean(\mathbf{w})}{2}$ (denotation: r in Fig.~\ref{fig:weightcastle-050610-derivation}) centred at [$\frac{mean(\mathbf{w})}{2}$,0]. This relates the coordinates of each vertex: $(\frac{mean(\mathbf{w})}{2} - x(V_j))^2 + y(V_j)^2 = (\frac{mean(\mathbf{w})}{2})^2$, and allows to express $y(V_j)$ as a function of $x(V_j)$: $y(V_j) = \sqrt{(\frac{mean(\mathbf{w})}{2})^2 - (\frac{mean(\mathbf{w})}{2} - x(V_j))^2}$. 

Now, consider aggregation $\mathsf{I}^\epsilon$: for $\epsilon = 1$ its isolines constitute circles centred at [$mean(\mathbf{w})$,0] (Fig.~\ref{fig:weightcastle-050610-derivation}). As opposed to Figs.~\ref{fig:WMSD-colours}--\ref{fig:aggregations-for-theta-100}, which depict the isolines (of different aggregations) only inside the WMSD-space, Fig.~\ref{fig:weightcastle-050610-derivation} depicts these isolines (of $\mathsf{I}^\epsilon$) also beyond (precisely: in the quadrant delineated by the $[0,mean(\mathbf{w})] \times [0,mean(\mathbf{w})]$ square).
In particular, the 0-isoline for $\epsilon = 1$ constitutes a circle of radius $mean(\mathbf{w})$ (denotation: R in Fig.~\ref{fig:weightcastle-050610-derivation}). As such, it has only one common point ($V_0$) with the WMSD-space, so the aggregation is 0 only in $V_0$ (simultaneously, the aggregation is 1 only in $V_n$, so it upholds the `maximality/minimality property').

Next, recall that within aggregations $\mathsf{I}^\epsilon$ and $\mathsf{A}^\epsilon$ feasible values $\epsilon \neq 1$ make the circles elongated either horizontally or vertically which, in practice, bring the isolines up ($\epsilon > 1$) or down ($\epsilon < 1$). In particular, $\epsilon < 1$ brings down the 0-isoline of $\mathsf{I}^\epsilon$. Incidentally, no $\epsilon$ may make the 0-isoline leave $V_0$, which means that only parts of the 0-isoline outside of $V_0$ are meant in the context of `bringing up or down'. Simultaneously, the fact guarantees that the value of $\mathsf{I}^\epsilon$ is always 0 in at least one point ($V_0$) of WMSD-space. 

Finally, consider $x(V_1)$ and $y(V_1)$ (identical denotations in Fig.~\ref{fig:weightcastle-050610-derivation}): the coordinates of the second leftmost vertex $V_1$.
Given $\mathbf{w} = [w_1, w_2, ..., w_n]$, it may be shown that $x(V_1) = \frac{(min_{w_j>0}(w_j))^2}{s \cdot \norm{\mathbf{w}}}$, which is positive, as opposed to $x(V_0)$, which is by definition zero. 
As already stated, point [$x(V_1)$,$y(V_1)$] lies on the circle of radius $\frac{mean(\mathbf{w})}{2}$ centred at [$\frac{mean(\mathbf{w})}{2}$,0] (which implies that also $y(V_1)$ is positive). It is clear from the relation between this circle and the 0-isoline that point [$x(V_1)$,$y(V_1)$] lies below the 0-isoline. So, there exist $h > 0$ (identical denotation in Fig.~\ref{fig:weightcastle-050610-derivation}) such that point [$x(V_1)$,$y(V_1)+h$] lies on the 0-isoline. This relates the $x(V_1)$ and $y(V_1)+h$ coordinates of this point: $(mean(\mathbf{w}) - x(V_1))^2 + (y(V_1)+h)^2 = mean(\mathbf{w})^2$, and allows to express $y(V_1)+h$ as a function of $x(V_1)$: $y(V_1)+h = \sqrt{mean(\mathbf{w})^2 - (mean(\mathbf{w}) - x(V_1))^2}$. 

Assume that a particular value of $\epsilon$, referred to as $E$, is chosen so that at $x(V_1)$ the following holds: $(y(V_1)+h) \cdot E = y(V_1)$. The value of $\epsilon = E$ brings thus down the 0-isoline of $\mathsf{I}^\epsilon$ from $y(V_1)+h$ to $y(V_1)$. This is exactly what results in the violation of the `maximality/minimality property' -- the value of the aggregation is now 0 in both $V_0$ and $V_1$. Of course, every value of $\epsilon$ lower than $E$ brings down the 0-isoline even further, continuing to violate the property (in such cases the value of $\mathsf{I}^\epsilon$ is 0 in more than two points).

Equation $(y(V_1)+h) \cdot E = y(V_1)$ may now be used to express $E$ as: $E = \frac{y(V_1)}{y(V_1)+h}$. Next, 
functional dependencies: $y(V_1) = \sqrt{(\frac{mean(\mathbf{w})}{2})^2 - (\frac{mean(\mathbf{w})}{2} - x(V_1))^2}$, $y(V_1)+h = \sqrt{mean(\mathbf{w})^2 - (mean(\mathbf{w}) - x(V_1))^2}$ and $x(V_1) = \frac{(min_{w_j>0}(w_j))^2}{s \cdot \norm{\mathbf{w}}}$ may be used to express $E$ as:\\
$
E = 
\frac
{\sqrt{(\frac{mean(\mathbf{w})}{2})^2 - (\frac{mean(\mathbf{w})}{2} - x(V_1))^2}}
{\sqrt{mean(\mathbf{w})^2 - (mean(\mathbf{w}) - x(V_1))^2}}
=
\frac
{\sqrt{mean(\mathbf{w})^2 - (mean(\mathbf{w}) - 2 \cdot x(V_1))^2}}
{2 \cdot \sqrt{mean(\mathbf{w})^2 - (mean(\mathbf{w}) - x(V_1))^2}}
=
\sqrt
{
\frac
{(mean(\mathbf{w})^2 - (mean(\mathbf{w}) - 2 \cdot x(V_1))^2)}
{4 \cdot mean(\mathbf{w})^2 - (mean(\mathbf{w}) - x(V_1))^2}
}
$.

Finally, 
$
E = \sqrt
{
\frac
{(mean(\mathbf{w})^2 - (mean(\mathbf{w}) - 2 \cdot \frac{(min_{w_j>0}(w_j))^2}{s \cdot \norm{\mathbf{w}}})^2)}
{4 \cdot mean(\mathbf{w})^2 - (mean(\mathbf{w}) - \frac{(min_{w_j>0}(w_j))^2}{s \cdot \norm{\mathbf{w}}})^2}
}
$.

Notice that an analogous line of reasoning might be employed for $\mathsf{A}^\epsilon$, but with its 1-isoline, the second rightmost vertex $V_{n_p-1}$ and the `maximality/minimality property' violated by the value of $\mathsf{A}^\epsilon$ being 1 in points other than $V_{n_p}$. However, because a vertical line at $x = 0$ constitutes the symmetry axis of:
\begin{itemize}
\item the 1-isoline of $\mathsf{A}^\epsilon$ and the 0-isoline of $\mathsf{I}^\epsilon$,
\item the $V_{n_p-1}$ vertex and the $V_{1}$ vertex,
\item the $V_{n_p}$ vertex and the $V_0$ vertex,
\end{itemize}
the derived value of $E$ would be identical.

Generally speaking, $E$ is a function of the aggregation and the WMSD-space coordinates of the second leftmost/rightmost vertex, which depend merely on the vector of weights, so it is generally denoted as $E(\mathit{G},\mathbf{w})$.

In summary, as far as aggregations $\mathsf{I}^\epsilon$ and $\mathsf{A}^\epsilon$ are concerned, the `maximality/minimality property' is violated for every $\epsilon \in (0,E]$. On the other hand, for every $\epsilon \in (E,+\infty)$ the property holds.

\end{document}